\documentclass[11pt, a4paper]{deepmind}

\usepackage{url}
\usepackage[square, numbers, sort&compress]{natbib} 
\usepackage{xcolor}
\usepackage{xspace}
\usepackage{listings}
\usepackage{cleveref}
\usepackage{url}
\usepackage{multirow}
\usepackage{bibunits} 
\usepackage{adjustbox}
\usepackage[tight]{units}

\graphicspath{{figures/}}
\title{Skillful Precipitation Nowcasting using Deep Generative Models of Radar}
\author[*,1]{Suman Ravuri}
\author[*,1]{Karel Lenc}
\author[*,1]{Matthew Willson}
\author[2,3]{Dmitry Kangin}
\author[1]{Remi Lam}
\author[1]{Piotr Mirowski}
\author[2]{{Megan Fitzsimons}}
\author[2]{Maria Athanassiadou}
\author[1]{Sheleem Kashem}
\author[2]{Sam Madge}
\author[2,3]{Rachel Prudden}
\author[1]{Amol Mandhane}
\author[1]{Aidan Clark}
\author[1]{Andrew Brock}
\author[1]{Karen Simonyan}
\author[1]{Raia Hadsell}
\author[2,3]{Niall Robinson}
\author[1]{Ellen Clancy}
\author[$\dagger$,2,4]{Alberto Arribas}
\author[1]{Shakir Mohamed}

\affil[*]{Equal contributions}

\renewcommand{\today}{}

\newcommand{\numforecasters}{56}
\newcommand{\modelname}{the generative method}
\begin{abstract}
Precipitation nowcasting, the high-resolution forecasting of precipitation up to two hours ahead, supports the real-world socio-economic needs of many sectors reliant on weather-dependent decision-making.
State-of-the-art operational nowcasting methods typically advect precipitation fields with radar-based wind estimates, and struggle to capture important non-linear events such as convective initiations.
Recently introduced deep learning methods use radar to directly predict future rain rates, free of physical constraints. While they accurately predict low-intensity rainfall, their operational utility is limited because their lack of constraints produces blurry nowcasts at longer lead times, yielding poor performance on more rare medium-to-heavy rain events.
To address these challenges, we present a Deep Generative Model for the probabilistic nowcasting of precipitation from radar.
Our model produces realistic and spatio-temporally consistent predictions over regions up to \unit[1536]{km} $\times$ \unit[1280]{km} and with lead times from 5--\unit[90]{min} ahead.
In a systematic evaluation by more than fifty expert forecasters from the Met Office, our generative model ranked first for its accuracy and usefulness in 88\% of cases against two competitive methods, demonstrating its decision-making value and ability to provide physical insight to real-world experts. When verified quantitatively, these nowcasts are skillful without resorting to blurring.
We show that generative nowcasting can provide probabilistic predictions that improve forecast value and support operational utility, and at resolutions and lead times where alternative methods struggle.
\end{abstract}

\begin{document}
\maketitle

\begin{bibunit}[ieeetr]

High-resolution forecasts of rainfall and hydrometeors zero to two hours in the future, known as precipitation nowcasting \citep{WMOStatement}, are crucial for weather-dependent decision-making. 
Such nowcasts inform the operations of a wide variety of sectors, including emergency services, energy management, retail, flood early-warning systems, air traffic control, and marine services \citep{wilson2010nowcasting, wmoguidelines2017}. For nowcasting to be useful in these applications the forecast must provide accurate predictions across multiple spatial and temporal scales, account for uncertainty and be verified probabilistically, and perform well on heavier precipitation events that are rarer but more critically affect human life and economy.

Ensemble Numerical Weather Prediction (NWP) systems, which simulate coupled physical equations of the atmosphere to generate multiple realistic precipitation forecasts, are natural candidates for nowcasting since one can derive probabilistic forecasts and uncertainty estimates from the ensemble of future predictions  \citep{toth1997ensemble}. For precipitation at zero to two hours lead time, however, NWPs tend to provide poor forecasts since this is less than the time needed to reach steady-state simulation \citep{pierce2012nowcasting}. 
As a result, alternative methods that make predictions using composite radar observations have been used; radar data is now available every five minutes and at \unit[1]{km} $\times$ \unit[1]{km} grid resolution 
\citep{harrison2015evolution}. Established probabilistic nowcasting methods, such as STEPS and PySTEPS \citep{bowler2006steps, pulkkinen2019pysteps}, follow the NWP approach of using ensembles to account for uncertainty, but model precipitation following the advection equation with a radar source term. In these models, motion fields are estimated by optical flow, smoothness penalties are used to approximate an advection forecast, and stochastic perturbations are added to the motion field and intensity model \citep{germann2004scale, pulkkinen2019pysteps, bowler2006steps}.
These stochastic simulations allow for ensemble nowcasts from which both probabilistic and deterministic forecasts can be derived, and are applicable and consistent at multiple spatial scales, from the kilometer scale to the size of a catchment area \cite{imhoff2020spatial}. %

Approaches based on deep learning have been developed that move beyond reliance on the advection equation \citep{ayzel2020rainnet, lebedev2019precipitation, agrawal2019machine, xingjian2015convolutional, foresti2019using, sonderby2020metnet, shi2017deep}. By training these models on large corpora of radar observations rather than relying on in-built physical assumptions, deep learning methods aim to better model traditionally difficult non-linear precipitation phenomena, such as convective initiation and high precipitation. This class of methods directly predicts precipitation rates at each grid location,
and models have been developed for both deterministic and probabilistic forecasts. 
As a result of their direct optimization and fewer inductive biases, the forecast quality of deep learning methods---as measured by per-grid-cell metrics such as Critical Success Index (CSI) \citep{jolliffe2012forecast} at low precipitation levels (less than \unit[2]{mm/hr})---has greatly improved.

While existing deep learning systems have improved forecast quality, indirect evidence suggests the resulting nowcasts do not allow end users to make better decisions, i.e. do not improve forecast value.
As a number of authors have noted \cite{ayzel2020rainnet, shi2017deep}, forecasts issued by current deep learning systems 
express uncertainty at increasing lead times with blurrier precipitation fields,
and may not include small-scale weather patterns important for improving forecast value.
Furthermore, the focus in existing approaches on location-specific predictions, rather than probabilistic predictions of entire precipitation fields, limits their operational utility and usefulness, being unable to provide simultaneously consistent predictions across multiple spatial and temporal aggregations. The ability to make skillful probabilistic predictions is also known to provide greater economic and decision-making value than deterministic forecasts \citep{palmer2002quantifying, richardson2000skill}.

In this work, we demonstrate that improving the
skill of probabilistic precipitation nowcasting
improves the value of precipitation nowcasts.
To create these more skillful predictions,
we develop an observations-driven approach for probabilistic nowcasting using Deep Generative Models (DGMs). DGMs are statistical models that learn probability distributions of data, and allow for easy generation of samples from their learned distributions. Since generative models are fundamentally probabilistic they have the ability to simulate many samples from the conditional distribution of future radar given historical radar, generating a collection of forecasts similar to ensemble methods. The ability of DGMs to both learn from observational data as well as represent uncertainty across multiple spatial and temporal scales makes them a powerful method for developing new types of operationally-useful nowcasting. 
These models can predict smaller-scale weather phenomena that are inherently difficult to predict due to underlying stochasticity, which is a critical issue for nowcasting research. DGMs 
predict the location of precipitation as accurately as systems tuned to this task while preserving spatio-temporal properties useful for decision-making. Importantly, they are judged by professional forecasters as substantially more accurate and useful than PySTEPS or deep learning systems.

\section*{Generative Models of Radar}
Our nowcasting algorithm is a conditional generative model that predicts $N$ future radar fields given $M$ past, or contextual, radar fields, using radar-based estimates of surface precipitation
$X_T$ at a given time point $T$. Our model includes
latent random variables $Z$ and parameters $\boldsymbol{\theta}$, described by
\begin{equation}
P(X_{M+1:M+N}|X_{1:M}) = \int P(X_{M+1:M+N}|\mathbf{Z}, X_{1:M}, \boldsymbol{\theta})P(\mathbf{Z}|X_{1:M}) d\mathbf{Z}.
\label{eq:prob_model}
\end{equation}

\begin{figure}[tp]
     \begin{minipage}{0.49\textwidth}
    \textbf{a}\\
    \includegraphics[width=.95\textwidth]{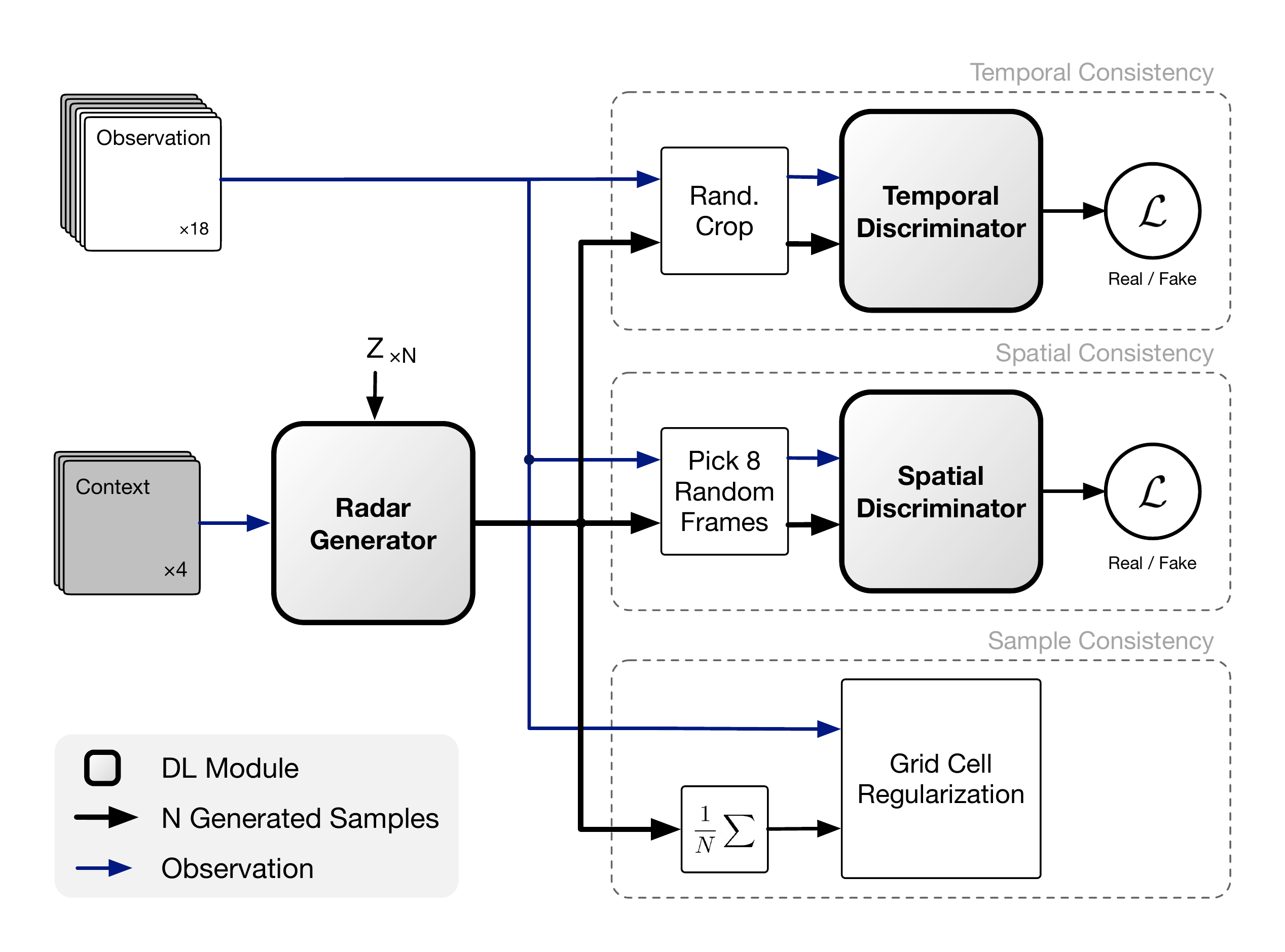}\\
    \vspace{5mm}
    \textbf{b}\\
    \includegraphics[width=.8\textwidth]{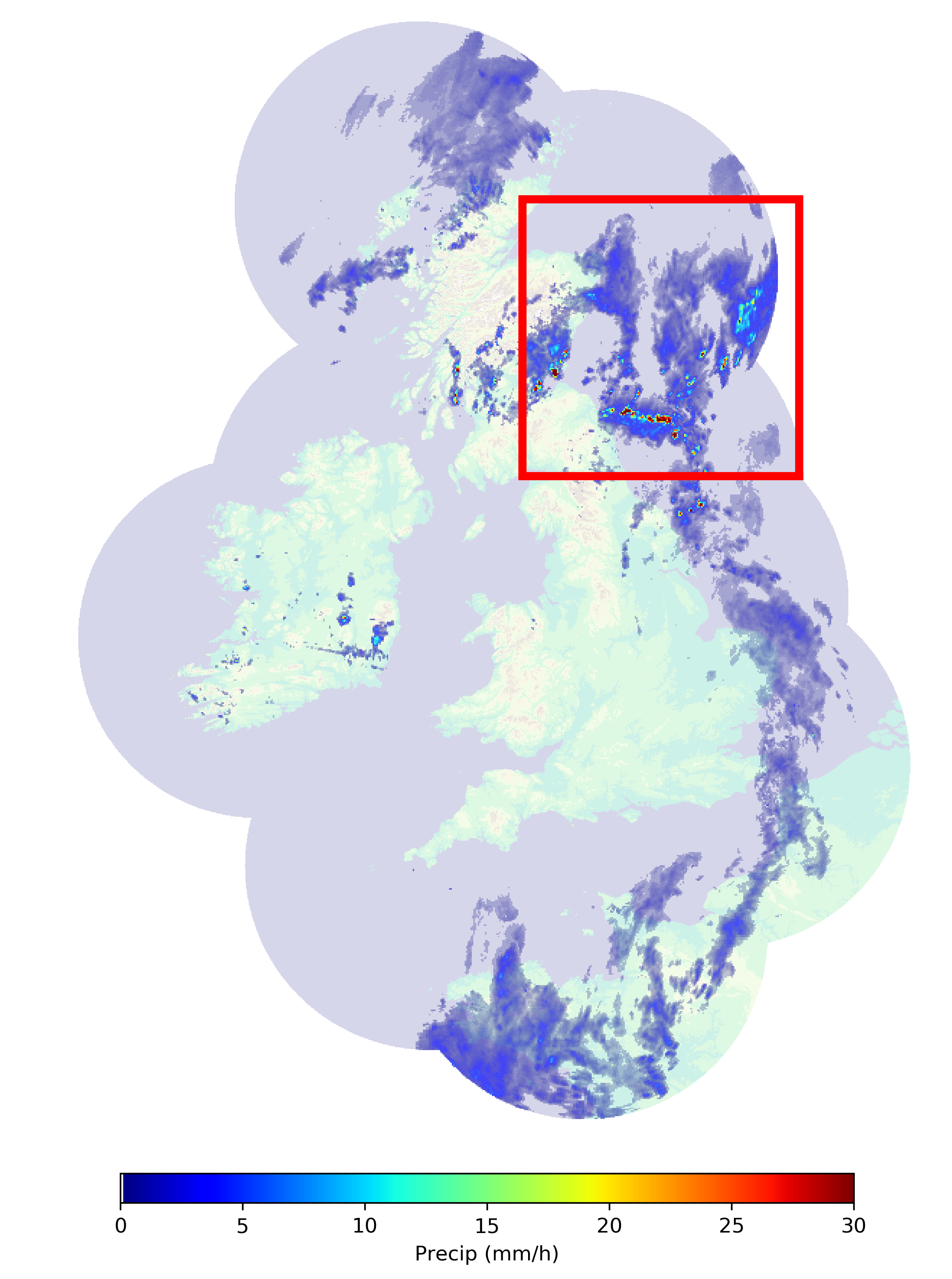}\\
    \end{minipage}
    \begin{minipage}{0.49\textwidth}
    \newcommand{\insertevent}[1]{\includegraphics[width=0.32\textwidth]{generated_qualitative_figures_20210211/event_a2_white/#1}}
    \textbf{c}\\
    \setlength{\tabcolsep}{0pt}
    \begin{tabular}{m{1em} c c c}
        & T+\unit[30]{min} & T+\unit[60]{min} & T+\unit[90]{min} \\ 
        \vspace{-25mm}
        \rotatebox{90}{\hspace{0.8em}Observations} &
            \insertevent{steps_target_30.png} &
            \insertevent{steps_target_60.png} &
            \insertevent{steps_target_90.png}
            \\
            \vspace{-25mm}
        \rotatebox{90}{\hspace{0.7em}Gen. Method} &
            \insertevent{ganspd256_pred_samples_30.png} &
            \insertevent{ganspd256_pred_samples_60.png} &
            \insertevent{ganspd256_pred_samples_90.png}
            \\
        \vspace{-25mm}
        \rotatebox{90}{\hspace{1.7em}PySTEPS} &
            \insertevent{steps_pred_samples_30.png} &
            \insertevent{steps_pred_samples_60.png} &
            \insertevent{steps_pred_samples_90.png}
            \\
        \vspace{-25mm}
        \rotatebox{90}{\hspace{2.5em}UNet} &
            \insertevent{unet_pred_30.png} &
            \insertevent{unet_pred_60.png} &
            \insertevent{unet_pred_90.png}
            \\
        \vspace{-25mm}
        \rotatebox{90}{~Axial Attention} &
            \insertevent{metnet_pred_30.png} &
			\insertevent{metnet_pred_60.png} &
			\insertevent{metnet_pred_90.png}
            \\
            \\
    \end{tabular}
    \end{minipage}
    \caption{\textbf{Model overview and case study of performance on a challenging precipitation event starting at T=2019-06-24 at 16:15 UK, showing convective cells over eastern Scotland. The Generative Method is better able to predict the spatial coverage and convection compared to other methods over a longer time period, while not over-estimating the intensities, and is significantly preferred by forecasters (93\% first choice, $N=\numforecasters$, $p<10^{-4}$).} 
    \textbf{a:} Schematic diagram of the model architecture showing the generator with spatial latent variables $Z$. 
    \textbf{b:} Geographic context for predictions at T+0. \textbf{c:} A single prediction at T+\unit[30], T+\unit[60], and T+\unit[90]{mins} lead time for different models. Critical Success Index (CSI) at thresholds \unit[2]{mm/hr} and \unit[8]{mm/hr} and Continuous Ranked Probability Score (CRPS) for an ensemble of 4 samples shown in the bottom left corner. For Axial Attention we show the mode prediction.
    }
    
    \label{fig:model-visualization-a}
\end{figure}

The integration over latent variables ensures that the model makes predictions that are spatially dependent.
Learning is framed in the algorithmic framework of a conditional Generative Adversarial Network \citep{goodfellow2014generative, brock2018large, mirza2014conditional}, specialized for the precipitation prediction problem. 
Four consecutive radar observations (the previous \unit[20]{mins}) are used as context for a generator (figure \ref{fig:model-visualization-a}a) that allows sampling of multiple realizations of future precipitation, each realization being 18 frames (\unit[90]{minutes}).

Learning is driven by two loss functions and a regularization term, which guide parameter adjustment by comparing real radar observations to those generated by the model. 
The first loss is defined by a spatial discriminator, which is a convolutional neural network that aims to distinguish individual observed radar fields from generated fields, and ensures spatial consistency, and discourages blurry predictions.
The second loss is defined by a temporal discriminator, which is a 3D convolutional neural network that aims to distinguish observed and generated radar sequences, and
imposes temporal consistency, and penalizes jumpy predictions.
These two discriminators share similar architectures to existing work in video generation \cite{clark2019efficient}. When used alone, these losses lead to accuracy on par with Eulerian persistence.
To improve accuracy, we introduce a regularization term that penalizes deviations at the grid cell level between the real radar sequences and the model predictive mean (computed with multiple samples).
This third term is important for the model to produce location-accurate predictions and improve performance.
In the Supplement, we show an ablation study supporting the necessity of each loss term.
Finally, we introduce a fully convolutional latent module for the generator, allowing for predictions over precipitation fields larger than the ones used at training time, while maintaining spatio-temporal consistency.

The model is trained on a large corpus of precipitation events, which are 256 $\times$ 256 crops extracted from the radar stream, of length \unit[110]{minutes} (22 frames). An importance sampling scheme is used to create a dataset more representative of high precipitation (see Methods). Throughout, all models are trained on radar observations for the UK for years 2016-2018, and evaluated on a test set from 2019. 
Analysis on other data splits and the US is reported in the Supplement.
Once trained, this model allows fast full-resolution nowcasts to be produced, with a single prediction (using an NVIDIA V100 GPU) needing just over a second to generate.

\section*{Intercomparison Case Study}

We use a single case study to compare the nowcasting performance of the generative method to three strong baselines: PySTEPS, a widely-used precipitation nowcasting system based on ensembles, considered to be state-of-the-art \citep{bowler2006steps, pulkkinen2019pysteps, imhoff2020spatial}; UNet, a popular deep learning method for nowcasting \cite{agrawal2019machine}; and to an Axial Attention model, a radar-only implementation of MetNet \cite{sonderby2020metnet}. Three case studies were chosen by an expert forecaster as meteorologically challenging, and as useful for understanding the value of competing methods on real-world events. We expand one of these cases in Figure \ref{fig:model-visualization-a}b,c, showing the ground truth and predicted precipitation fields at T+\unit[30], T+\unit[60] and T+\unit[90]{mins}, quantitative scores on different verification metrics, and comparisons of expert forecaster preferences among the competing methods; the two other cases are included in the Extended Data. 

The event in Figure \ref{fig:model-visualization-a} starts at 2019-06-24 at 16:15 and  shows convective cells in eastern Scotland (Fife, the Firth of Forth and the North Sea), with intense showers making landfall over Fife. Maintaining such cells is difficult and a traditional method such as PySTEPS overestimates the rainfall intensity over time, which is not observed in reality, and does not sufficiently cover the spatial extent of the rainfall.
The UNet and Axial Attention models
roughly predict the location of rain, but due to aggressive blurring, over-predict areas of rain, miss intensity, and fail to capture any small-scale structure.
By comparison, the generative method preserves a good spatial envelope, representing the convection over the Firth of Forth, and  maintains the heavy rainfall in the early prediction although with less accurate rates at T+\unit[90]{min} and at the edge of the radar than at previous time steps. When expert forecasters judged these predictions against ground truth observations, they significantly preferred the generative nowcasts, with 93\% of forecasters choosing it as their first choice; see figure \ref{fig:forecaster_preferences}a. 

In addition to the expert assessment, the nowcasts in figure \ref{fig:model-visualization-a}c also indicate  two widely-used skill scores, the Critical Success Index (CSI) and Continuous Rank Probability Score (CRPS) \citep{jolliffe2012forecast}.
As the scores show, UNet, Axial Attention, and the generative method all outperform PySTEPS on CSI and perform similarly to each other. Since their predictions are qualitatively different, however, they are judged as significantly different by experts, and the scores do not measure whether the forecasts capture other salient qualities of precipitation.
This study highlights a limitation of using these popular metrics to evaluate deep learning systems: while standard metrics implicitly assume that models, such as NWPs and advection-based systems, preserve the physical plausibility of forecasts, deep learning systems may outperform on certain metrics by failing to satisfy other characteristics of useful predictions. 

\section*{Quantitative Evaluation}

We verify the performance of competing methods using a suite of metrics as is standard practice, since no single verification score can capture all desired properties of a forecast. We report the Critical Success Index (CSI) \cite{schaefer1990critical} to measure location accuracy of the forecast at various rain rates. 
We report the radially-averaged power spectral density (PSD) \cite{harris2001multiscale, sinclair2005empirical} to compare the precipitation variability of nowcasts to that of the radar observations. We report the continuous ranked probability score (CRPS) \cite{gneiting2007strictly} to determine how well the probabilistic forecast aligns with the ground truth. Details of these metrics and results on other standard metrics can be found in the Supplement. We report results here using data from the UK, and results consistent with these showing generalization of the method on a dataset from the US in the Extended Data. 

\begin{figure}[htp]
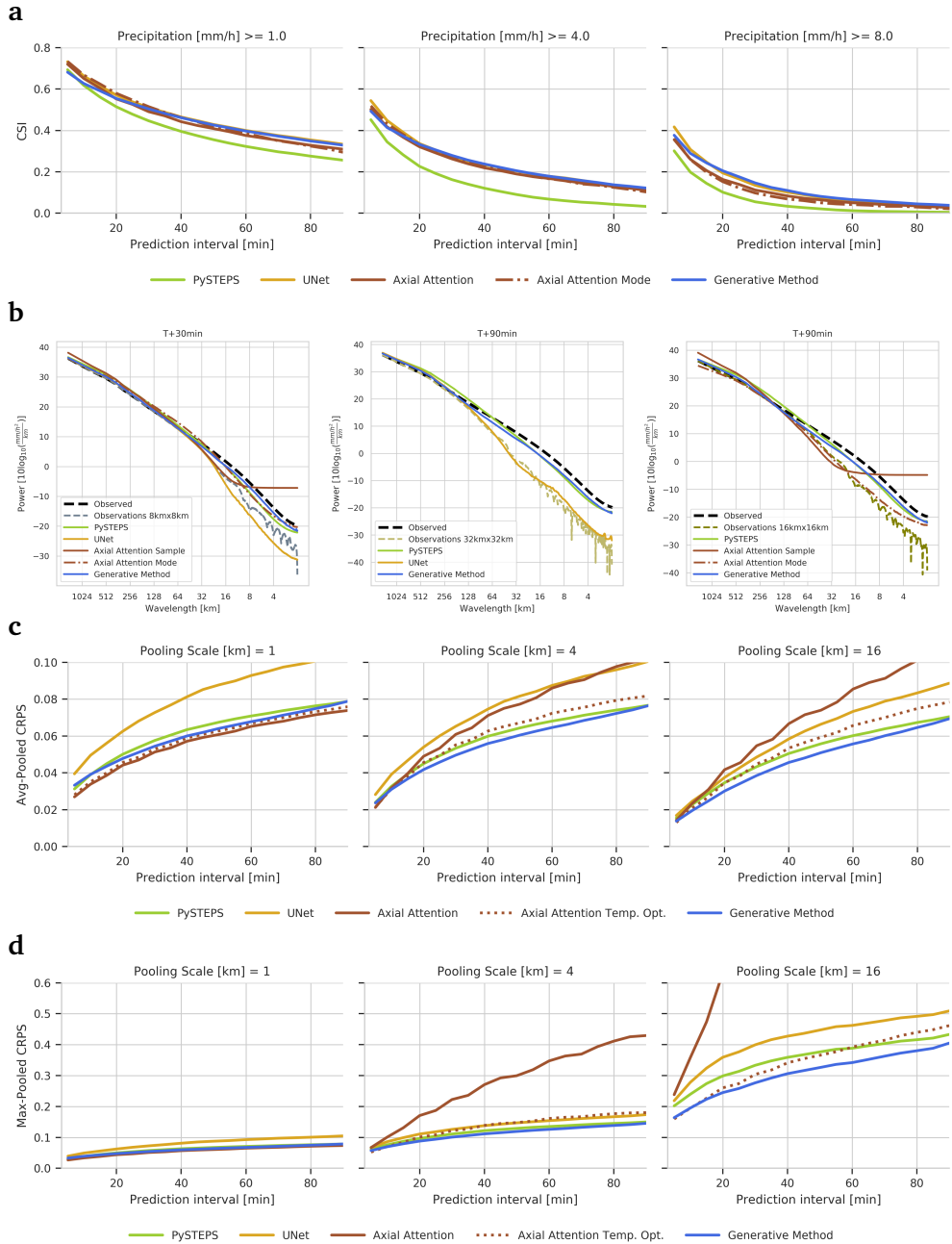

    \centering
    \newcommand{\insertfigure}[3]{\includegraphics[width=#1\textwidth]{figures/figures_UK/yearly/single_run/#2/#3}}
    \begin{minipage}{0.8\textwidth}
        \begin{minipage}{0.99\textwidth}
        \textbf{a}\\
        \insertfigure{0.99}{20210210}{prob-csicc-uk512-yearly-test-1-4-8classifier-sample_mean-matching_thresholds-csi.png}
        \end{minipage} \\
        \begin{minipage}{0.99\textwidth}
        \textbf{b} \\
		\insertfigure{0.32}{20210210}{custom_psd_nimrod_yearly_splits_30min_8kmObs.png}
		\insertfigure{0.32}{20210210}{custom_psd_nimrod_yearly_splits_90min_32kmObs_unet.png}
		\insertfigure{0.32}{20210210}{custom_psd_nimrod_yearly_splits_90min_16kmObs_aa.png}
        \end{minipage} \\
        \begin{minipage}{0.99\textwidth}
        \textbf{c}\\
        \insertfigure{0.99}{20210210}{prob-crps-uk512-yearly-test-1-4-8crps-crps.png}
        \end{minipage} \\
        \begin{minipage}{0.99\textwidth}
        \textbf{d}\\
        \insertfigure{0.99}{20210210}{prob-crps-uk512-yearly-test-1-4-8crps_max-crps.png}
        \end{minipage} \\

    \end{minipage}
    
    \caption{ \textbf{Verification scores for the United Kingdom in 2019. While all deep learning methods perform similarly on CSI, the UNet and Axial Attention methods achieve these scores by blurring predictions, as shown in the radially-averaged PSD plot. The spectral properties of the generative method, by contrast, mirrors the radar observations. Furthermore, the generative method has similarly strong CRPS performance across spatial scales, unlike the baseline models.} \\
    \textbf{a:} Critical Success Index across 20 samples of different models across precipitation thresholds \unit[1]{mm/hr} (left), \unit[4]{mm/hr}, \unit[8]{mm/hr} (right). We also report results for the Axial Attention mode prediction.
    \textbf{b:} Radially-averaged power spectral density for full-frame 2019 predictions for all models at T+\unit[30]{min} (left) and T+\unit[90]{min} (middle and right). At T+\unit[90]{min}, the effective resolution is \unit[32]{km} for the UNet (middle plot) and \unit[16]{km} for the Axial Attention (both sample and mode prediction).
    \textbf{c:} CRPS of various models for original predictions (left), average rain rate over a catchment area of size \unit[4]{km} $\times$ \unit[4]{km} (middle) and \unit[16]{km} $\times$ \unit[16]{km} (right).
    \textbf{d:} CRPS of various models for original predictions (left), maximum rain rate over a catchment area of size \unit[4]{km} $\times$ \unit[4]{km} (middle) and \unit[16]{km} $\times$ \unit[16]{km} (right).
     }
    \label{fig:uk_yearly_test_results}
\end{figure}

Figure \ref{fig:uk_yearly_test_results}a shows that all three deep learning systems produce forecasts that are significantly more location-accurate than the PySTEPS baseline when compared using CSI. 
We assessed the statistical significance across lead times and thresholds, using paired permutation tests with alternating weeks as independent units. In particular, the generative method has significant skill compared to PySTEPS for all precipitation thresholds ($N=26$, $p<10^{-4}$); see Methods. 

The PSD in Figure \ref{fig:uk_yearly_test_results}b shows that both the generative method and PySTEPS match the observations in their spectral characteristics, but the Axial Attention and UNet models produce forecasts with medium- and small-scale precipitation variability that decreases with increasing lead time. Since they produce blurred predictions, the effective resolution of the Axial Attention and UNet nowcasts is far less than the \unit[1]{km} $\times$ \unit[1]{km} resolution of the data. At T+\unit[30]{min} lead time, UNet forecasts share the same precipitation variability as radar upscaled to \unit[8]{km} $\times$ \unit[8]{km}. At T+\unit[90]{min}, its effective resolution is \unit[32]{km} $\times$ \unit[32]{km}. Since the Axial Attention model uses a location-specific probabilistic output distribution, sample forecasts from the model introduce high frequency noise at wavelengths \unit[8]{km} and \unit[16]{km} for T+\unit[30]{min} and T+\unit[90]{min}, respectively. At T+\unit[90]{min}, Axial Attention has
the same precipitation variability as radar upscaled to \unit[16]{km} $\times$ \unit[16]{km}, which reduces the value of the nowcasts for forecasters. 

For probabilistic verification, figure \ref{fig:uk_yearly_test_results} c,d shows the CRPS of the average and maximum precipitation rate aggregated over regions of increasing size \cite{gilleland2009intercomparison}, which assesses the forecast behaviour across space. When measured at the grid-resolution level, the generative method, PySTEPS, and Axial Attention perform similarly; we also show an Axial Attention model with improved performance obtained by rescaling its output probabilities \cite{guo2017calibration} (denoted Axial Attention-Temp-Opt).
As the spatial aggregation is increased, the generative method and PySTEPS share similarly strong performance, with the generative method performing better on maximum precipitation. The Axial Attention model is significantly poorer for larger aggregations and underperforms all other methods at scale four and above. 
Using alternating weeks as independent units, paired permutation tests show that the performance differences between the generative method and the Axial Attention-Temp-Opt are significant ($N=26$, $p<10^{-3}$).

In additional quantitative analysis, we show that our approach generalizes across seasons, types of rain, and regions. We also show evaluation on other metrics such as fractions skill score (FSS), performance on a data split over weeks rather than years, and evaluation showing the inability of NWPs to make predictions at nowcasting timescales; see Extended Data Figures \ref{fig:uk_yearly_test_results2}--\ref{fig:uk_weekly_test_results}.

Together, these results show that the generative method verifies competitively compared to alternatives: it outperforms (on CSI) the incumbent STEPS nowcasting approach,
provides probabilistic forecasts that are more location accurate, and preserves the statistical properties of precipitation across spatial and temporal scales without blurring, whereas other deep learning methods do so at the expense of them.

\section*{Expert Forecaster Assessment}
\begin{figure}[tb]
     \centering
    \begin{minipage}{0.99\textwidth}
        \begin{minipage}{0.49\textwidth}
    \textbf{a}\\
    \includegraphics[width=\textwidth]{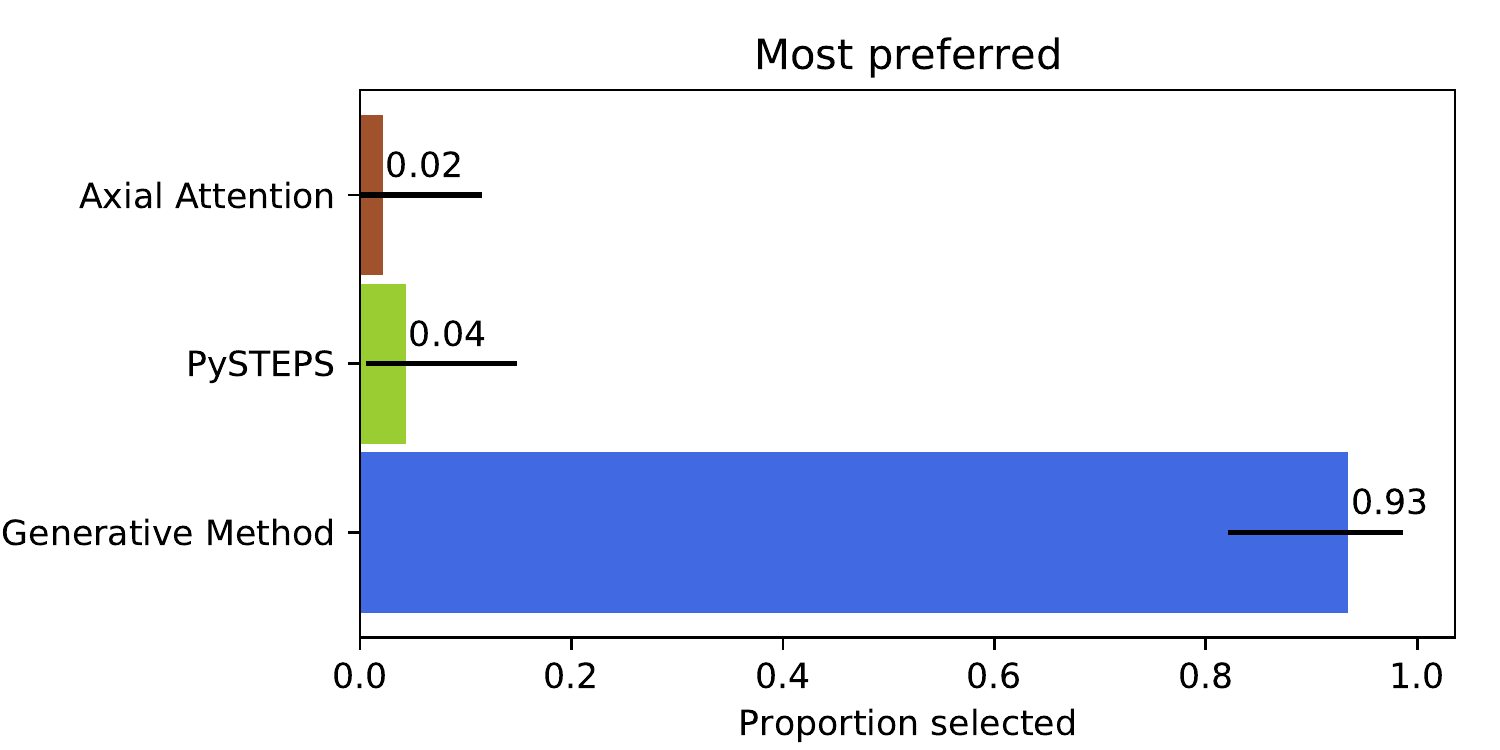}
    \end{minipage}
    \begin{minipage}{0.35\textwidth}
    \textbf{b}\\
    \includegraphics[width=\textwidth]{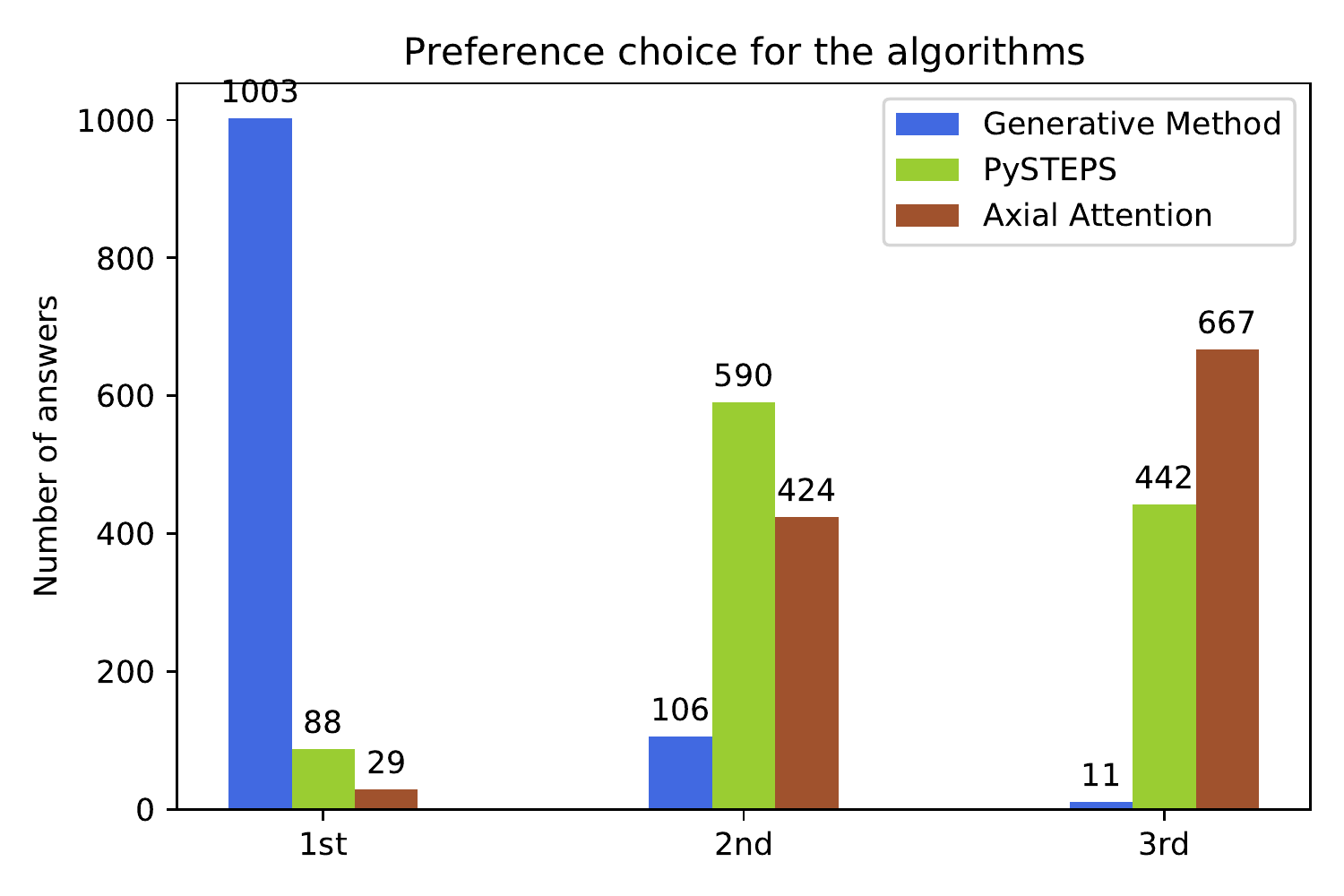}
    \end{minipage}
    \\
        \begin{minipage}{0.45\textwidth}
        \textbf{c}\\
        \includegraphics[width=\textwidth]{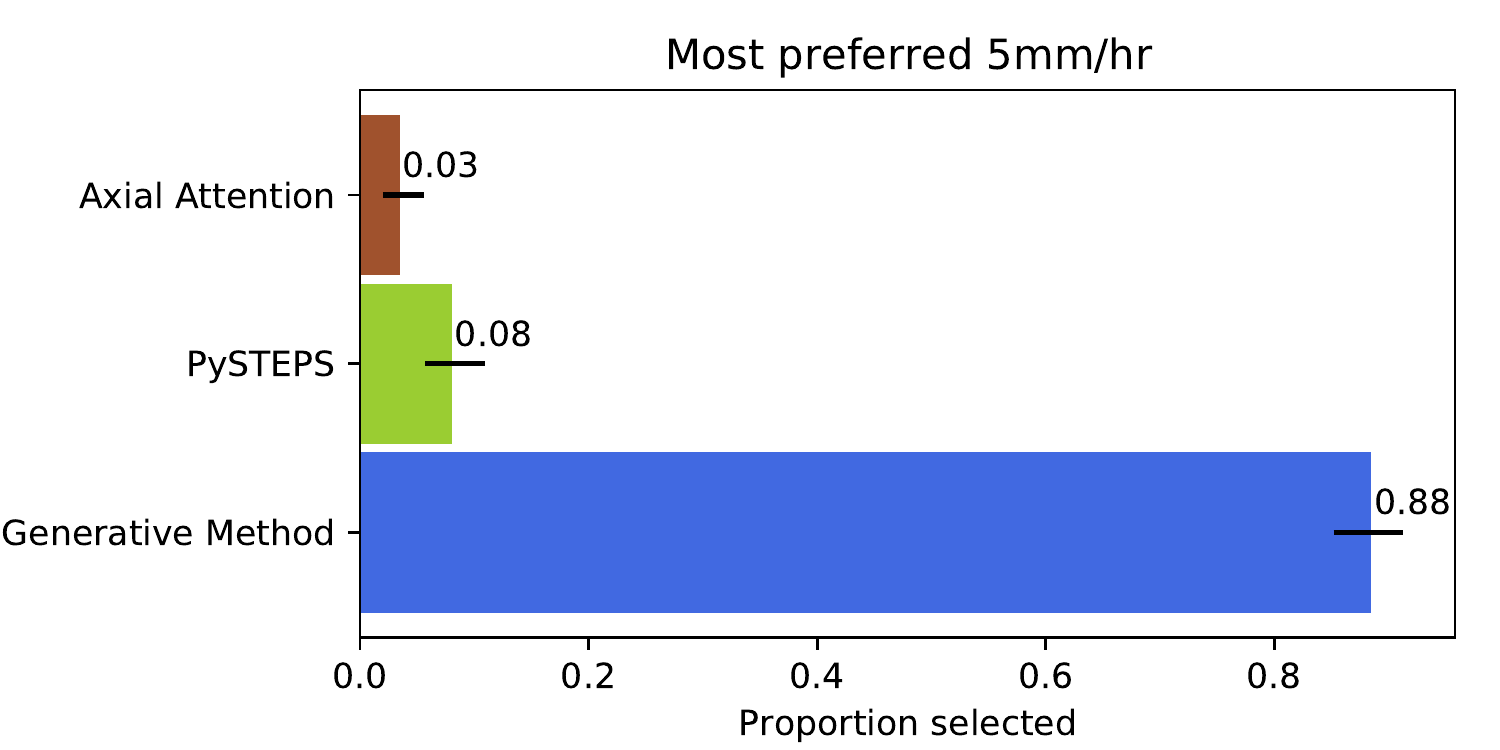}
        \end{minipage}
        \begin{minipage}{0.45\textwidth}
        \textbf{d}\\
        \includegraphics[width=\textwidth]{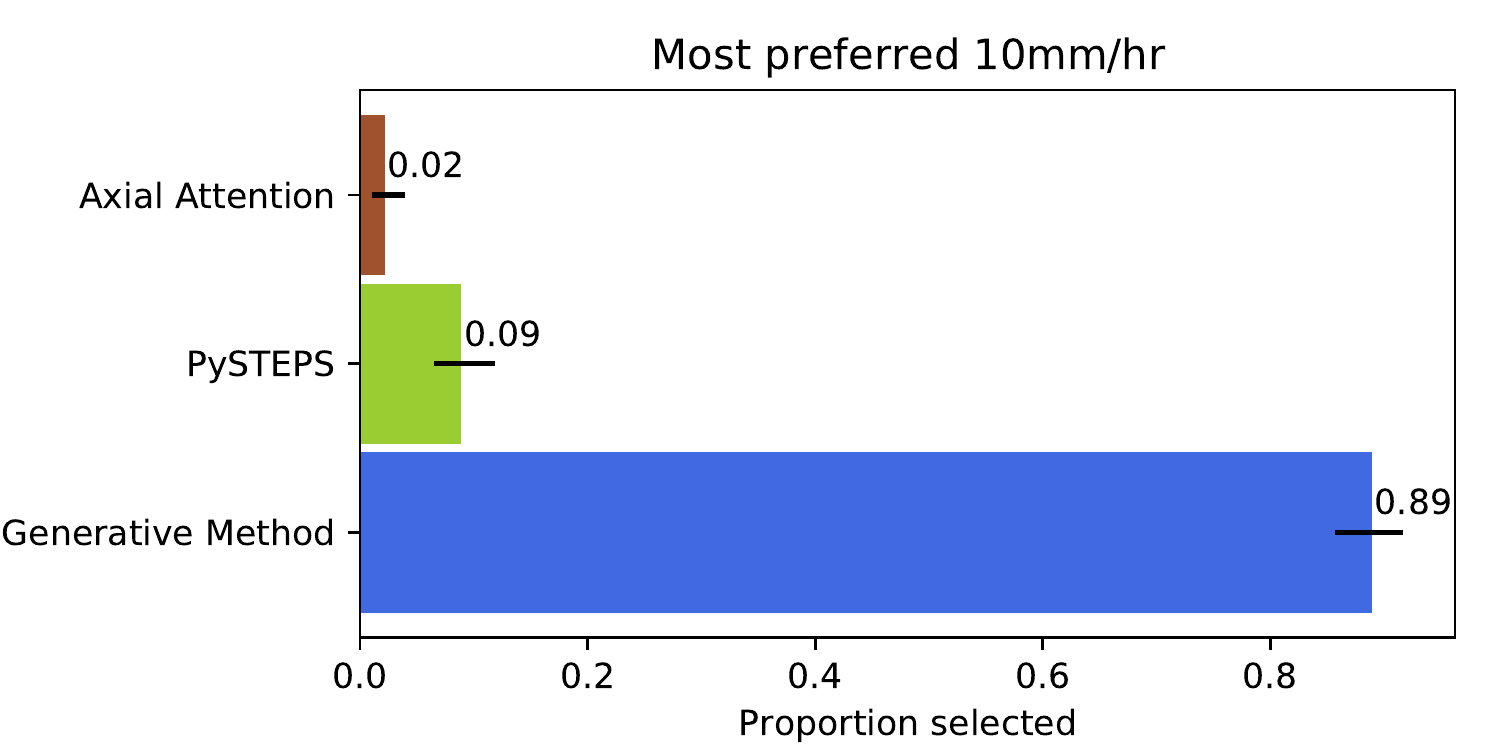}
        \end{minipage}
    \end{minipage}
    
     \caption{\textbf{Expert forecaster results. Results show that forecasters significantly preferred the Generative Method to alternatives ($N=\numforecasters$, $p<10^{-4}$).} 
     \textbf{a:} Forecaster preferences for the case study in figure \ref{fig:model-visualization-a}.
     \textbf{b:} Counts of forecaster ranking assignments amongst the three alternatives assessed.
     \textbf{c:} Forecaster rankings for medium rain (\unit[5]{mm/hr}) cases.
     \textbf{d:} Forecaster rankings for heavy rain (\unit[10]{mm/hr}) cases.
      We report the percentage of forecasters for their first choice rating as well as the Clopper-Pearson 95\% confidence interval. The experimental assessment involved two types of rain-intensity with a total of 10 trials for each type per forecaster. 
     }
     \label{fig:forecaster_preferences}
 \end{figure}

To directly assess the decision-making value and operational usefulness of the generative nowcasting approach, we use the judgments of \numforecasters~expert forecasters working in the 24/7 operational center at the Met Office (the UK's national meteorology service). This assessment was conducted using a two-phase experimental protocol. In phase 1, all forecasters were asked to provided a ranked preference assessment on a set of nowcasts with the instruction that `preference is based on [their] opinion of accuracy and value'. 20\% of forecasters were randomly selected to participate in a phase 2 retrospective recall interview \citep{crandall2013cognitive}.

Operational forecasters seek utility in forecasts for critical events, safety, and planning guidance. Therefore, to make meaningful statements of operational usefulness, our evaluation assessed nowcasts for high-intensity events, specifically medium rain (rates above \unit[5]{mm/hr}, top 10\% of rain events) and heavy rain (rates above \unit[10]{mm/hr}, top 1\% of rain events). Forecasters were asked to rank their preferences on a sample of 20 nowcasts (from a corpus of 2,126 high-intensity precipitation events in 2019). 
Data was presented in the form shown in figure \ref{fig:model-visualization-a}b,c, showing clearly the initial context at T+0, the ground truth at T+\unit[30]{min}, T+\unit[60]{min}, and T+\unit[90]{min}, and nowcasts from PySTEPS, Axial Attention, and \modelname. The identity of the methods in each panel was anonymized and their order randomized. 
See the Methods for further details of the protocol and of the ethics approval for human subjects research.

Figure \ref{fig:forecaster_preferences}b shows the counts of all rankings. The generative nowcasting approach was significantly preferred by forecasters when asked to make judgments of accuracy and value of the nowcast, being their most preferred 88\% of the time for the \unit[5]{mm/hr} nowcasts (figure \ref{fig:forecaster_preferences}c; $p<10^{-4}$), and 89\% for the \unit[10]{mm/hr} nowcasts (figure \ref{fig:forecaster_preferences}d, $p<10^{-4}$). We compute the $p$-value assessing the binary decision whether forecasters chose \modelname~as their first choice using a permutation test with $10,000$ resamplings, and indicate the Clopper-Pearson 95\% confidence interval. 

This significant forecaster preference is important since it is strong evidence that generative nowcasting can provide to forecasters critical physical insight not provided by alternative methods. 
In the subsequent structured interviews, PySTEPS was described as `being too developmental which would be misleading', i.e. as having many `positional errors' and `much higher intensity compared with reality'. The Axial Attention model was described as `too bland', i.e. as being `blocky' and `unrealistic', but had `good spatial extent'. 
Forecasters described \modelname~as having the `best envelope', `representing the risk best', as having `much higher detail compared to what [expert forecasters] are used to at the moment', and as capturing `both the size of convection cells and intensity the best'. In the cases where forecasters chose PySTEPS or the Axial Attention as their first choice, they pointed out that the generative method showed decay in the intensity for heavy rainfall at T+\unit[90]{min} and had difficulty predicting isolated showers, which are important future improvements for the method.

\section*{Conclusion}

We showed that Deep Generative Models are a valuable forecasting tool, providing fast and accurate short-term predictions for lead times where existing methods struggle.
Yet, there remain challenges for our approach to probabilistic nowcasting. Rank histograms show that ensemble members from the generative approach are under-dispersed compared to PySTEPS.
As the forecaster assessment demonstrated, our generative method provides skillful predictions compared to other solutions, but the prediction of high precipitation at long lead times remains difficult for all approaches.
Critically, our work reveals that standard verification metrics and expert judgments are not mutually indicative of value, highlighting the need for newer quantitative measurements that are better aligned with operational utility when evaluating models with few inductive biases and high capacity.

Skillful nowcasting is a long-standing problem of importance for much of weather-dependent decision-making. Our approach using Deep Generative Models directly tackles this important problem, improves upon existing solutions, and provides the insight needed for real-world decision-makers. Whereas existing practice focuses on quantitative improvements without concern for operational utility, we hope this work will serve as a foundation for new data, code and verification methods---as well as the greater integration of machine learning and environmental science in forecasting larger sets of environmental variables---that makes it possible to both provide competitive verification and operational utility.

\putbib[refs]
\end{bibunit}

\begin{bibunit}[ieeetr]
\renewcommand{\figurename}{Extended Data Figure}
\setcounter{figure}{0}  
\newpage
\section*{Methods}
We provide additional details of the data, models and evaluation here, with references to extended data that add to the results provided in the main text.

\section*{Datasets}
A dataset of radar for the United Kingdom  was used for all the experiments in the main text. Additional quantitative results on a United States dataset are available in Supplement \ref{sec:suppl-datasets}. 

\subsection*{United Kingdom Dataset}
\label{sub:nimrod_dataset}
To train and evaluate nowcasting models over the United Kingdom, we use a collection of radar composites from the Met Office RadarNet4 network.
This network comprises more than 15 operational, proprietary C-band dual polarization radars covering $99\%$ of the UK (see Figure 1 in \cite{fairman17climatology}).
We refer to \citep{harrison2015evolution} for details about how radar reflectivity is post-processed to obtain the 2D radar composite field, which includes orographic enhancement and mean field adjustment using rain gauges.
Each grid cell in the $1536 \times 1280$ composite represents the surface-level precipitation rate (in \unit{mm/hr}) over a \unit[1]{km} $\times$ \unit[1]{km} region in the OSGB36 coordinate system.
If a precipitation rate is missing (e.g., because the location is not covered by any radar, or if a radar is out of order), the corresponding grid cell is assigned a negative value which is used to mask the grid cell at training and evaluation time.
The radar composites are quantized in increments of \unit[1/32]{mm/hr}.

We use radar collected every five minutes between 1 January 2016 and 31 December  2019. We use the following data splits for model development. Fields from the first day of each month from 2016 to 2018 are assigned to the validation set. All other days from 2016 to 2018 are assigned to the training set. Finally, data from 2019 is used for the test set, preventing data leakage and testing for out of distribution generalization. For further experiments testing in-distribution performance using a different data split, see section \ref{sec:suppl_baselines} of the supplementary material.

\subsection*{Training Set Preparation}
\label{sub:dataset_preparation}

Most of the radar composites contain little to no rain.
Supplementary Table \ref{tab:dataset-distn} shows that approximately 89\%  
of grid cells contain no rain in the United Kingdom. Medium to high precipitation (using rain rate above \unit[4]{mm/hr}) comprises fewer than 0.4\% of grid cells in the dataset. To account for this unbalanced distribution, the dataset is rebalanced to include more data with higher precipitation radar observations, which allows the models to learn useful precipitation predictions.

Each example in the dataset is a sequence of 24 radar observations of size $1536 \times 1280$, representing two continuous hours of data. The maximum rain rate is capped at \unit[128]{mm/hr}, and sequences that are missing one or more radar observations are removed.
$256\times256$ crops are extracted every 32 grid cells, and an importance sampling scheme is used to reduce the number of examples containing little precipitation.
We describe this importance sampling and the parameters used in the Supplement \ref{sec:importance-sampling-scheme}.
After subsampling and removing entirely-masked examples, the number of examples in the training set is roughly 1.5 million.

\section*{Model Details and Baselines}
\label{sec:models}
Here, we describe the proposed method and the three baselines to which we compare performance. When applicable, we describe both the architectures of the models and the training methods. There is a wealth of prior work, and we survey them as additional background in Supplement \ref{sup:related_work}. 

\subsection*{Generative Method}
\label{sub:models-gan}

A high-level description of the model was given in the main text and in Figure \ref{fig:model-visualization-a}a, which we expand here. The model is a generator that is trained using two discriminators and an additional regularization term. The generator is fully convolutional, and both the generator and discriminators use spectrally normalized convolutions throughout, similar to \cite{zhang2019self} .

\begin{figure}[tbp]
    \textbf{a} \\
    \includegraphics[width=0.8\linewidth]{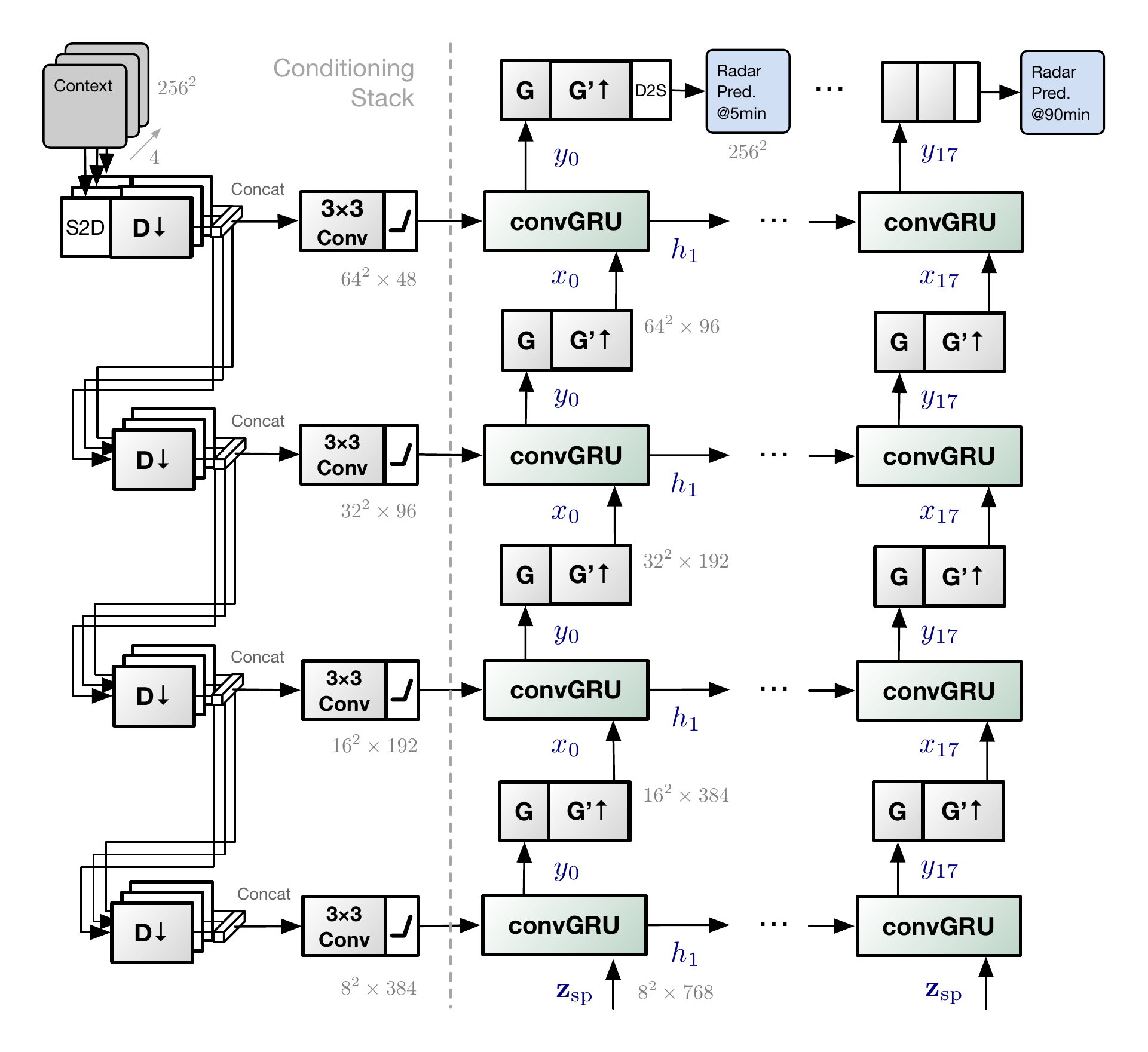} \\
    \textbf{b} \\
    \includegraphics[width=0.95\linewidth]{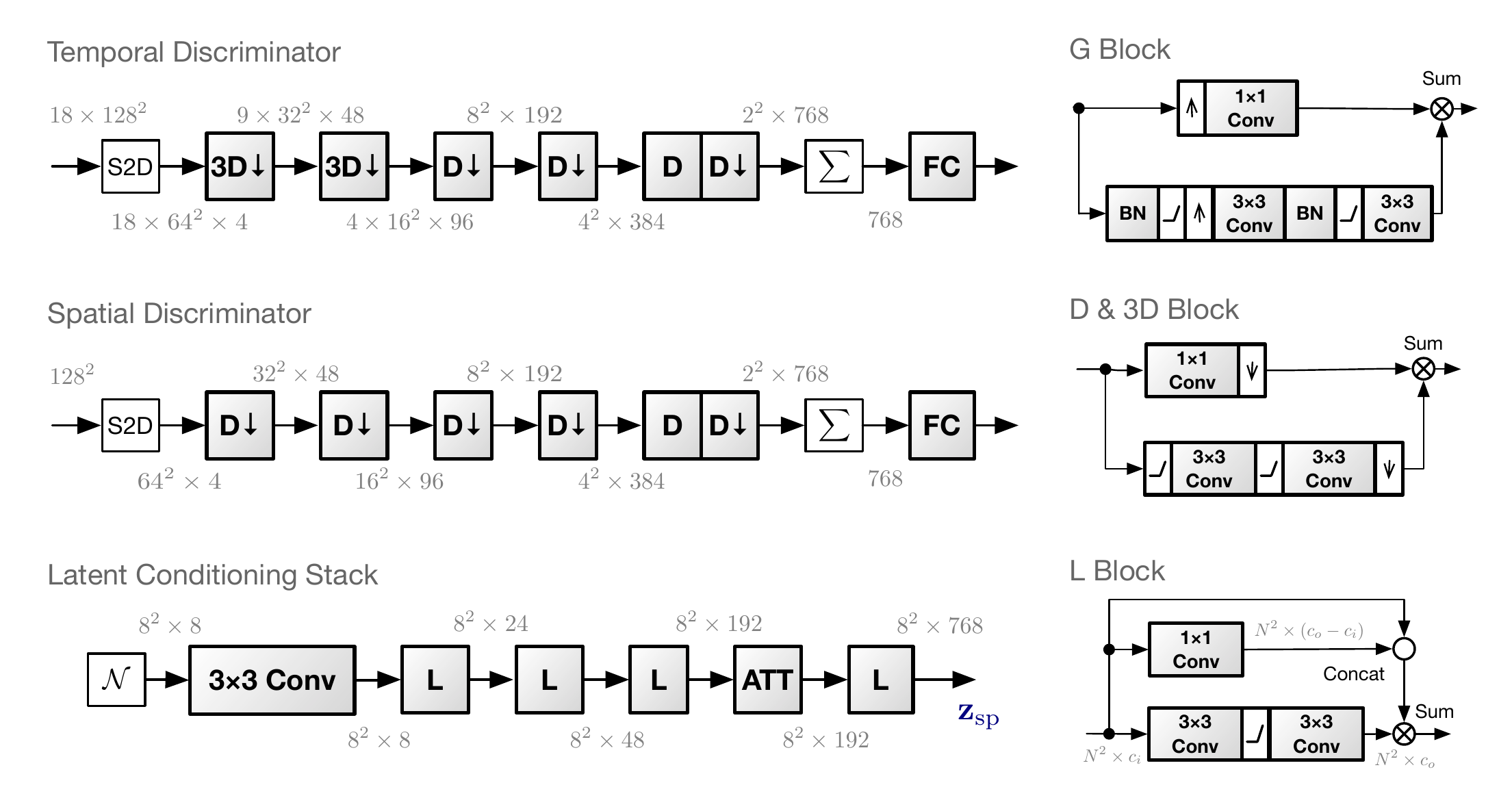}
    \caption{Detailed architecture of the proposed method. \textbf{a}: Architecture of the generator. S2D stands for space-to-depth operation and D2S for depth to space. \textbf{b}: Architecture of the temporal discriminator (top left), spatial discriminator (middle left), and latent conditioning stack (bottom left) of the generator. On the right hand side are the architectures of the \textbf{G} block (top), \textbf{D} and \textbf{3D} block (middle), and \textbf{L} Block (right). For all panels, blocks with ($\uparrow$) or ($\downarrow$) indicate that they perform spatial up-sampling or down-sampling, respectively.}
    \label{fig:gan-architecture-generator}
\end{figure}

\paragraph{Architectural Details.}
The generator consists of two main modules: the conditioning stack, which creates a conditioning representation from previous radar observations; and the sampler, which generates 18 predictions of future radar from the conditioning representation. As the sizes of inputs to, and predictions from, the model are different during training than evaluation, we first describe the model during training before discussing modifications for evaluation. These descriptions are accompanied by schematic description in Extended Data Figure \ref{fig:gan-architecture-generator}. 

The \textit{conditioning stack} is a feed-forward convolutional neural network (Extended Data Figure \ref{fig:gan-architecture-generator}a) that generates the conditioning representation from four radar observations. First, each $256\times256\times1$ radar observation is converted to a $128\times128\times4$ input by stacking $2\times2$ patches into the channel layer (space-to-depth). Then, each radar observation is processed separately using four downsampling residual blocks (D Block in Extended Data Figure \ref{fig:gan-architecture-generator}b), which decreases the resolution and increases the number of channels by a factor of two. The four outputs of each residual block are concatenated across the channel dimension, and, for each output, a $3\times3$ spectrally normalized convolution is applied to reduce the number of channels by a factor of two, followed by a rectified linear unit. This yields a %
stack of conditioning representations of sizes $64\times64\times48$, $32\times32\times96$, $16\times16\times192$, and $8\times8\times384$.

The \textit{sampler} (Extended Data Figure \ref{fig:gan-architecture-generator}a), which is a stack of four ConvGRU units, uses the conditioning representations as initial states for each of its recurrent modules.  
Along with the initial states, $18$ copies (one for each lead time) of an $8\times8\times768$ latent representation are given as input to the lowest-resolution ConvGRU block. This representation is generated by the latent conditioning stack, a small feed-forward convolutional network that converts an $8\times8\times8$ input to the latent representation by gradually increasing the number of channels. %

For the \textit{latent conditioning stack},
entries in the $8\times8\times8$ input are independent draws from a normal distribution. The first two dimensions of the input are height and width, which are 1/32 of the height and width of the $256\times256\times1$ radar observations.
The latent conditioning stack comprises one $3\times3$ convolution, three L Blocks, a spatial attention module \cite{vaswani2017attention, zhang2019self}, and one L Block. The L Block is a modified residual block designed specifically for increasing the number of channels of its respective input.

As described, the output of the latent conditioning stack is repeated $18$ times and is used as input to the lowest resolution ConvGRU.
The output of each ConvGRU is then upsampled to an input of the next ConvGRU with one spectrally normalized convolution and two residual blocks that process all $18$ temporal representations independently. The second residual block doubles the input's spatial resolution with nearest neighbor interpolation, and halves its number of channels. After the last ConvGRU, the intermediate feature vector is of size $128\times128\times48$. After batch normalization, a ReLU and a $1\times1$ spectrally normalized convolution is applied, yielding an output of size $128\times128\times4$. Similar to super-resolution, this is converted to $18$ predictions of size $256\times256\times1$ with a depth-to-space operation.

The spatial and temporal discriminators used to train the generative method are similar to \cite{luc2020transformation}, and either operate on predictions (for generator steps), or predictions and targets (for discriminator steps).
The \textit{spatial discriminator} picks uniformly at random $8$ out of $18$ lead times, which are first downsampled to $128\times128\times1$ using $2\times2$ mean pooling, and then converted to a $64\times64\times4$ input by stacking $2\times2$ patches into the channel layer (space-to-depth). This is followed by five residual blocks (D Block), each of which halve the resolution while doubling the number of channels. The first D Block does not apply a ReLU before the first $3\times3$ convolution. The outputs of the five blocks are $32\times32\times48$, $16\times16\times96$, $8\times8\times192$, $4\times4\times384$, and $2\times2\times768$, respectively. After being processed by one more D Block that preserves the spatial resolution and number of channels, the representations are sum-pooled along the height and width dimensions. The $8$ resulting representations are inputs to a spectrally normalized linear layer, which are then summed together before a ReLU is applied for binary classification output. 
The input to the \textit{temporal discriminator} is a sequence comprised of the four contextual radar frames concatenated along the time axis to either the predictions (for generator steps), or the predictions or targets (for discriminator steps). A random crop of height and width $128\times128$ is extracted from the sequence, and each frame in the sequence is then converted to $64\times64\times4$ using a space-to-depth operation. This output is processed by two 3D Blocks, which mimic the processing of the first two D Blocks in the spatial discriminator, but with $3\times3\times3$ spectrally normalized convolutions. The first 3D Block does not apply a ReLU before the first $3\times3\times3$ convolution. Each frame of the resulting length-five $16\times16\times96$ representation is processed by four residual blocks with same architecture as that of the spatial discriminator. The remaining steps are identical to those after the last D Block of the spatial discriminator.

During model development, we initially found that including a batch normalization layer (without variance scaling) prior to the linear layer of the two discriminators improved training stability. The results presented use batch normalization, but we later were able to obtain nearly identical quantitative and qualitative results without it, and it is likely unneeded to reproduce the model.

\paragraph{Objective Function.}
The generator is trained with losses from the two discriminators and a grid cell regularization term (denoted $\mathcal{L}_R(\theta)$). The spatial discriminator $D_\phi$ has parameters $\phi$, the temporal discriminator $T_\psi$ has parameters $\psi$, and the generator $G_\theta$ has parameters $\theta$. We indicate the concatenation of two fields using the notation $\{X;G\}$. The generator loss that is maximized is
\begin{eqnarray}
\mathcal{L}_G(\theta) & = & \mathbb{E}_{X_{1:M+N}}\bigl[\mathbb{E}_{Z}[D(G_{\theta}(Z; X_{1:M})) + T(\{X_{1:M}; G_{\theta}(Z; X_{1:M})\})] - \lambda \mathcal{L}_R(\theta)\bigr];
\label{eq:gen_loss}
\end{eqnarray}
\begin{equation}
\mathcal{L}_R(\theta) = \tfrac{1}{HWN} \left\| \bigl(\mathbb{E}_Z[G_{\theta}(Z; X_{1:M})] - X_{M+1:M+N}\bigl) \odot w(X_{M+1:M+N}) \,\right\|_1.
\label{eq:xcorr_loss}
\end{equation}

We use Monte Carlo estimates for expectations over the latent Z in \eqref{eq:gen_loss} and \eqref{eq:xcorr_loss}. These are calculated using six samples per input $X_{1:M}$, which comprises $M=4$ radar observations. The grid cell regularizer ensures that the mean prediction remains close to the ground truth, and is averaged across all grid cells along the height $H$, width $W$, and lead-time $N$ axes. It is weighted towards higher rainfall targets using the function $w(y) = \max(y+1, 24)$, which operate element-wise for input vectors, and is clipped at $24$ for robustness to spuriously large values in the radar. The GAN spatial discriminator loss $\mathcal{L}_D(\phi)$ and temporal discriminator loss $\mathcal{L}_T(\psi)$ are minimized with respect to parameters $\phi$ and $\psi$, respectively. The discriminator losses uses a hinge loss formulation \cite{clark2019efficient}:
\begin{eqnarray}
\mathcal{L}_D(\phi) & = & \mathbb{E}_{X_{1:M+N},Z}[ReLU(1-D_{\phi}(X_{M+1:M+N})) + ReLU(1+D_{\phi}(G(Z; X_{1:M})))]. \\
\mathcal{L}_T(\psi) & = & \mathbb{E}_{X_{1:M+N},Z}[ReLU(1-T_{\psi}(X_{1:M+N})) + ReLU(1+T_{\psi}(\{X_{1:M}; G(Z; X_{1:M})\}))].
\label{eq:disc_loss}
\end{eqnarray}

During evaluation, the generator architecture is the same, but unless otherwise noted, full radar observations of size $1536\times1280$, and latent variables with height and width 1/32 of the radar observation size ($48\times40\times8$ of independent draws from a normal distribution), are used as inputs to the conditioning stack and latent conditioning stack, respectively. In particular, the latent conditioning stack allows for spatiotemporally consistent predictions for much larger regions than on which the generator is trained. 
 
The model is trained for $5 \times 10^5$ generator steps, with two discriminator steps per generator step. The learning rate for the generator is $5 \times 10^{-5}$, and is $2 \times 10^{-4}$ for the discriminator and uses Adam optimizer \cite{kingma2015adam} with $\beta_1=0.0$ and $\beta_2=0.999$. The scaling parameter for the grid cell regularization is set to $\lambda=20$, as this produced the best continuous ranked probability score results on the validation set. We train on 16 Tensor Processing Units cores\footnote{see \url{https://cloud.google.com/tpu/docs/tpus}} for 1 week on random crops of the training dataset of size $256\times256$ measurements using a batch-size of 16 per training step. The supplement contains additional comparisons showing the contributions of the different loss components to overall performance. We evaluated the speed of sampling by comparing speed on both CPU (10 core AMD EPYC) and GPU (NVIDIA V100) hardware. We generate 10 samples and report the median time: for CPU the median time per sample was \unit[25.7]{s}, and \unit[1.3]{s} for the GPU.

\subsection*{UNet Baseline}
\label{sub:models-unet}
We use a UNet encoder-decoder model as strong baseline similarly to how it was used in related studies \cite{ayzel2020rainnet, agrawal2019machine}, but we make architectural and loss function changes that improve its performance at longer lead times and higher precipitation. First, we replace all convolutional layers with residual blocks, as the latter provided a small but consistent improvement across all prediction thresholds. Second, rather than predicting only a single output and using autoregressive sampling during evaluation, the model predicts all frames in a single forward pass. This somewhat mitigates the excessive blurring found in \cite{ayzel2020rainnet} and improves results on quantitative evaluation. Our architecture consists of six residual blocks, where each block doubles the number of channels of the latent representation followed by spatial down-sampling by factor of 2. The representation with the highest resolution has 32 channels which increases up to 1024 channels.

Similarly to \cite{shi2017deep}, we use a loss weighted by precipitation intensity.
Rather than weighting by precipitation bins, however, we reweight the loss directly by the precipitation to improve results on thresholds outside of the bins specified by \cite{shi2017deep}. Additionally, we truncate the maximum weight to \unit[24]{mm/hr} as an error in reflectivity of observations leads to larger error in the precipitation values. We also found that including a mean squared error loss made predictions more sensitive to radar artifacts; as a result, the model is only trained with precipitation weighted mean average error loss.

The model is trained with batch size eight for $1 \times 10^{6}$ steps, with learning rate $2 \times 10^{-4}$ with weight decay, 
using the Adam Optimizer with default exponential rates.
We select a model using early stopping on the average area under the Precision-Recall curve on the validation set.
The UNet baselines are trained with 4 frames of size $256 \times 256$ as context.

\subsection*{Axial Attention Baseline}
\label{sub:models-axial}
As a second strong deep learning-based baseline, we adapt the MetNet model~\cite{sonderby2020metnet}, which is a combination of a Convolutional LSTM encoder~\cite{xingjian2015convolutional} and an Axial Attention decoder~\cite{ho2019axial}, for radar-only nowcasting. MetNet was demonstrated to achieve strong results on short-term (up to 8 hours) low precipitation forecasting using radar and satellite data of the continental US, making pointwise probabilistic predictions and by factorizing spatial dependencies using alternating layers of Axial Attention.

We modified that Axial Attention encoder decoder model to use only radar observations, as well as to cover the spatial and temporal extent of data in this study. We rescaled the targets of the model to improve its performance on forecasts of high precipitation events. After evaluation on our UK and US data, we observed that additional satellite or topographical data as well as the spatio-temporal embeddings did not provide statistically significant CSI improvement. An extended description of the model and its adaptations is provided in Supplement \ref{sec:suppl_baselines}.

\subsection*{PySTEPS Baseline}
We use the PySTEPS implementation from \cite{pulkkinen2019pysteps} using the default configuration available at \url{https://github.com/pySTEPS/pysteps}. Please refer to \cite{pulkkinen2019pysteps, bowler2006steps} for more details of this approach.
In our evaluation, unlike other models evaluated that use inputs of size 256 $\times$ 256, PySTEPS is given the advantage of being fed inputs of size 512 $\times$ 512, which was found to improve its performance.

\section*{Evaluation}
\label{sec:evaluation}
We evaluate our model and baselines using commonly-used quantitative verification measures, as well as qualitatively using a cognitive assessment task with expert forecasters. 
Unless otherwise noted, models are trained on years 2016-2018 and evaluated on 2019 (i.e. a yearly split).

\subsection*{Expert Forecaster Study}

The expert forecaster study described is a two phase protocol consisting of a ranked comparison task followed by a retrospective recall interview. The study was submitted for ethical assessment by an independent ethics committee and received favourable review. Key elements of the protocol involved consent forms that clearly explained the task and time commitment, clear messaging on the ability to withdraw from the study at any point, and that the study was not an assessment of the forecaster's skills and would not impact their employment and role in any way. Forecasters were not paid for participation, since involvement in these types of studies is considered part of the broader role of the forecaster. The study was anonymized, and only the study lead had access to the assignment of experimental IDs. The study was restricted to forecasters in guidance-related roles, i.e., forecasters whose role is to interpret weather forecasts, synthesize forecasts, and provide interpretations, warnings and watches. \numforecasters~forecasters agreed to participate in the study. 

\begin{figure}[tp]
    \centering
    \begin{minipage}{0.5\textwidth}
    \textbf{a}\\
   \includegraphics[width=.8\textwidth]{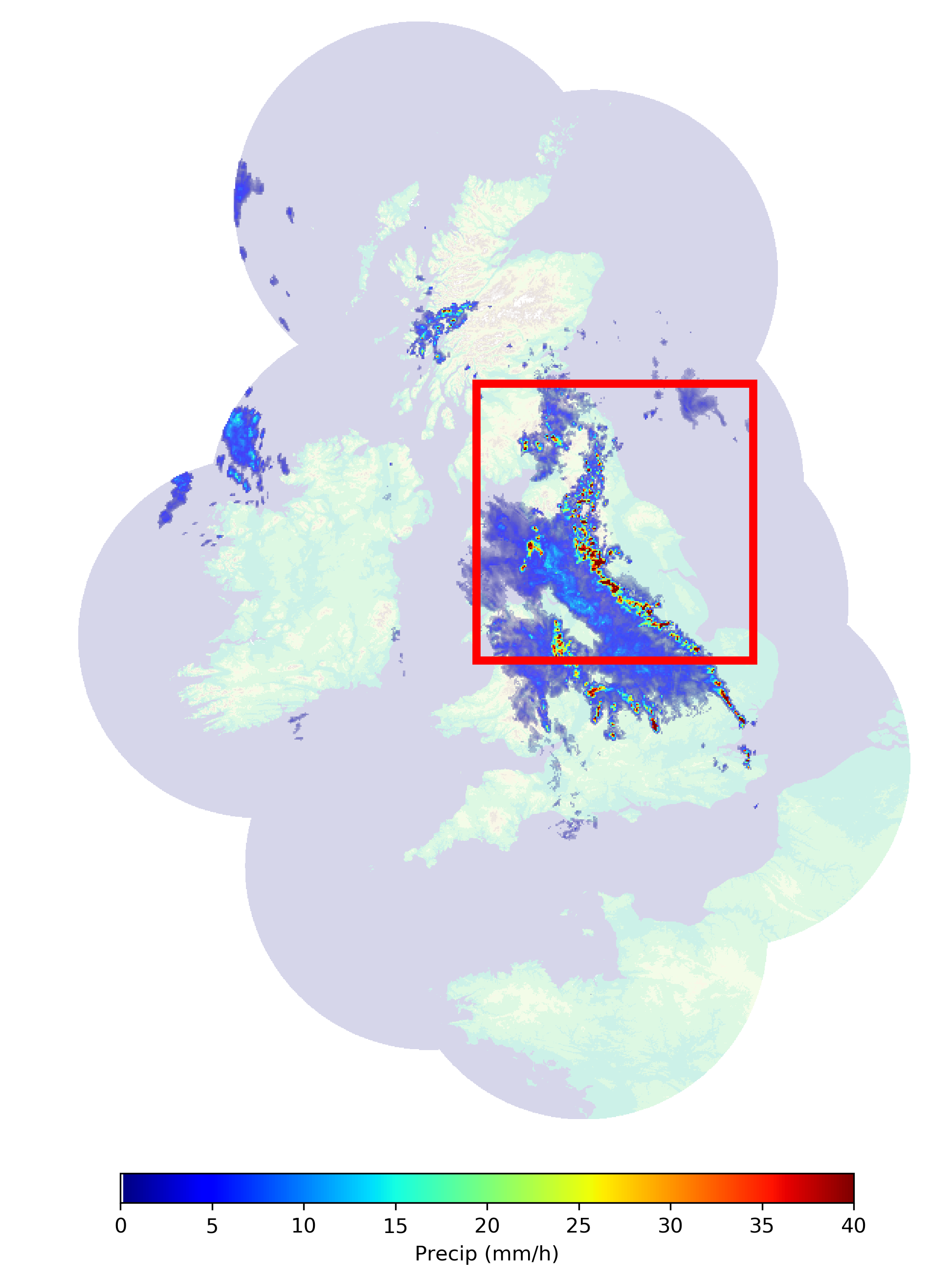}\\
    \textbf{c}\\
    \includegraphics[width=.95\textwidth]{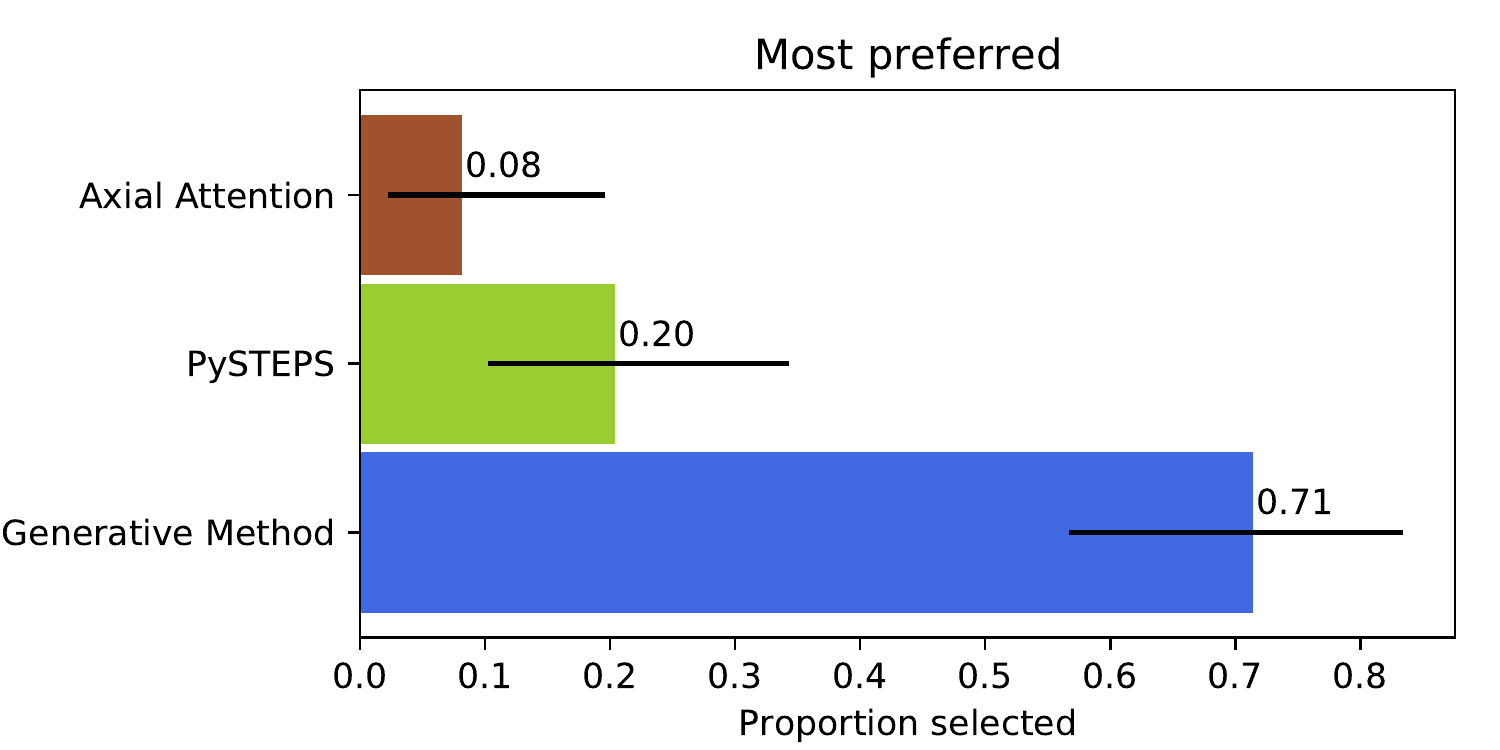}\\
    
    \end{minipage}
    \begin{minipage}{0.49\textwidth}
    \textbf{b}\\
    \newcommand{\insertevent}[1]{\includegraphics[width=0.32\textwidth]{generated_qualitative_figures_20210211/event_b_white/#1}}
    \setlength{\tabcolsep}{0pt}
    \begin{tabular}{m{1em} c c c}
        & T+\unit[30]{min} &T+\unit[60]{min} & T+\unit[90]{min} \\ 
        \vspace{-25mm}
        \rotatebox{90}{\hspace{0.8em}Observations} &
            \insertevent{steps_target_30.png} &
            \insertevent{steps_target_60.png} &
            \insertevent{steps_target_90.png}
            \\
            \vspace{-25mm}
            \rotatebox{90}{\hspace{0.7em}Gen. Method} &
            \insertevent{ganspd256_pred_samples_30.png} &
            \insertevent{ganspd256_pred_samples_60.png} &
            \insertevent{ganspd256_pred_samples_90.png}
            \\
        \vspace{-25mm}
        \rotatebox{90}{\hspace{1.7em}PySTEPS} &
            \insertevent{steps_pred_samples_30.png} &
            \insertevent{steps_pred_samples_60.png} &
            \insertevent{steps_pred_samples_90.png}
            \\
        \vspace{-25mm}
        \rotatebox{90}{\hspace{2.5em}UNet} &
            \insertevent{unet_pred_30.png} &
            \insertevent{unet_pred_60.png} &
            \insertevent{unet_pred_90.png}
            \\
        \vspace{-25mm}
        \rotatebox{90}{Axial Att. Mode} &
            \insertevent{metnet_pred_30.png} &
			\insertevent{metnet_pred_60.png} &
			\insertevent{metnet_pred_90.png}
            \\
        \vspace{-25mm}
        \rotatebox{90}{Axial Att. Smpl} &
            \insertevent{metnet_samples_pred_samples_30.png} &
            \insertevent{metnet_samples_pred_samples_60.png} &
            \insertevent{metnet_samples_pred_samples_90.png}
            \\
    \end{tabular}
    \end{minipage}
    \caption{\textbf{Case study of performance on a challenging precipitation event starting at 2019-07-24 at 03:15 UK, showing two separate banded structures of intense rainfall in the north-east and south-west over northern England. The Generative Method is better able to predict the spatial coverage and convection compared to other methods over a longer time period, while not over-estimating the intensities, and is significantly preferred by forecasters (71\% first choice, $N=\numforecasters$, $p<2\times10^{-4}$).} 
    	\textbf{a:} Geographic context for predictions. \textbf{b:} A single prediction at T+\unit[30], T+\unit[60], and T+\unit[90]{mins} lead time for different models. Critical Success Index (CSI) at thresholds \unit[2]{mm/hr} and \unit[8]{mm/hr} and Continuous Ranked Probability Score (CRPS) for an ensemble of 4 samples shown in a bottom left corner. For Axial Attention we show the mode prediction and the single sample. \textbf{c:} Expert forecaster preference for the visualized prediction (Axial Attention uses the mode prediction; we report the percentage of forecasters for their first choice rating as well as the Clopper-Pearson 95\% confidence interval).}
    
    \label{fig:model-visualization-b}
\end{figure}

\begin{figure}[tp]
    \centering
    \begin{minipage}{0.5\textwidth}
    \textbf{a}\\
    \includegraphics[width=.8\textwidth]{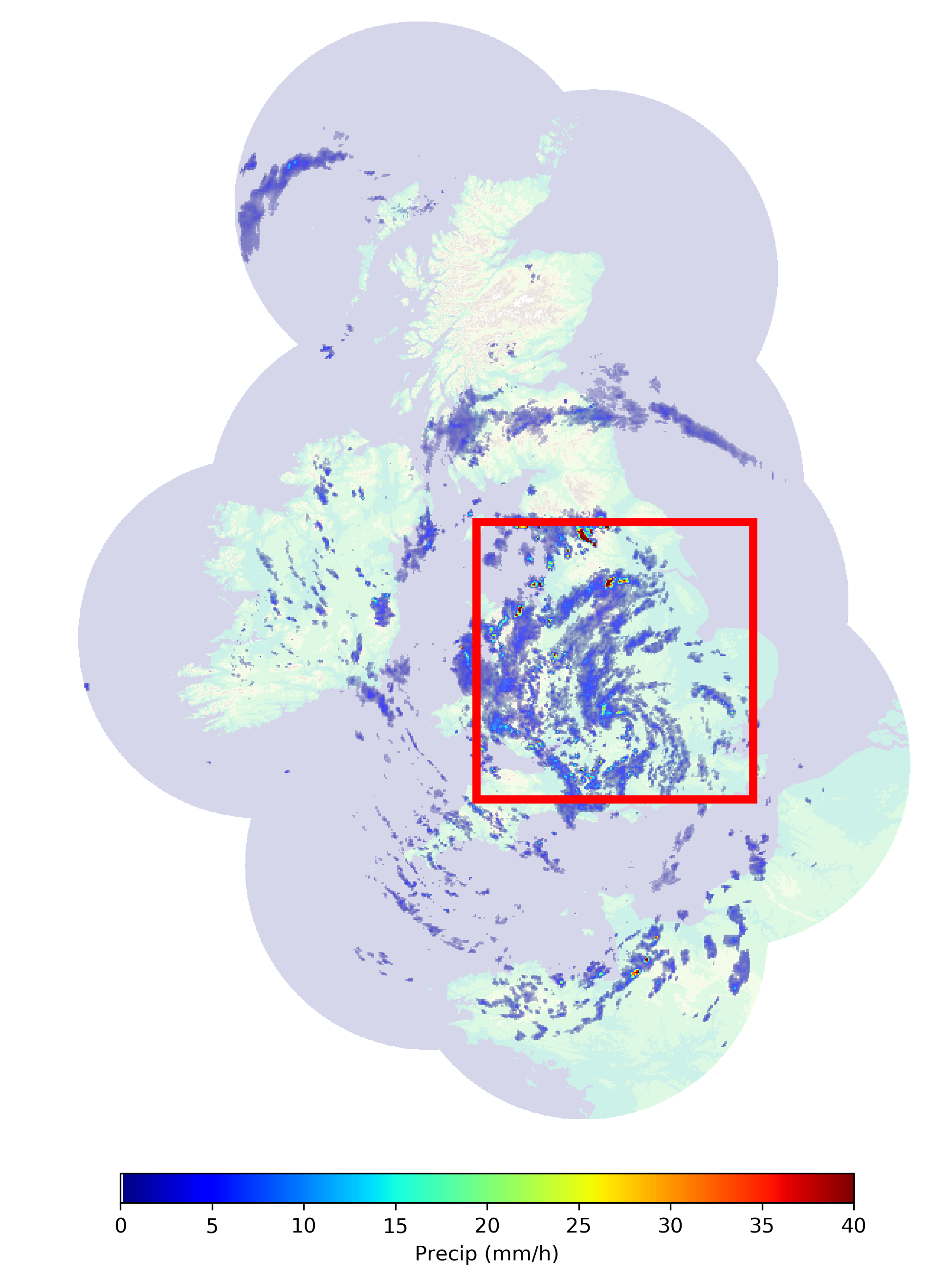}\\
    \textbf{c}\\
    \includegraphics[width=.95\textwidth]{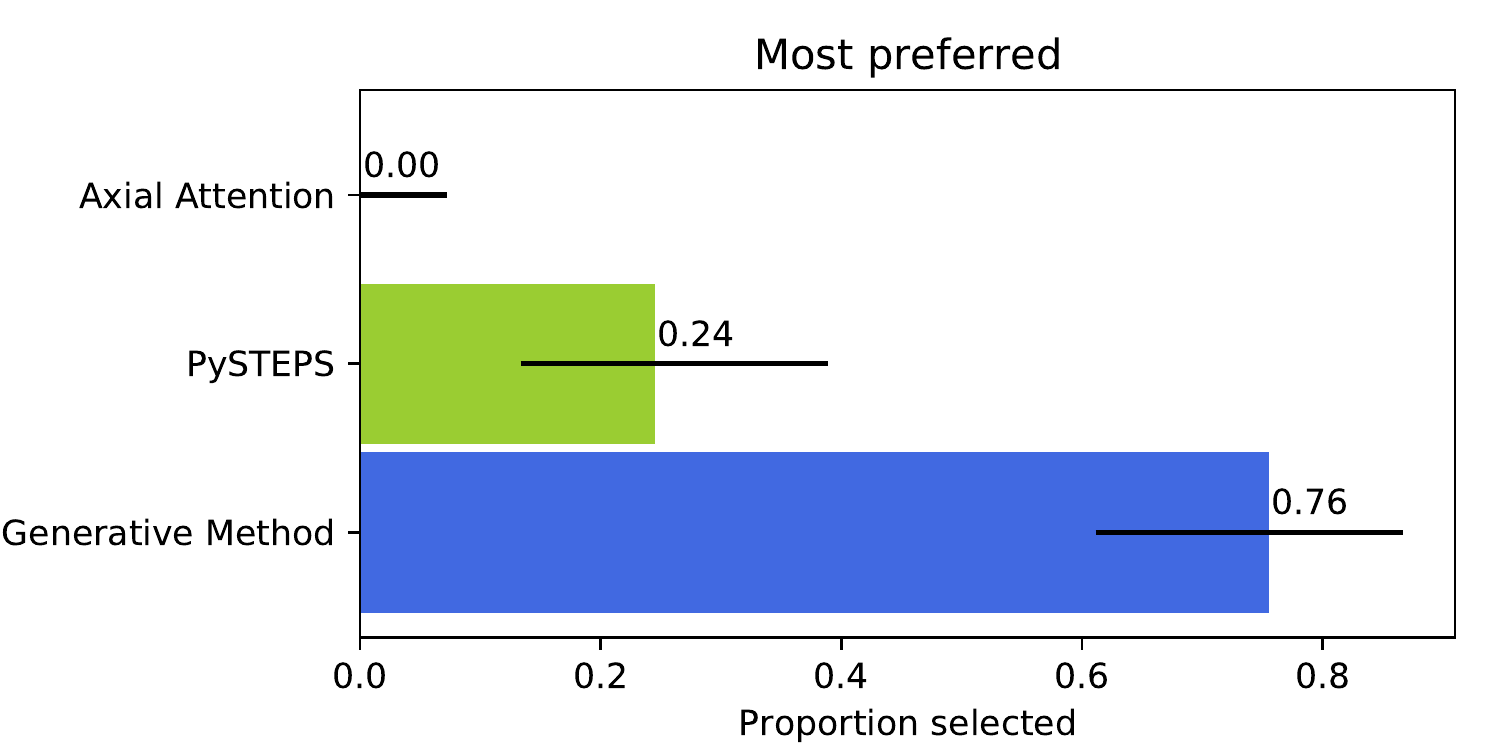}\\
    \end{minipage}
    \begin{minipage}{0.49\textwidth}
    \textbf{b}\\
    \newcommand{\insertevent}[1]{\includegraphics[width=0.32\textwidth]{generated_qualitative_figures_20210211/event_c_white/#1}}
    \newcommand{\insertoldevent}[1]{\includegraphics[width=0.32\textwidth]{qualitative_samples_c_white/#1}}
    \setlength{\tabcolsep}{0pt}
    \begin{tabular}{m{1em} c c c}
        & T+\unit[30]{min} &T+\unit[60]{min} & T+\unit[90]{min} \\ 
        \vspace{-25mm}
        \rotatebox{90}{\hspace{0.8em}Observations} &
            \insertevent{steps_target_30.png} &
            \insertevent{steps_target_60.png} &
            \insertevent{steps_target_90.png}
            \\
            \vspace{-25mm}
             \rotatebox{90}{\hspace{0.7em}Gen. Method} &
            \insertevent{ganspd256_pred_samples_30.png} &
            \insertevent{ganspd256_pred_samples_60.png} &
            \insertevent{ganspd256_pred_samples_90.png}
            \\
        \vspace{-25mm}
        \rotatebox{90}{\hspace{1.7em}PySTEPS} &
            \insertevent{steps_pred_samples_30.png} &
            \insertevent{steps_pred_samples_60.png} &
            \insertevent{steps_pred_samples_90.png}
            \\
        \vspace{-25mm}
        \rotatebox{90}{\hspace{2.5em}UNet} &
            \insertevent{unet_pred_30.png} &
            \insertevent{unet_pred_60.png} &
            \insertevent{unet_pred_90.png}
            \\
        \vspace{-25mm}
        \rotatebox{90}{Axial Att. Mode} &
            \insertevent{metnet_pred_30.png} &
            \insertevent{metnet_pred_60.png} &
            \insertevent{metnet_pred_90.png}
            \\
        \vspace{-25mm}
        \rotatebox{90}{Axial Att. Smpl} &
            \insertevent{metnet_samples_pred_samples_30.png} &
            \insertevent{metnet_samples_pred_samples_60.png} &
            \insertevent{metnet_samples_pred_samples_90.png}
            \\
    \end{tabular}
    \end{minipage} \\
    \caption{ 
    	\textbf{Case study of performance on a challenging precipitation event starting 2019-07-30 at 15:15 UK, showing a pattern of precipitation around a low pressure area which is slow moving, resulting in the cyclonic banded structures over England. The Generative Method captures extent of precipitation overall over the area, though slightly overdoes rain coverage between bands, and is significantly preferred by forecasters (76\% first choice, $N=\numforecasters$, $p<2\times10^{-4}$).} 
    	   \textbf{a:} Geographic context for predictions. \textbf{b:} A single prediction at T+\unit[30], T+\unit[60], and T+\unit[90]{mins} lead time for different models. Critical Success Index (CSI) at thresholds \unit[2]{mm/hr} and \unit[8]{mm/hr} and Continuous Ranked Probability Score (CRPS) for an ensemble of 4 samples shown in a bottom left corner. For Axial Attention we show the mode prediction and the single sample. \textbf{c:} Expert forecaster preference for the visualized prediction (Axial Attention uses the mode prediction; we report the percentage of forecasters for their first choice rating as well as the Clopper-Pearson 95\% confidence interval).}

    \label{fig:model-visualization-c}
\end{figure}

Phase 1 of the study, the rating assessment, involved each forecaster receiving a unique form as part of their experimental evaluation. The Axial Attention mode prediction is used in the assessment, and this was selected as the most appropriate prediction during the pilot assessment of the protocol by the Chief Forecaster. The phase 1 evaluation comprised an initial practice phase of three judgments for forecasters to understand how to use the form and assign ratings, followed by an experimental phase that involved 20 trials that were different for every forecaster, and a final case study phase in which all forecasters rated the same three scenarios (the scenarios in panel 1a, and Extended Data \ref{fig:model-visualization-b} and \ref{fig:model-visualization-c}); these three events were chosen by the Chief Forecaster---who is independent of the research team and also did not take part in the study---as difficult events that would expose challenges for the nowcasting approaches we compare. Ten forecasters  participated in the subsequent retrospective recall interview. This interview involved an in-person interview in which experts were asked to explain the reasoning for their assigned rating and what aspects informed their decision-making. These interviews all used the same script for consistency, and these sessions were recorded with audio only. Once all the audio was transcribed, the recordings were deleted.

The twenty trials of the experimental phase were split into two parts, each containing ten trials. The first 10 trials comprised medium rain events (rainfall greater than \unit[5]{mm/hr}) and the second 10 trials comprised heavy rain events (rainfall greater than \unit[10]{mm/hr}). 141 days from 2019 were chosen by the Chief Forecaster as having high-precipitation events. From these dates, radar fields were chosen algorithmically according to the following procedure.
First, we excluded from the crop selection procedure the \unit[192]{km} that forms the image margins of each side of the radar field. 
Then, the crop over \unit[256]{km} regions, containing the maximum fraction of grid cells above the given threshold, $5$ or \unit[10]{mm/hr}, was selected from the radar image. 
If there was no precipitation in the frame above the given threshold, the selected crop was the one with the maximum average intensity.

 Extended Data \ref{fig:model-visualization-b} shows a high-precipitation front with decay and Extended Data \ref{fig:model-visualization-c} shows a cyclonic circulation event (low-pressure area), both of which are difficult for current deep learning models to predict. These two cases were also assessed by all expert forecasters as part of the evaluative study, and in both cases, forecasters significantly preferred the generative approach ($N=\numforecasters$, $p<10^{-4}$) to competing methods.
For the high-precipitation front in Extended Data \ref{fig:model-visualization-b}, forecasters ranked first the generative approach in 71\% of cases. Forecasters reported that \modelname~has `decent accuracy with both the shape and intensity of the feature', `but loses most of the signal for embedded convection by T+90'. PySTEPS is `too extensive with convective cells and lacks the organisation seen in the observations', and the axial attention model being able to `highlighting the worst areas' but `looks wrong'. 

For the cyclonic circulation in Extended Data \ref{fig:model-visualization-c}, forecasters ranked first the generative approach in 76\% of cases. Forecasters reported that it was difficult to judge this case between \modelname~and PySTEPS. When making their judgment, they chose \modelname~since it has `best fit and rates overall'. \modelname~`captures extent of precipitation overall [in the] area, though slightly overdoes rain coverage between bands', whereas PySTEPS `looks less spatially accurate as time goes on'. The Axial Attention model `highlights the area of heaviest rain although its structure is unrealistic and too binary'.

\subsection*{Quantitative Evaluation}
We evaluate all models using established metrics \citep{jolliffe2012forecast}: Critical Success Index (CSI), Continuous Ranked Probability Score (CRPS), Pearson Correlation Coefficient (PCC), Fractions Skill Score (FSS) \cite{roberts2008scale,schwartz2010toward}, and radially-averaged Power Spectral Density (PSD). These are described in Supplement \ref{sec:suppl_detailed_metrics}.

To make evaluation computationally feasible, for all metrics except PSD, we evaluate the models on a subsampled test set, consisting of $512\times512$ crops drawn from the full radar images. We use an importance sampling scheme (described in Supplement \ref{sec:importance-sampling-scheme}) to ensure that this subsampling does not unduly compromise the statistical efficiency of our estimators of the evaluation metrics. The subsampling reduces the size of the test set to 66,851 and Supplement \ref{ssec:ssot_vs_ff} shows that results obtained when evaluating CSI are not different when using the dataset with or without subsampling.
All models except PySTEPS are given the center $256\times256$ crop as input. PySTEPS is given the entire $512\times512$ crop as input as this improves its performance. The predictions are evaluated on the center $64\times64$ grid cells, ensuring that models are not unfairly penalized by boundary effects.
Our statistical significance tests use every other week of data in the test set (leaving N=26 weeks) as independent units. We test the null hypothesis that performance metrics are equal for the two models, against the two-sided alternative, using a paired permutation test \citep{edgington2007randomization} with $10^6$ permutations.

Extended Data \ref{fig:uk_yearly_test_results2}, shows additional probabilistic metrics that measure the calibration of the evaluated methods. 
Extended Data \ref{fig:uk_results_rain_type} compares the performances of competing methods against different rain types. We annotated examples covering the entire United Kingdom during 2019 as belonging to several rain types. These examples were annotated by an author of the paper but not involved in model development. These annotations were then reviewed and approved by an independent forecaster at the Met Office. These precipitation-event types are: frontal, non-frontal, or mixed type precipitation; precipitation with coherent motion, incoherent motion, or mixed type; and scattered, non-scattered, or mixed scattered and non-scattered precipitation.
In nearly all cases, \modelname~outperformed competing methods on both CRPS and CSI metrics. \modelname~performs particularly well for non-frontal rain (panels a,b), which is an important result as non-frontal rain is known to be very difficult to predict \citep{sun2014use}.

Extended Data \ref{fig:uk_yearly_test_results_nwp_preliminary} compares the performance to that of an NWP, using the UKV deterministic forecast \cite{bush2020first}, showing that NWPs are not competitive in this regime. See the supplement for further details of the NWP configuration.

To verify other generalization characteristics of our approach---as an alternative to the yearly data split that uses training data of 2016-2018 and tests on 2019---we also use a weekly split: where the training, validation, and test sets comprise Thursday through Monday, Tuesday, and Wednesday, respectively. The sizes of the training and test datasets are 1.48M and 36,106, respectively. Extended Data \ref{fig:uk_weekly_test_results} shows the same competitive verification performance of \modelname~in this generalization test. 

To further assess the generalization of our method, we evaluate on a second data set from the United States using the Multi-Radar Multi-Sensitivity (MRMS) data set, which consists of radar composites for years 2017-2019 \cite{smith2016multi}. We use 2 years for training and 1 year for testing, and even with this more limited data source, our model still obtains comparable results relative to the other baselines. Extended Data \ref{fig:us_yearly_test_results}, \ref{fig:us_yearly_psd_by_region}, and \ref{fig:us_yearly_test_results2} compares all methods on all metrics we have described, showing both the generalization and skillful performance on this second data set. The Supplement contains additional comparisons on performance with different initializations and performance of different loss function components. 

\paragraph{Data Availability.}
Processed radar data over the UK used is to be released under open source licence. 
For the raw data, other licences, and alternative time periods, the data from the UK can be obtained with appropriate agreements from the Met Office; see \url{https://www.metoffice.gov.uk/research/weather/observations-research/radar-products}. The multi-radar multi-sensor (MRMS) dataset is available with appropriate agreements from NOAA; see \url{https://www.nssl.noaa.gov/projects/mrms/}.

\paragraph{Author Affiliations.}
$^1$DeepMind; $^2$Met Office; $^3$University of Exeter; $^4$University of Reading; $\dagger$Alberto Arribas contributed to this research while at the Met Office, and now at Microsoft.


\begin{figure}[tp]
    \centering
    \newcommand{\insertfigure}[3]{\includegraphics[width=#1\textwidth]{figures/figures_UK/yearly/single_run/#2/#3}}
    \begin{minipage}{0.8\textwidth}
        \begin{minipage}{0.99\textwidth}
        \textbf{a}\\
        \insertfigure{0.99}{20210210}{prob-bysample-uk512-yearly-test-1-4-8fss.png}
        \end{minipage} \\
        \begin{minipage}{0.99\textwidth}
        \textbf{b}\\
        \insertfigure{0.99}{20210210}{prob-csicc-uk512-yearly-test-1-4-8cross_correlation.png}
        \end{minipage} \\
        \begin{minipage}{0.57\textwidth}
        \textbf{c} \\
        \insertfigure{1}{20210210}{prob-csicc-uk512-yearly-test-1-4-8reliability.png}
        \end{minipage}
        \begin{minipage}{0.42\textwidth}
        \textbf{d} \\
        \insertfigure{1}{20210210}{prob-csicc-uk512-yearly-test-1-4-8rank-histogram.png}
        \end{minipage} 
    \end{minipage}
    
    \caption{ \textbf{Further verification scores for the United Kingdom in 2019.}
    \textbf{a:} Fractions Skill Score across 20 samples of different models across spatial scale thresholds \unit[1]{mm/hr} (left), \unit[4]{mm/hr}, \unit[8]{mm/hr} (right).
    \textbf{b:} Pearson Correlation Coefficient of various models for original predictions (left), average rain rate over a \unit[4]{km} $\times$ \unit[4]{km} catchment area (middle), and average rain rate over a \unit[16]{km} $\times$ \unit[16]{km} catchment area (right).
     \textbf{c:} Reliability diagrams and sharpness plots for two precipitation thresholds for T+\unit[60]{min} predictions.
     \textbf{d:} Rank Histogram at T+\unit[60]{min}.}
    \label{fig:uk_yearly_test_results2}
\end{figure}

\begin{figure}[tp]
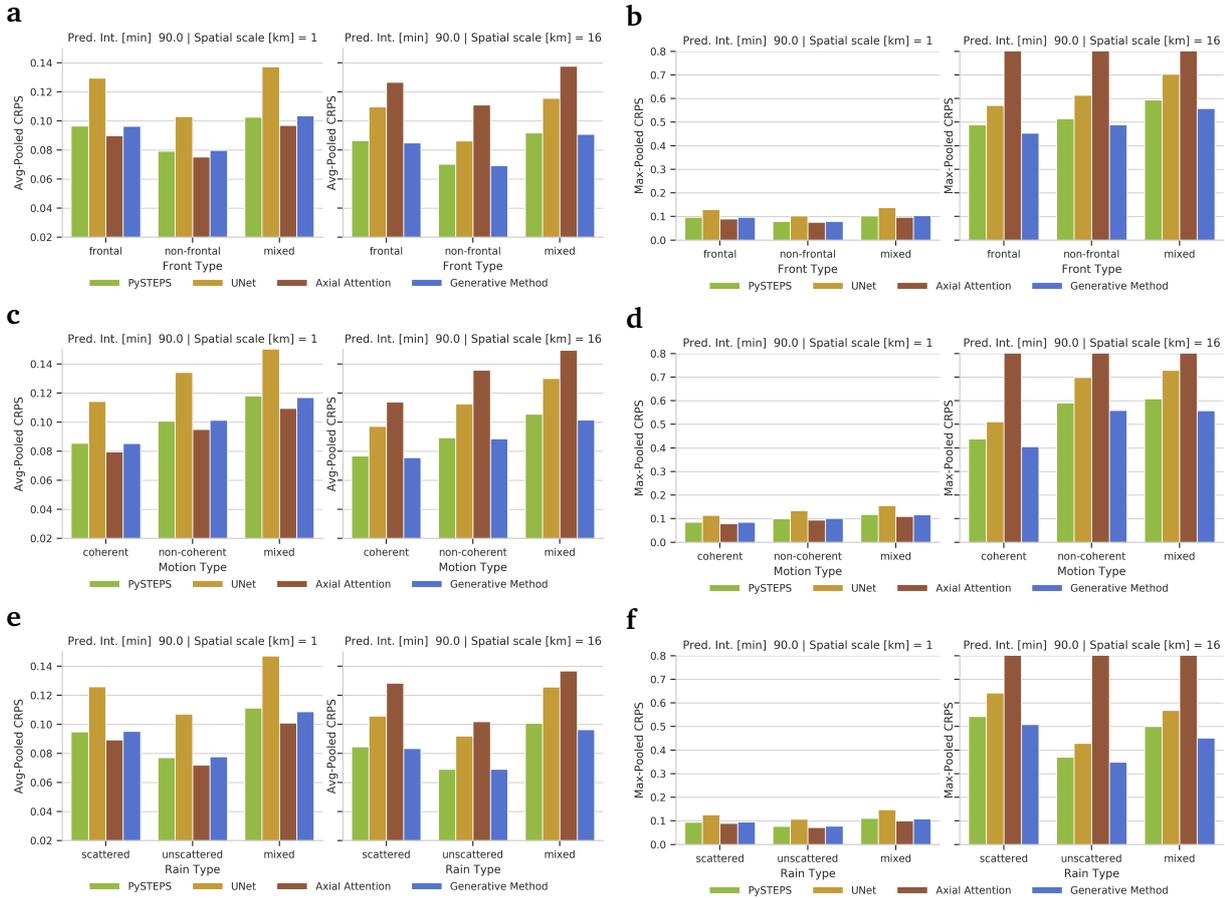

    \newcommand{\insertfigure}[3]{\includegraphics[width=#1\textwidth]{figures/figures_UK/yearly/by_rain_type/#2/#3}}
    \begin{minipage}{0.99\textwidth}
        \begin{minipage}{0.49\textwidth}
        \textbf{a}\\
        \insertfigure{0.99}{20210210}{probbin-uk512-yearly-test-frontcrps-crps.png}
        \end{minipage}
        \begin{minipage}{0.49\textwidth}
        \textbf{b}\\
        \insertfigure{0.99}{20210210}{probbin-uk512-yearly-test-frontcrps_max-crps.png}
        \end{minipage}
    \end{minipage} \\
    \begin{minipage}{0.99\textwidth}
        \begin{minipage}{0.49\textwidth}
        \textbf{c}\\
        \insertfigure{0.99}{20210210}{probbin-uk512-yearly-test-motioncrps-crps.png}
        \end{minipage}
        \begin{minipage}{0.49\textwidth}
        \textbf{d}\\
        \insertfigure{0.99}{20210210}{probbin-uk512-yearly-test-motioncrps_max-crps.png}
        \end{minipage}
    \end{minipage} \\
    \begin{minipage}{0.99\textwidth}
        \begin{minipage}{0.49\textwidth}
        \textbf{e}\\
        \insertfigure{0.99}{20210210}{probbin-uk512-yearly-test-scatteredcrps-crps.png}
        \end{minipage}
        \begin{minipage}{0.49\textwidth}
        \textbf{f}\\
        \insertfigure{0.99}{20210210}{probbin-uk512-yearly-test-scatteredcrps_max-crps.png}
        \end{minipage} \\
    \end{minipage} \\
    \caption{\textbf{Rain-type analysis for predictions T+\unit[90]{min} lead time, comparing CRPS metrics at \unit[1]{km} and \unit[16]{km} spatial scales for the United Kingdom.} 
    \textbf{a:} Average pooled Continuous Ranked Probability Score across 20 samples of different models at scales \unit[1]{km} and \unit[16]{km}, for frontal, non-frontal, and mixed-type precipitation events. 
    \textbf{b:} Max-pooled CRPS across 20 samples of different models for frontal, non-frontal, and mixed-type precipitation events.
    \textbf{c:} Avg-pooled CRPS for coherent, non-coherent, and mixed-type precipitation events.
    \textbf{d:} Max-pooled CRPS for coherent, non-coherent, and mixed-type precipitation events.
    \textbf{e:} Avg-pooled CRPS for scattered, non-scattered, and mixed-type precipitation events.
    \textbf{f:} Max-pooled for scattered, non-scattered, and mixed-type precipitation events.
    }
    \label{fig:uk_results_rain_type}
    
\end{figure}

\begin{figure}[tp]
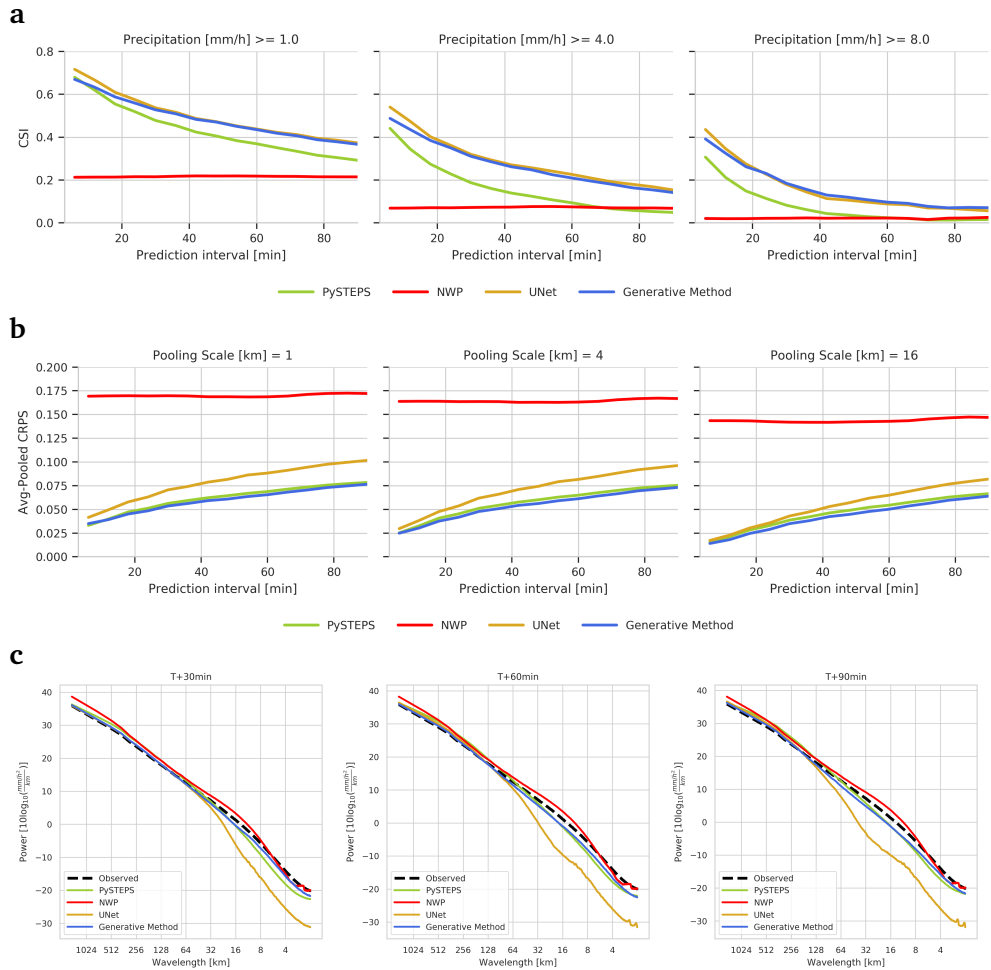

    \centering
    \newcommand{\insertfigure}[3]{\includegraphics[width=#1\textwidth]{figures/figures_UK/yearly_nwp/#2/#3}}
    \begin{minipage}{0.8\textwidth}
        \begin{minipage}{0.99\textwidth}
        \textbf{a}\\
        \insertfigure{0.99}{20210210}{prob-uk512-yearly-nwp-testclassifier-sample_mean-matching_thresholds-csi.png}
        \end{minipage} \\
        
        \begin{minipage}{0.99\textwidth}
        \textbf{b}\\
        \insertfigure{0.99}{20210210}{prob-uk512-yearly-nwp-testcrps-crps.png}
        \end{minipage} \\
        
        \begin{minipage}{0.99\textwidth}
        \textbf{c} \\
        \insertfigure{0.32}{20210210}{psd_nwp_nimrod_yearly_splits_30min.png}
        \insertfigure{0.32}{20210210}{psd_nwp_nimrod_yearly_splits_60min.png}
        \insertfigure{0.32}{20210210}{psd_nwp_nimrod_yearly_splits_90min.png}
        \end{minipage} \\
    \end{minipage}
    
    \caption{ \textbf{Verification scores for the United Kingdom by yearly splits aligned with NWP initialization times.}
    \textbf{a:} Critical Success Index across 20 samples of PySTEPS across precipitation thresholds \unit[1]{mm/hr} (left), \unit[4]{mm/hr}, \unit[8]{mm/hr} (right).
     \textbf{b:} CRPS of various models for original predictions (left), average rain rate over a \unit[4]{km} $\times$ \unit[4]{km} catchment area (middle), and average rain rate over a \unit[16]{km} $\times$ \unit[16]{km} catchment area (right). 
     \textbf{c:} Radially-averaged power spectral density for full-frame 2019 predictions for different models.
     Please note that these results are computed with significantly fewer examples of the UK yearly dataset due to the NWP lead time alignment. }
    \label{fig:uk_yearly_test_results_nwp_preliminary}
\end{figure}

\begin{figure}[tp]
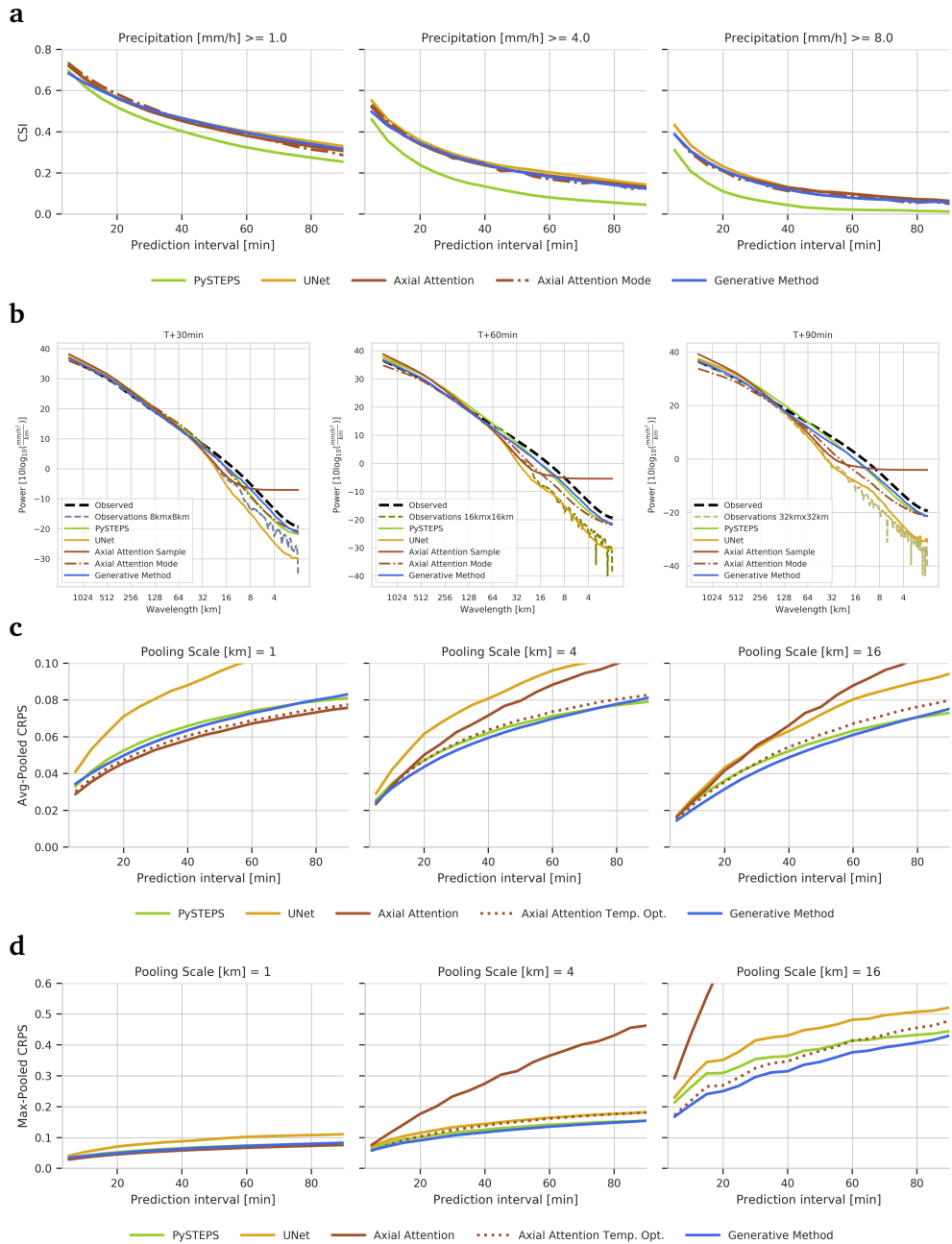

    \centering
    \newcommand{\insertfigure}[3]{\includegraphics[width=#1\textwidth]{figures/figures_UK/weekly/single_run/#2/#3}}
    \begin{minipage}{0.8\textwidth}
        \begin{minipage}{0.99\textwidth}
        \textbf{a}\\
        \insertfigure{0.99}{20210210}{prob-csicc-uk512-weekly-test-1-4-8classifier-sample_mean-matching_thresholds-csi.png}
        \end{minipage} \\
        \begin{minipage}{0.99\textwidth}
        \textbf{b} \\
        \insertfigure{0.32}{20210210}{psd_nimrod_weekly_splits_30min_8kmObs.png}
        \insertfigure{0.32}{20210210}{psd_nimrod_weekly_splits_60min_16kmObs.png}
        \insertfigure{0.32}{20210210}{psd_nimrod_weekly_splits_90min_32kmObs.png}
        \end{minipage} \\
        \begin{minipage}{0.99\textwidth}
        \textbf{c} \\
        \insertfigure{0.99}{20210210}{prob-crps-uk512-weekly-test-1-4-8crps-crps.png}
        \end{minipage} \\
        \begin{minipage}{0.99\textwidth}
        \textbf{d}\\
        \insertfigure{0.99}{20210210}{prob-crps-uk512-weekly-test-1-4-8crps_max-crps.png}
        \end{minipage} \\
    \end{minipage}
    
    \caption{ \textbf{Verification scores for the United Kingdom by weekly splits.}
    \textbf{a:} Critical Success Index across 20 samples of different models across precipitation thresholds \unit[1]{mm/hr} (left), \unit[4]{mm/hr}, \unit[8]{mm/hr} (right). We also report results for the Axial Attention mode prediction.
    \textbf{b:} Radially-averaged power spectral density for full-frame predictions for different models at T+\unit[30] (left), T+\unit[60] (middle), and T+\unit[90]{mins} (right). 
    \textbf{c:} CRPS of various models for original predictions (left), average rain rate over a \unit[4]{km} $\times$ \unit[4]{km} catchment area (middle), and average rain rate over a \unit[16]{km} $\times$ \unit[16]{km} catchment area (right).
    \textbf{d:} CRPS of various models for original predictions (left), maximum rain rate over a \unit[4]{km} $\times$ \unit[4]{km} catchment area (middle), and maximum rain rate over a \unit[16]{km} $\times$ \unit[16]{km} catchment area (right).  }
    \label{fig:uk_weekly_test_results}
\end{figure}

\begin{figure}[tp]
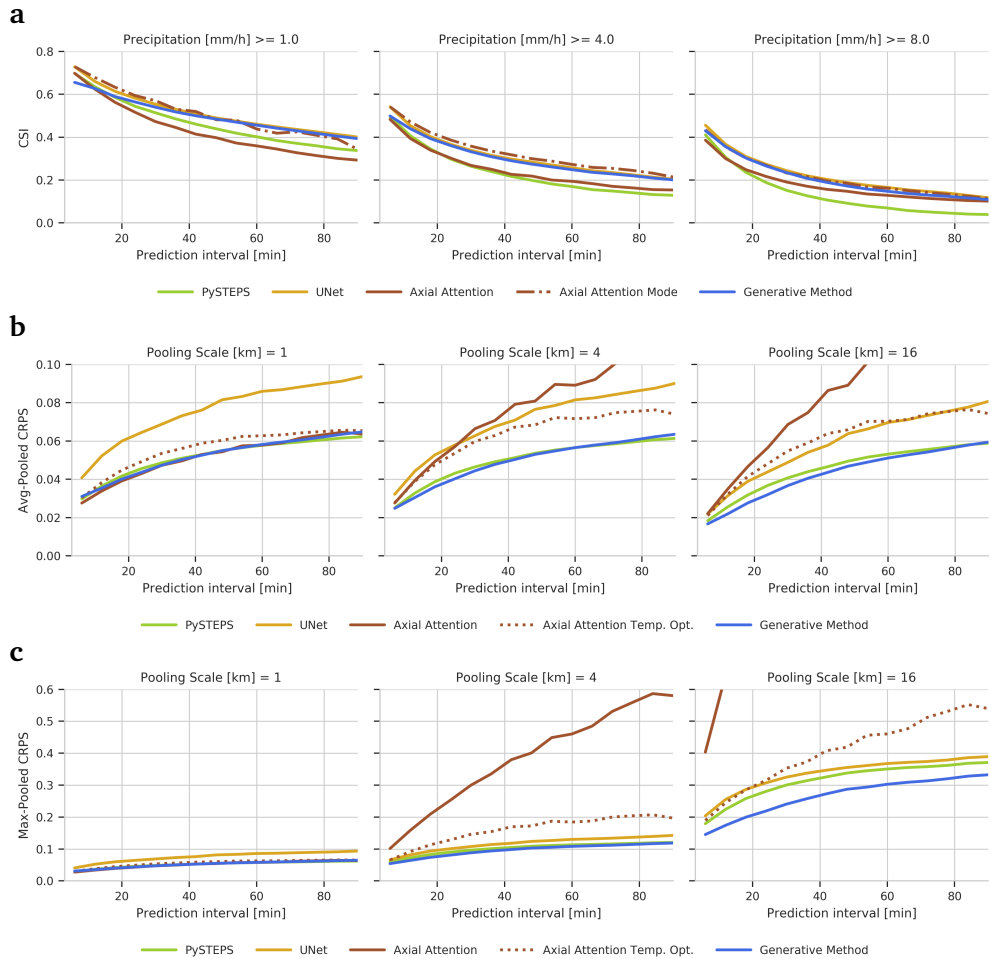

    \centering
    \newcommand{\insertfigure}[3]{\includegraphics[width=#1\textwidth]{figures/figures_US/yearly/single_run/#2/#3}}
    \begin{minipage}{0.8\textwidth}
        \begin{minipage}{0.99\textwidth}
        \textbf{a}\\
        \insertfigure{0.99}{20210210}{prob-csicc-us512-yearly-test-1-4-8classifier-sample_mean-matching_thresholds-csi.png}
        \end{minipage} \\
        \begin{minipage}{0.99\textwidth}
        \textbf{b}\\
         \insertfigure{0.99}{20210210}{prob-crps-us512-yearly-test-1-4-8crps-crps.png}
        \end{minipage} \\
        \begin{minipage}{0.99\textwidth}
        \textbf{c} \\
        \insertfigure{0.99}{20210210}{prob-crps-us512-yearly-test-1-4-8crps_max-crps.png}
        \end{minipage} \\
    \end{minipage}
    
    \caption{\textbf{Verification scores for the United States in 2019.}
    \textbf{a:} Critical Success Index across 20 samples of different models across precipitation thresholds \unit[1]{mm/hr} (left), \unit[4]{mm/hr}, \unit[8]{mm/hr} (right). We also report results for the Axial Attention mode prediction.
    \textbf{b:} CRPS of various models for original predictions (left), average rain rate over a \unit[4]{km} $\times$ \unit[4]{km} catchment area (middle), and average rain rate over a \unit[16]{km} $\times$ \unit[16]{km} catchment area (right).
    \textbf{c:} CRPS of various models for original predictions (left), maximum rain rate over a \unit[4]{km} $\times$ \unit[4]{km} catchment area (middle), and maximum rain rate over a \unit[16]{km} $\times$ \unit[16]{km} catchment area (right).
    }
    \label{fig:us_yearly_test_results}
\end{figure}

\begin{figure}[tp]
    \begin{minipage}{0.8\textwidth}
        \begin{minipage}{0.9\textwidth}
        \textbf{a} \\
        \includegraphics[width=0.99\textwidth]{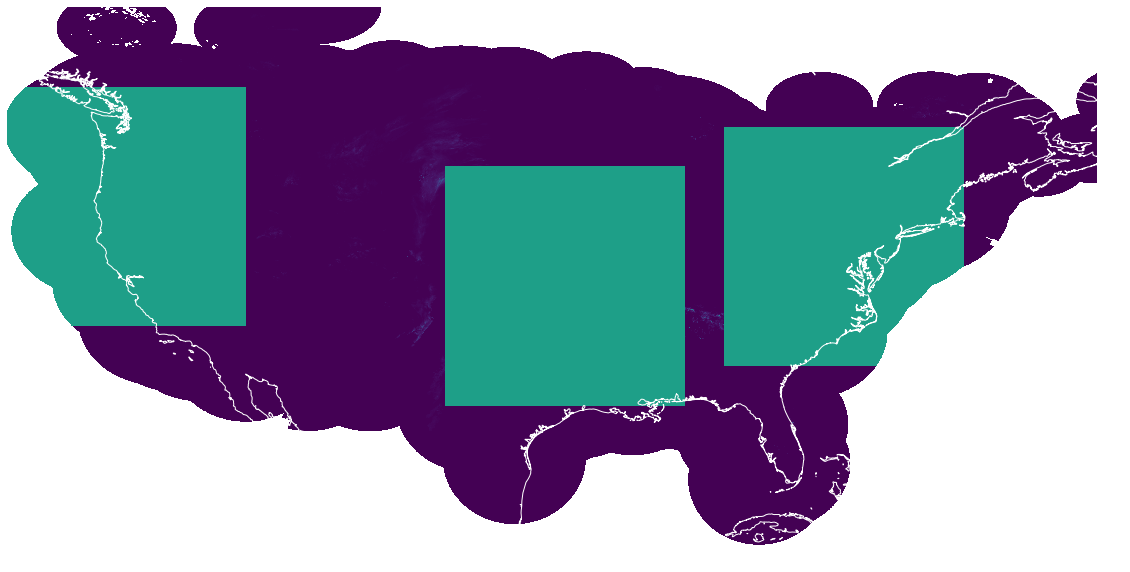}
        \end{minipage} \\
        \newcommand{\insertyrlyevent}[1]{\includegraphics[width=0.32\textwidth]{figures/figures_US/yearly/single_run/20210210/#1}}
        \textbf{b}\\
        \begin{tabular}{m{0.7em} c c c}
        \vspace{-40mm}
        \rotatebox{90}{\hspace{2.2em}Pacific Northwest} &
            \insertyrlyevent{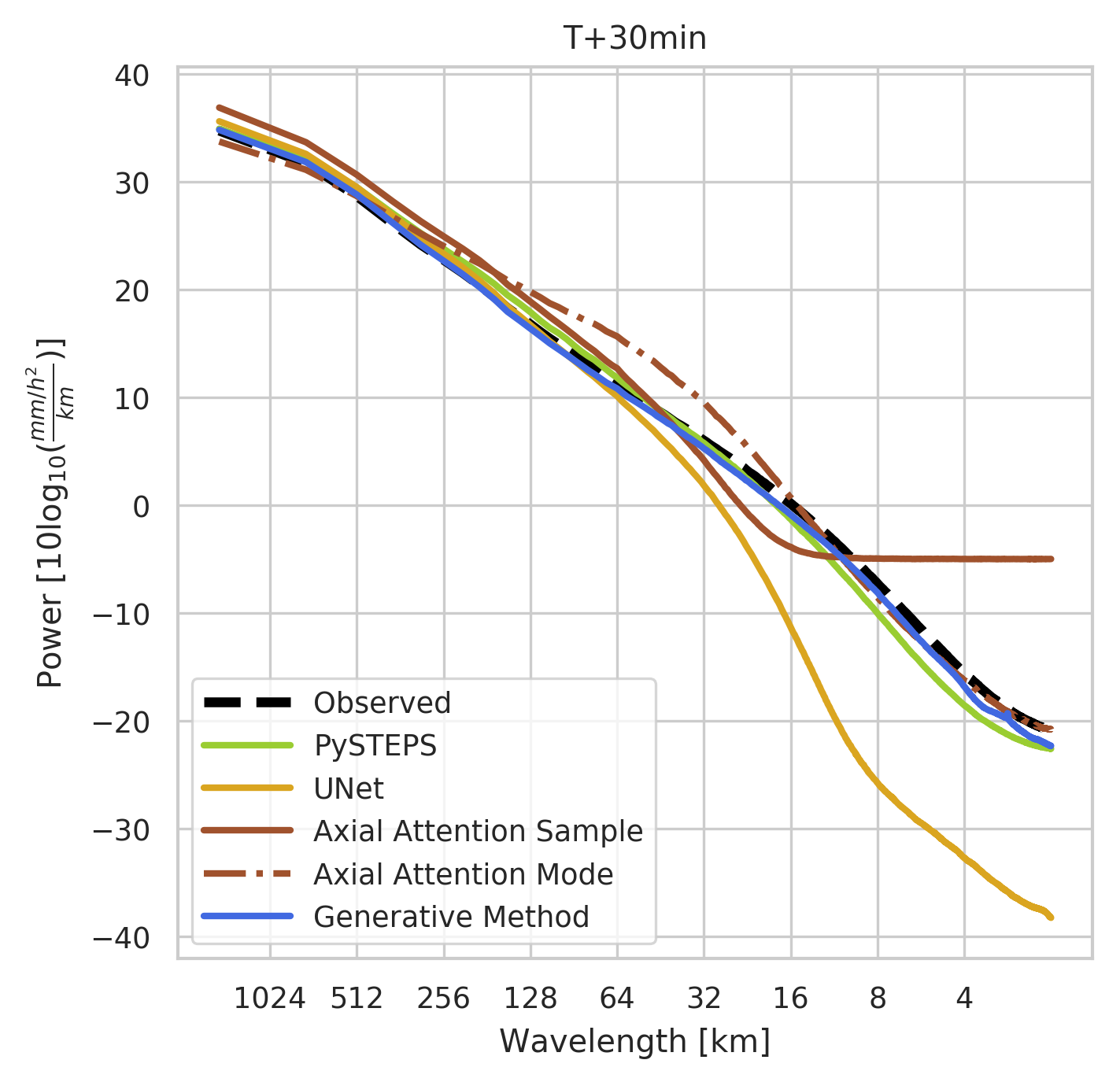} &
            \insertyrlyevent{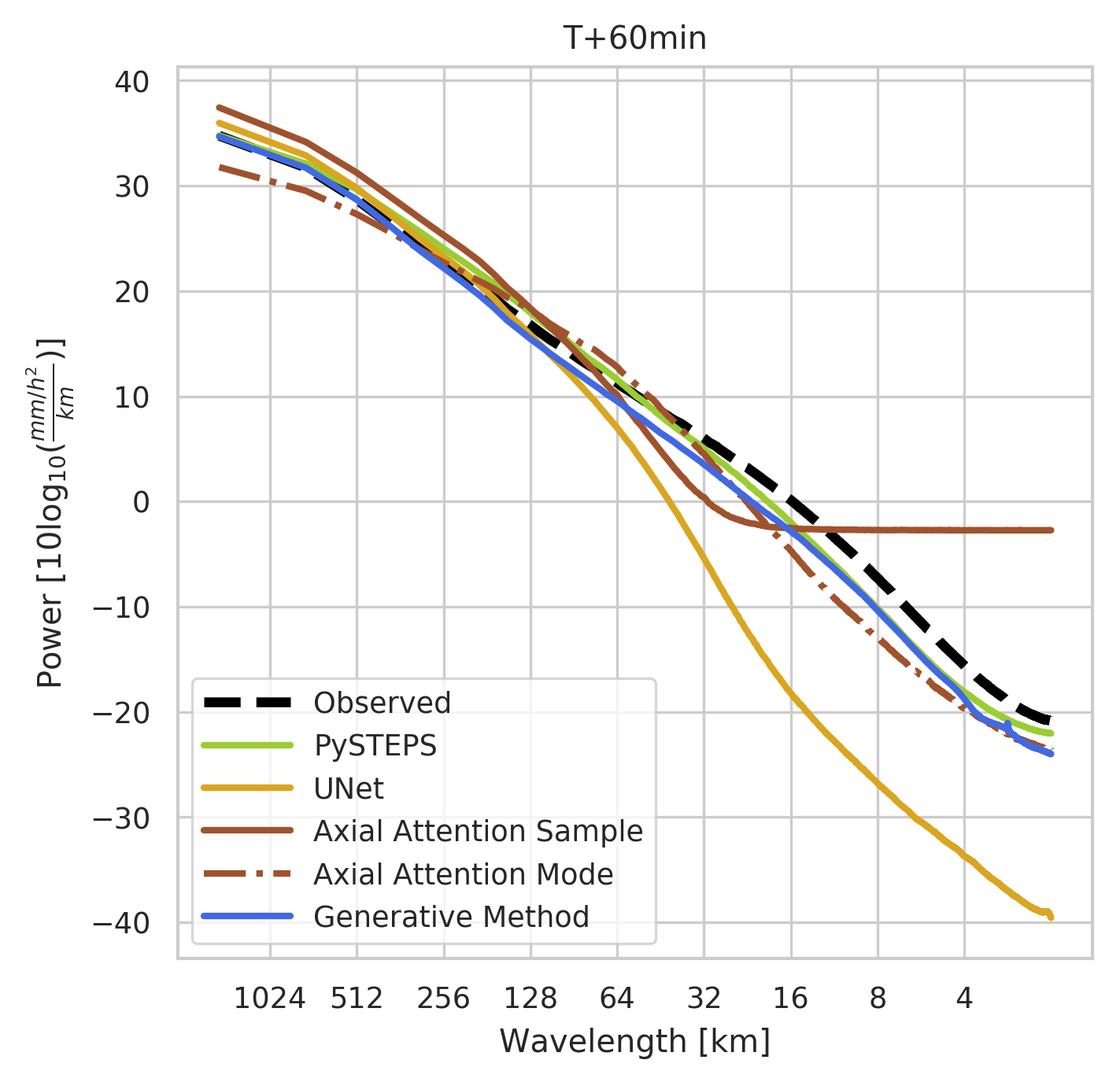} &
            \insertyrlyevent{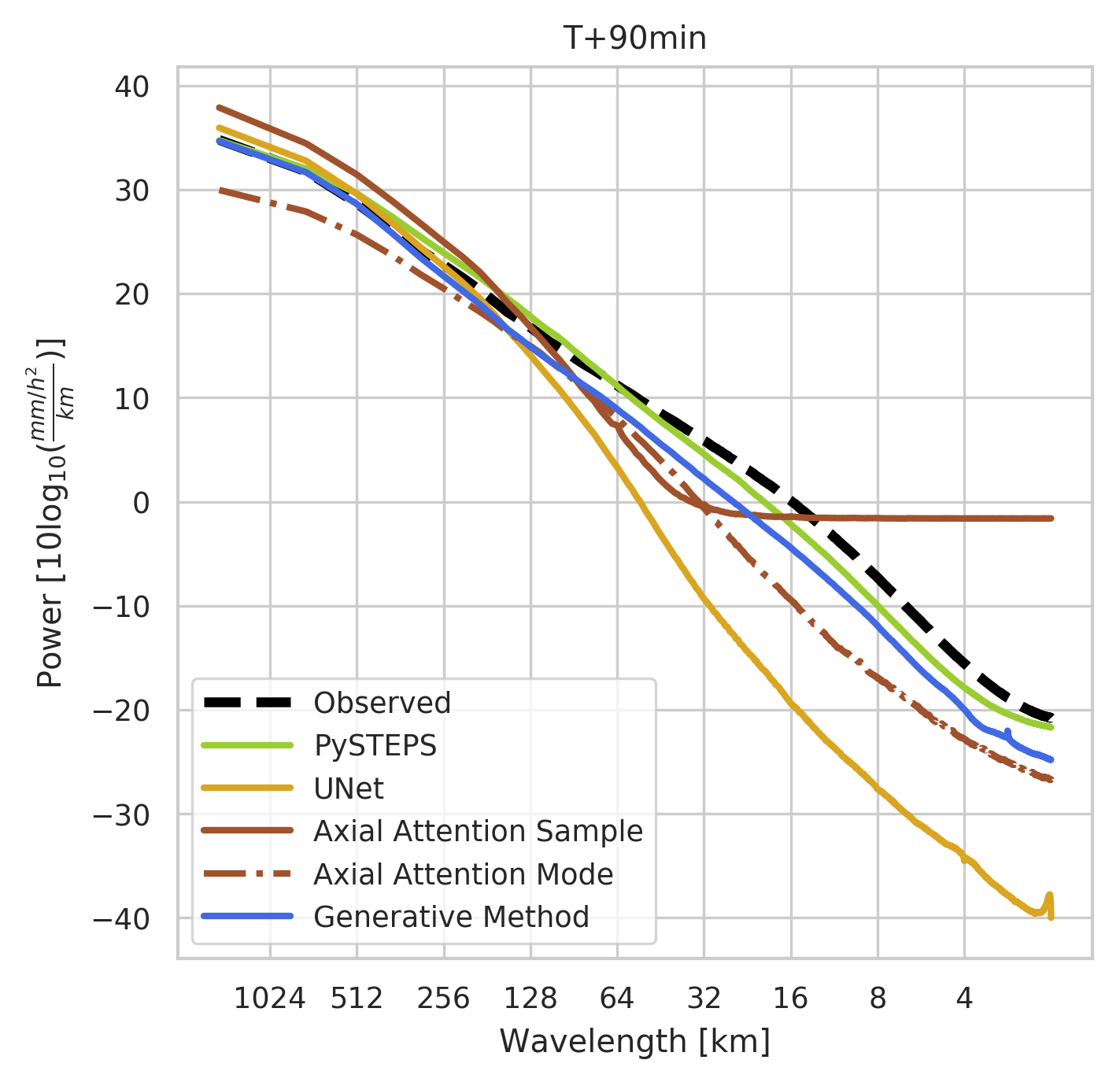}
            \\
        \end{tabular}
        \textbf{c}\\
        \begin{tabular}{m{0.7em} c c c}
        \vspace{-40mm}
        \rotatebox{90}{\hspace{4.1em}Midwest} &
            \insertyrlyevent{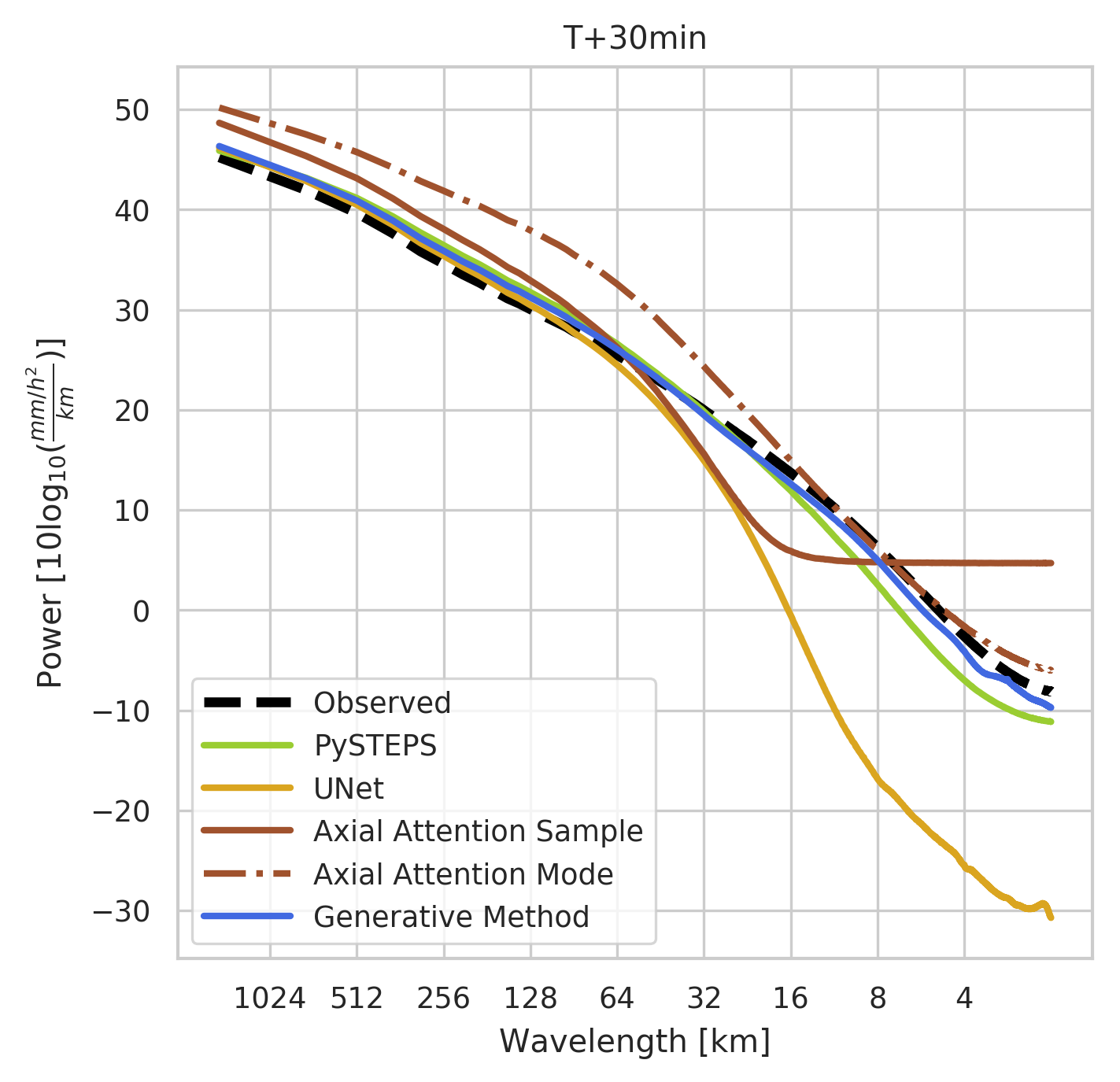} &
            \insertyrlyevent{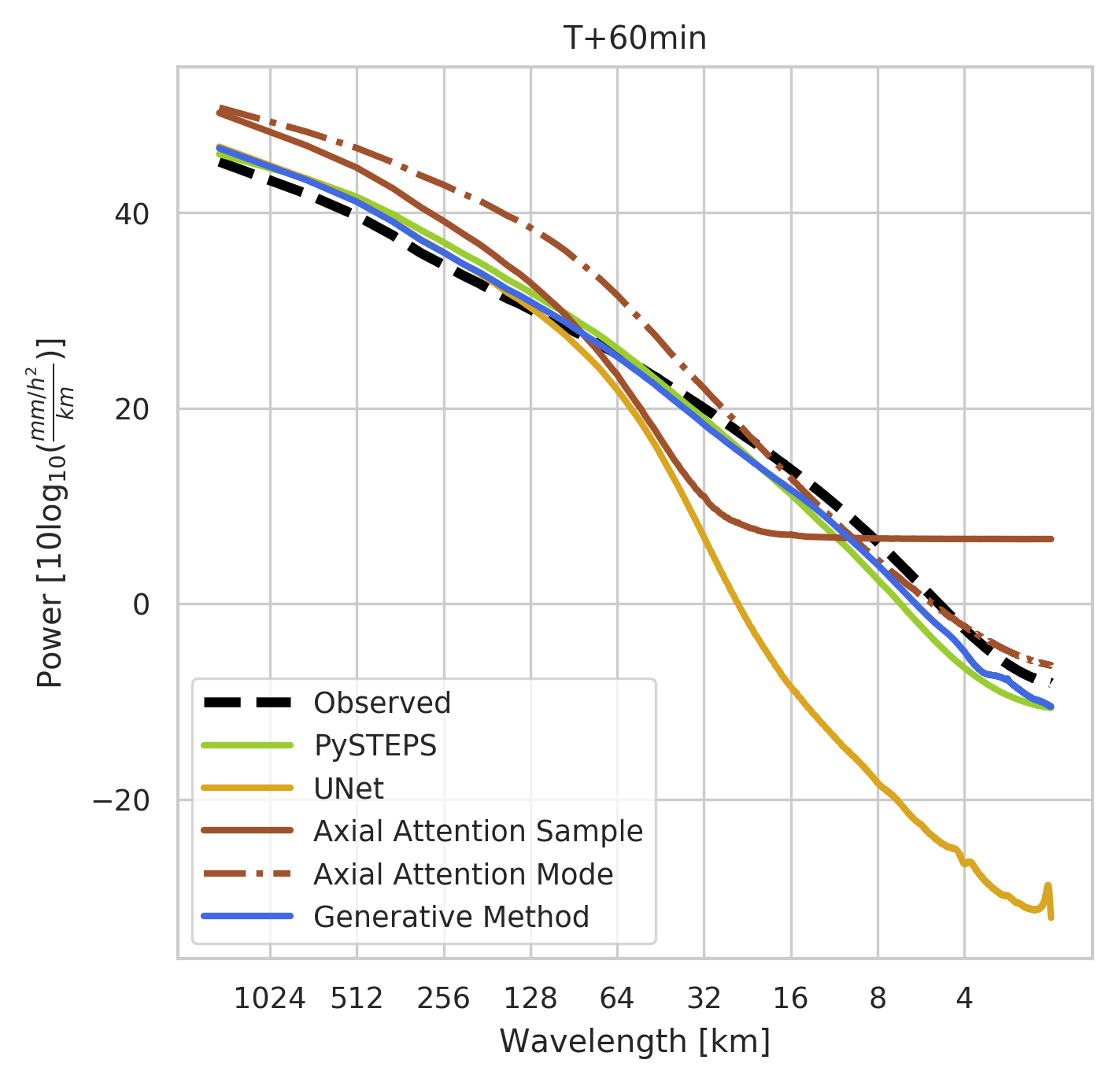} &
            \insertyrlyevent{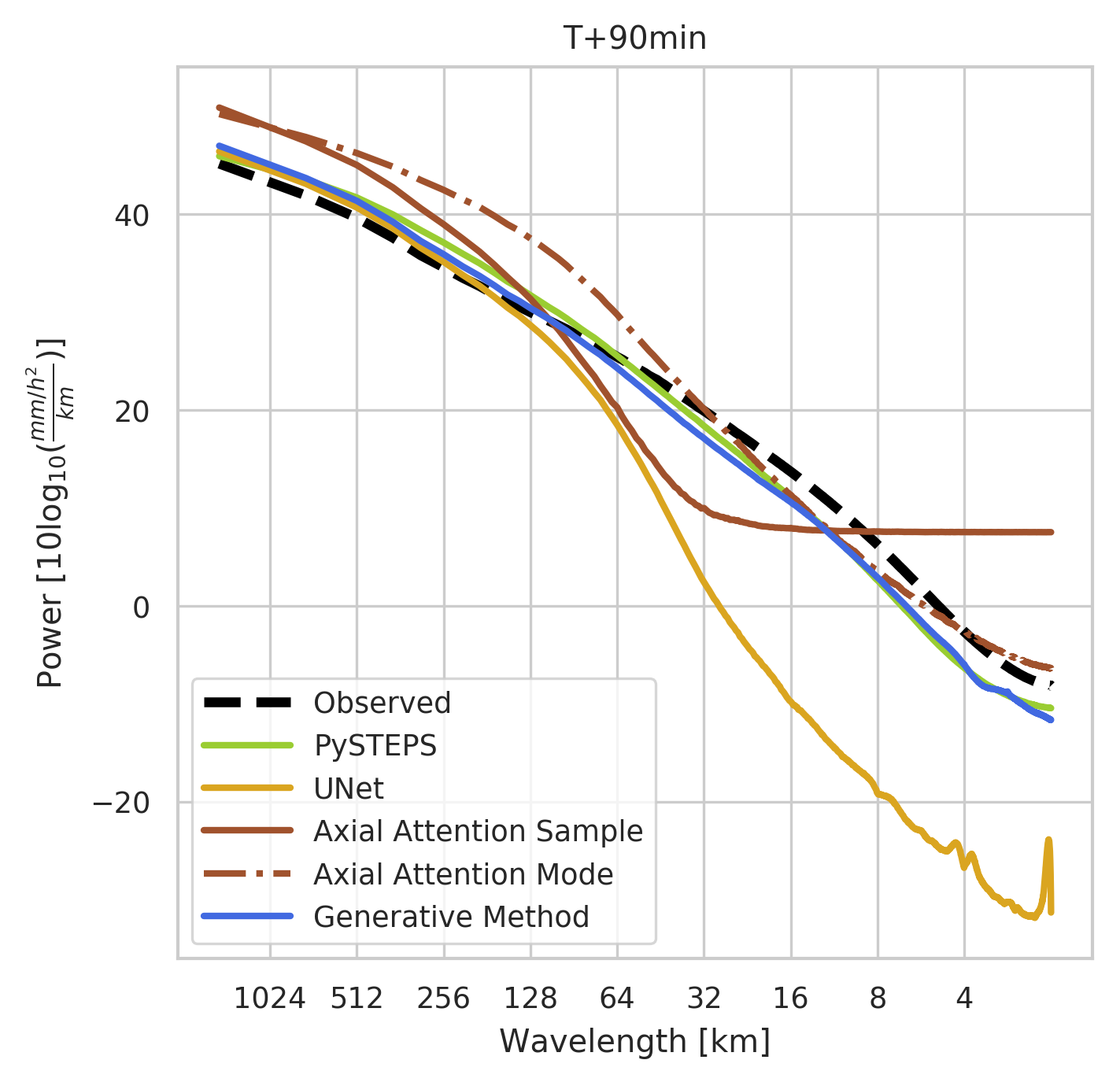}
            \\
        \end{tabular}
        \textbf{d}\\
        \begin{tabular}{m{0.7em} c c c}
        \vspace{-40mm}
        \rotatebox{90}{\hspace{3.8em}Northeast} &
            \insertyrlyevent{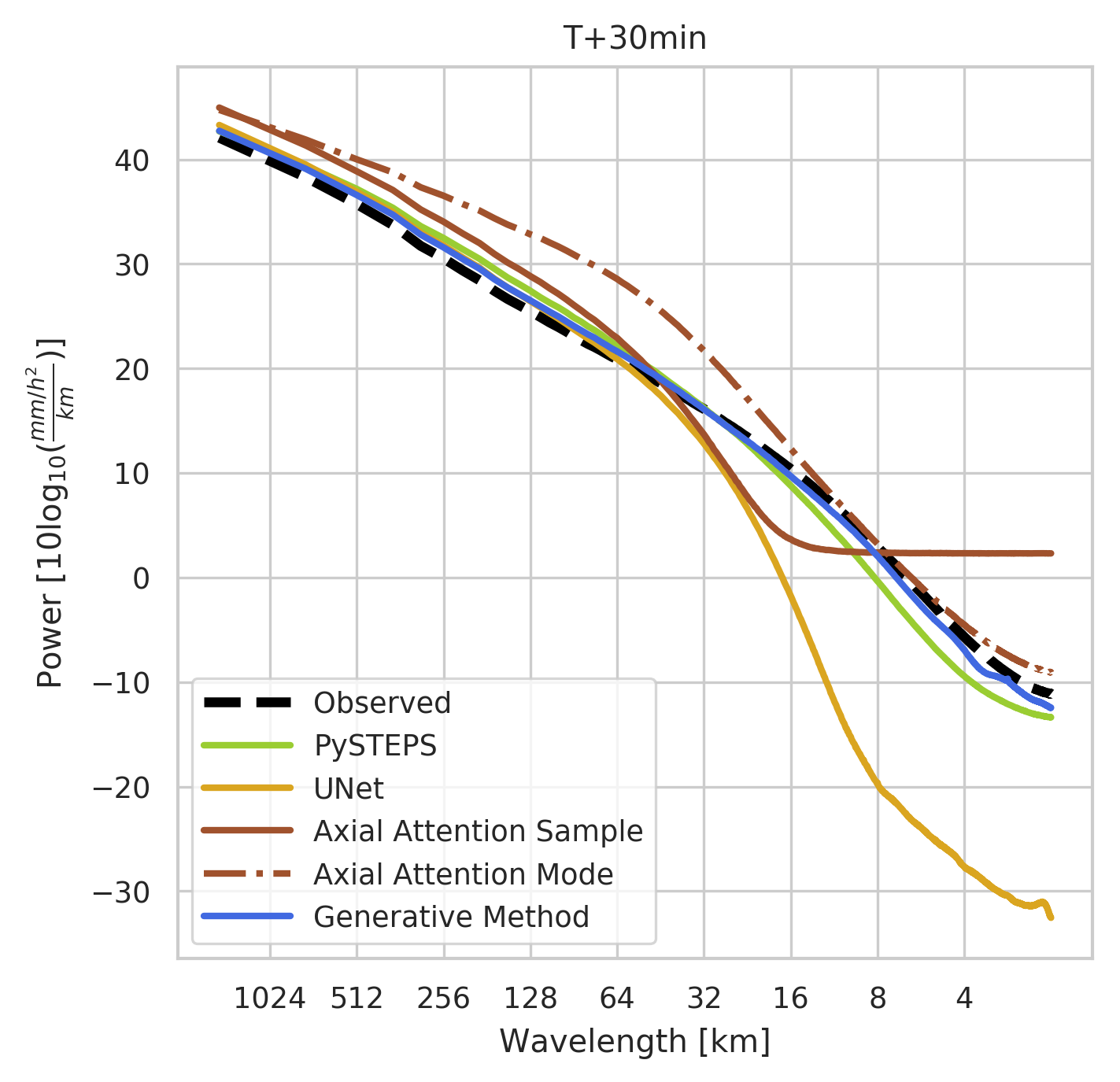} &
            \insertyrlyevent{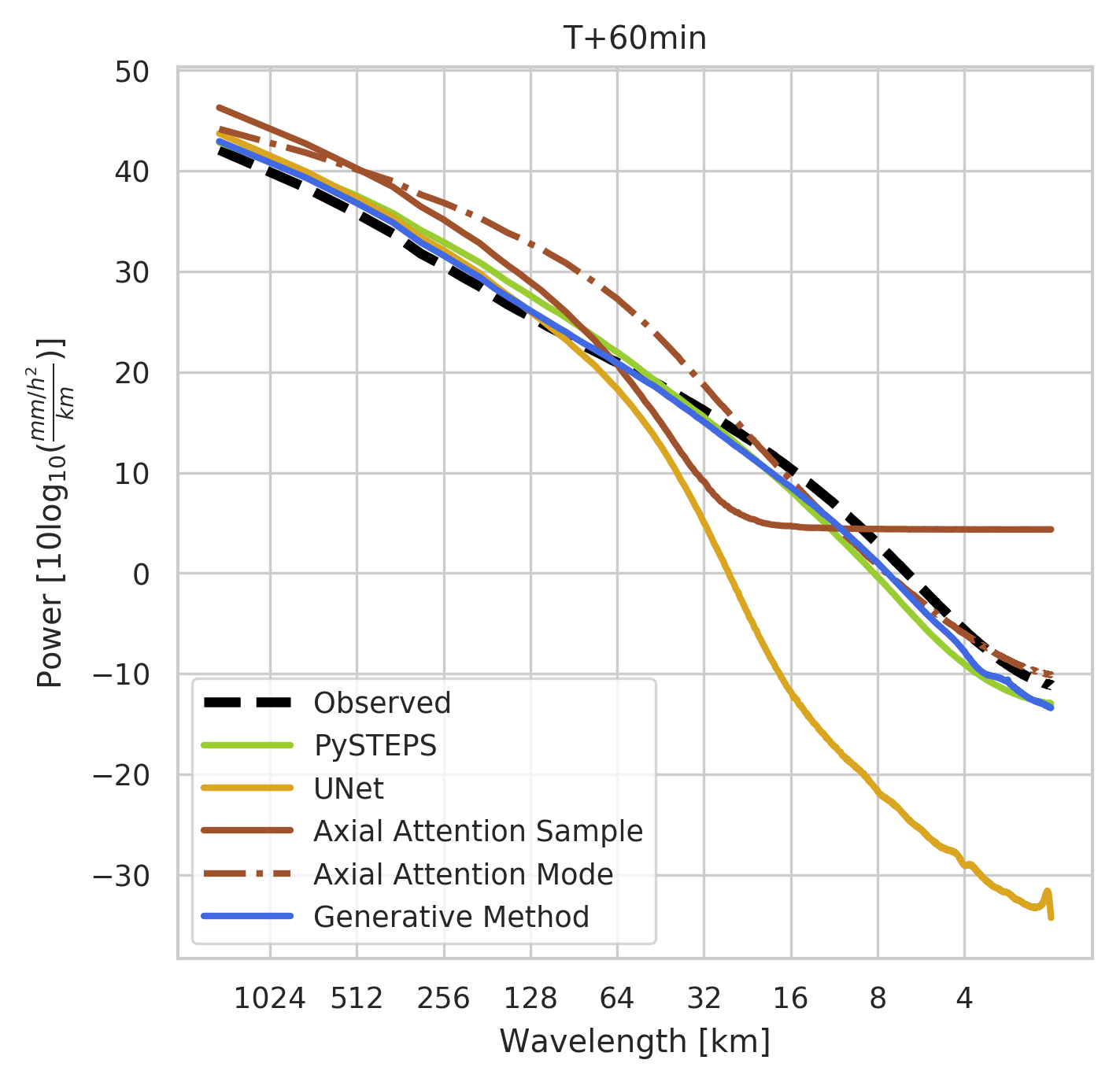} &
            \insertyrlyevent{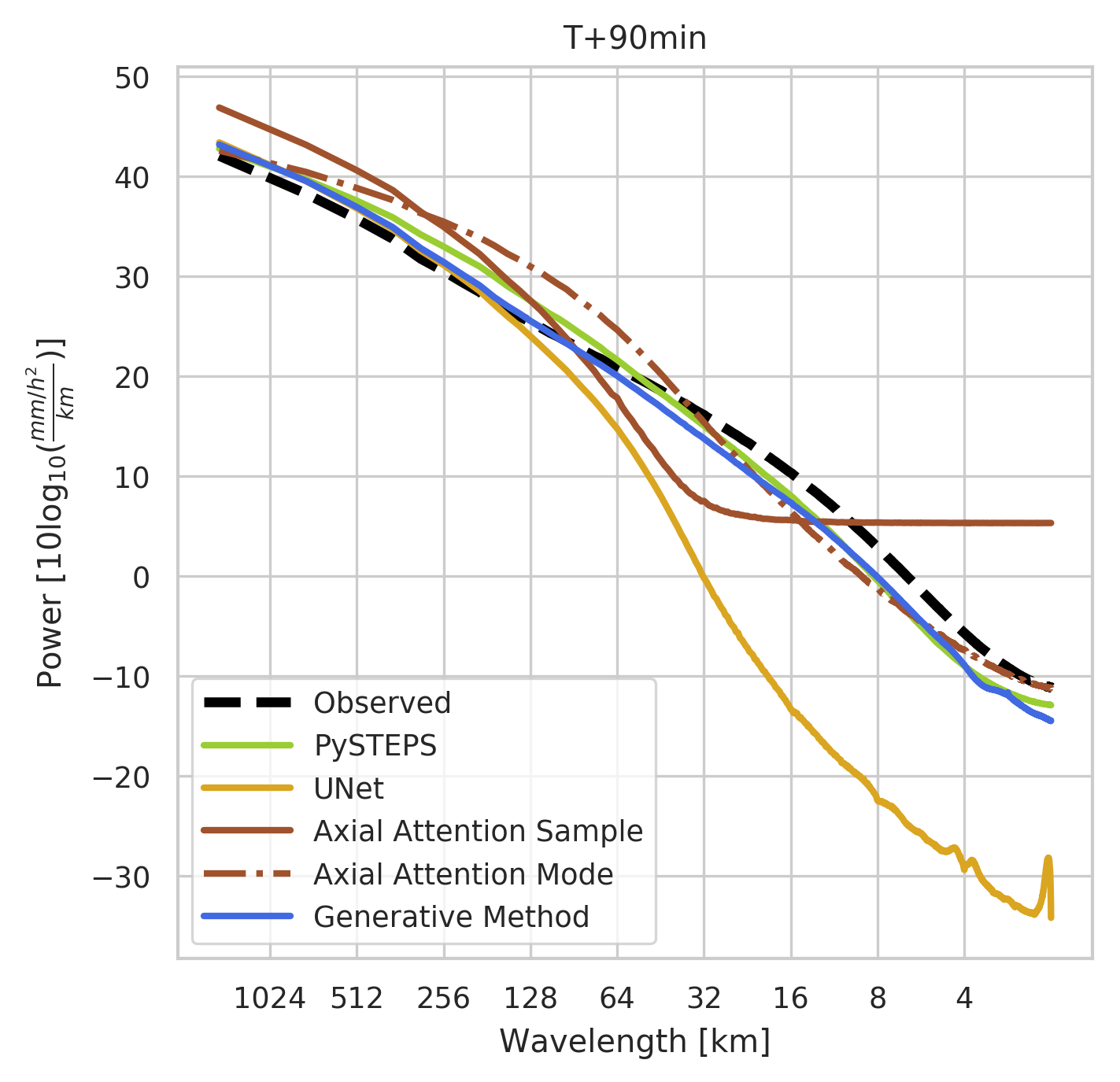}
            \\
    \end{tabular}
        
    \end{minipage}
    
    \caption{\textbf{Radially Averaged Power Spectral Density for the United States in 2019.}
    \textbf{a:} Map of United States with three 1536 $\times$ 1536 regions: Pacific Northwest (left), Midwest (middle), Northeast (right).
    \textbf{b:} Radially-averaged power spectral density for Pacific Northwest region for different models at T+\unit[30] (left), T+\unit[60] (middle), and T+\unit[90]{mins} (right).
     \textbf{c:} Radially-averaged power spectral density for Midwest region for different models at T+\unit[30] (left), T+\unit[60] (middle), and T+\unit[90]{mins} (right).
     \textbf{d:} Radially-averaged power spectral density for Northeast region for different models at T+\unit[30] (left), T+\unit[60] (middle), and T+\unit[90]{mins} (right).
     }
    \label{fig:us_yearly_psd_by_region}
\end{figure}

\begin{figure}[tp]
    \centering
    \newcommand{\insertfigure}[3]{\includegraphics[width=#1\textwidth]{figures/figures_US/yearly/single_run/#2/#3}}
    \begin{minipage}{0.8\textwidth}
        \begin{minipage}{0.99\textwidth}
        \textbf{a}\\
        \insertfigure{0.99}{20210210}{prob-bysample-us512-yearly-test-1-4-8fss.png}
        \end{minipage} \\
        \begin{minipage}{0.99\textwidth}
        \textbf{b}\\
        \insertfigure{0.99}{20210210}{prob-csicc-us512-yearly-test-1-4-8cross_correlation.png}
        \end{minipage} \\
        \begin{minipage}{0.57\textwidth}
        \textbf{c} \\
        \insertfigure{0.99}{20210210}{prob-csicc-us512-yearly-test-1-4-8reliability.png}
        \end{minipage}
        \begin{minipage}{0.42\textwidth}
        \textbf{d} \\
        \insertfigure{0.99}{20210210}{prob-csicc-us512-yearly-test-1-4-8rank-histogram.png}
        \end{minipage} 
    \end{minipage}
    
    \caption{ \textbf{Further verification scores for the United States in 2019.}
    \textbf{a:} Fractions Skill Score across 20 samples of different models across spatial scale thresholds \unit[1]{mm/hr} (left), \unit[4]{mm/hr}, \unit[8]{mm/hr} (right).
    \textbf{b:} Pearson Correlation Coefficient of various models for original predictions (left), average rain rate over a \unit[4]{km} $\times$ \unit[4]{km} catchment area (middle), and average rain rate over a \unit[16]{km} $\times$ \unit[16]{km} catchment area (right).
     \textbf{c:} Reliability diagrams and sharpness plots for two precipitation thresholds for T+\unit[60]{min} predictions.
     \textbf{d:} Rank Histogram at T+\unit[60]{min}.}
    \label{fig:us_yearly_test_results2}
\end{figure}

\putbib[refs]
\end{bibunit}

\begin{bibunit}[ieeetr]
\appendix
\renewcommand{\figurename}{Supplementary Figure}
\renewcommand{\tablename}{Supplementary Table}
\setcounter{figure}{0}  
\newpage
\section*{Supplementary Materials for Skillful Precipitation Nowcasting using Deep Generative Models of Radar}

In this supplement, we provide more details about the used datasets (supplement \ref{sec:suppl-datasets}), additional experiments mentioned in the methods (supplement \ref{sec:suppl_further_experiments}), a more detailed description of the re-implemented baselines (supplement \ref{sec:suppl_baselines}), context of the related work in nowcasting research in (supplement \ref{sup:related_work}), and the precise definitions of the used metrics and their pooled variants (supplement \ref{sec:suppl_detailed_metrics}) for reference.

\section{Additional Dataset Details}
\label{sec:suppl-datasets}

Here, we provide additional information regarding datasets.
We detail the important sampling scheme we employed to favor heavy rainfall examples in the dataset (supplement \ref{sec:importance-sampling-scheme}).
We then provide a description of the data used to construct the US datasets (supplement \ref{sub:mrms_dataset}).
Finally, we provide the precipitation rate statistics of our datasets (supplement \ref{sub:dataset-statistics}).

\subsection{Importance Sampling Scheme}
\label{sec:importance-sampling-scheme}

In the Methods we described the use of an importance sampling scheme to increase the frequency with which crops with rain were encountered during training. Most regions in the radar composite contain little or no rainfall, and such regions typically contribute little to the variance of our estimators for evaluation metrics and loss gradients.
We can take advantage of this to reduce the computational cost of evaluations without unduly compromising the statistical efficiency of these estimators. Specifically, we create sub-sampled datasets using an importance sampling scheme \cite{kahn1955use} that favours higher-rainfall examples, and use importance-weighted sums to address the bias which this introduces.

Starting from a full-frame example of size $T\times H\times W$ ---for time-length $T$, height $H$ and width $W$---the scheme samples smaller examples (i.e., crops) of size $T\times h\times w$, including examples of heavier rainfall with higher probability. Given one of these smaller examples $x_n$, the probability of sampling $x_n$ is a function of its rain rates $x_{n,c}$, where $c$ indexes over all $C = T\times h\times w$ grid cells in the example.
Saturated values are computed from rain rates as
$
x_{n,c}^{\text{sat}} = 1 - \exp(-x_{n,c} / s),
$
where $s$ is a saturation constant and $x_{n,c}$ is set to zero for masked grid cells.
These are averaged, scaled and clipped to give an acceptance probability
\begin{equation}
\label{eqn:inclusion-probability}
q_n = \min\bigl\{1, q_{\text{min}} + \frac{m}{C} \sum_{c} x_{n,c}^{\text{sat}}\bigr\},
\end{equation}
where $q_{\text{min}}$ is the minimum probability of inclusion and $m$ is a multiplier controlling the overall inclusion rate.
See Supplementary Table \ref{tab:dataset-sampling-param} for the values used for each dataset.

We consider different crops when sampling for the training, validation or test sets.
For validation and test, we only consider crops whose vertical and horizontal offsets are multiples of 64.
The probability of a given crop $x_n$ being included in the dataset is thus $q_n$ as detailed above.
In particular, for the $512\times 512$ test sets, for which we only compute metrics on the central $64\times 64$ grid cells, this ensures that every grid cell has a chance of being included in the evaluation, while reducing the overlap of crops.
For training, we want to create datasets as large as possible.
So, to ensure that examples can be sampled at all possible offsets, we first consider all the crops with a vertical and horizontal offsets multiple of 32.
If one of those examples $x_n$ is accepted with probability
$q_{n}$, instead of including it in the dataset,
we add uniform random offsets between 0 and 32 grid cells horizontally and vertically, defining a new example $x_{n'}$.
The example $x_{n'}$ is then added to the dataset and
its inclusion probability is effectively $q'_{n'} = q_{n}/32^{2}$.

When computing evaluation metrics on the subsampled validation and test sets, we use unbiased importance-weighted estimators in place of sums over the full dataset of examples $\{x_1, \ldots, x_N\}$.
We estimate $\sum_{n=1}^N S(x_n)$ using the subsample $\{x_{n_1}, \ldots, x_{n_L}\}$ as
$\sum_{l=1}^L q_{n_l}^{-1} S(x_{n_l})$.

We also explored using importance sampling weights at training time to correct bias in our estimates of loss gradients.
We found no significant advantage from this and therefore we did not use important weights at training time.

\begin{table}[tb]
\caption{\textbf{Summary of parameters used in generating full frame and sub-sampled data sets.}}
\label{tab:dataset-sampling-param}
\begin{center}
\begin{footnotesize}

\begin{tabular}{lcccccccc}
\toprule
Dataset & \multicolumn{2}{c}{Training} & \multicolumn{2}{c}{Validation} & \multicolumn{2}{c}{Test} & \multicolumn{2}{c}{Test}\\
& \multicolumn{2}{c}{Sub-sampled} & \multicolumn{2}{c}{Sub-sampled}& \multicolumn{2}{c}{Sub-sampled} & \multicolumn{2}{c}{Full-frame}  \\
\midrule
    & UK & US & UK & US & UK & US & UK & US\\
$s$ & 1.0 & 1.0 &  1.0 & 1.0 &  10.0 & 30.0 & - & - \\
$m$ & 0.1 & 0.1 &  2.2  & 0.2 &  1.0  & 0.2  & - & -\\
$q_{\text{min}}$ & 
$2\times 10^{-4}$ &  $2\times 10^{-4}$ & $5\times 10^{-3}$ & $5\times 10^{-3}$ & $2\times 10^{-5}$ & $2.5\times 10^{-4}$ &-&- \\
$T$ & 24 & 20 & 24 & 20 & 24 & 20 & 24 & 20 \\
$h$ & 256 & 256 & 256 & 256 & 512 & 512 & 1536 & 3584 \\
$w$ & 256 & 256 & 256 & 256 & 512 & 512 & 1280 & 7168 \\
Spatial offset & 32 & 32 & 256 & 256 & 64 & 64 & - & - \\
Random offset & True & True & False & False & False & False & False & False \\
Temporal offset & \multirow{2}{*}{5} & \multirow{2}{*}{6} & \multirow{2}{*}{20} & \multirow{2}{*}{24} & \multirow{2}{*}{20} & \multirow{2}{*}{24} & \multirow{2}{*}{20} & \multirow{2}{*}{24} \\
between examples (min) &  &  &  &  &  & & & \\
\bottomrule
\end{tabular}

\end{footnotesize}
\end{center}
\end{table}

\subsection{United States Dataset}
\label{sub:mrms_dataset}

To train and evaluate models of precipitation nowcasting over the United States, we use radar composites from the Multi-Radar Multi-Sensor (MRMS) system \cite{zhang2016multi, smith2016multi}.
The data is acquired with a network of 146 WSR-88D radars covering the conterminous US and 30 Canadian radars\footnote{\url{https://www.roc.noaa.gov/WSR88D/Maps.aspx}}.
We refer to \cite{zhang2016multi} for details on how reflectivity fields are converted to precipitation rates and how precipitation classification informs this transformation.
The radar composites cover latitudes between $20^{\circ}$ and $55^{\circ}$ North and longitudes between $130^{\circ}$ and $60^{\circ}$ West.
The resolution of the $3584\times7168$ composites is of $0.01^{\circ}$ in both latitude and longitude directions; this is equivalent to \unit[1.11]{km} uniformly in the North-South direction. However, in the West–East direction $0.01^{\circ}$ represents about \unit[0.6]{km} at the top of the image and about \unit[1]{km} at the bottom.
Similar to the UK data, missing values are identified by a negative value, which is used to mask irrelevant grid cells during training and evaluation.
To construct our datasets, we use the radar composites collected every 2 minutes between January 1, 2017 and December 31, 2019.
When generating an example from a sequence of radar composites, we downsample the temporal resolution of the data by 3$\times$,
ignoring 2 composites out of 3,
effectively making predictions by increments of 6 minutes.
We perform this downsampling operation since, while the nominal temporal resolution is two minutes, the duration to complete a volume scan by the radars varies from 3 to 10 minutes.
As a result, precipitation dynamics present ``skipping" patterns, with  measurements from different radars updating asynchronously (see \cite{zhang2016multi}).
Reducing the effective temporal resolution to 6 minutes mitigates this ``skipping" effect and makes the temporal resolution comparable to the UK dataset.
We cap the rain rates at the value of \unit[1024]{mm/hr}.

\subsection{Dataset Statistics}
\label{sub:dataset-statistics}

In Suplementary Table \ref{tab:dataset-distn}, we show summary statistics of the distribution of the rainfall amount in mm/hr for both the UK and US dataset. These statistics show the high proportion of no rain grid cells, differences in high-intensity grid cells between the US and UK data, and the effect of the importance sampling scheme towards higher intensity grid cells to support learning. 
\begin{table}[tb]
\caption{\textbf{Rainfall Distributions (in percentage).} Shown for the UK and US yearly test set per location and dataset type. Statistics are computed across 15 consecutive frames from $10^{4}$ randomly drawn examples for the sub-sampled datasets, and $10^{3}$ examples for the full frame datasets.
}
\label{tab:dataset-distn}
\begin{center}
\begin{small}
\begin{sc}
\begin{tabular}{c | cccc}
\toprule
Dataset & UK & UK & US & US \\
Interval in mm/hr & Full-Frame & Subs. & Full-Frame & Subs. \\
\midrule
$=0.0$        &  89.18     &   69.14    &   94.52   &   79.9	\\
$(0, 0.1]$    &   1.72      &    3.61     &    0.00   &    0.00	\\
$(0.1-1.0]$   &   5.96     &   16.18    &    3.46   &    9.82	\\
$(1.0-4.0]$   &   2.75     &    9.59     &    1.66   &    7.97	\\
$(4.0-5.0]$   &   0.16     &    0.62     &    0.11   &    0.74	\\
$(5.0-8.0]$   &   0.16     &    0.64     &    0.13   &    0.91	\\
$(8.0-10.0]$  &   0.03    &    0.11     &    0.03   &    0.21	\\
$> 10.0$      &   0.03     &    0.11     &    0.08   &    0.45	\\
\bottomrule
\end{tabular}
\end{sc}
\end{small}
\end{center}
\end{table}

\section{Additional Experimental Analysis}
\label{sec:suppl_further_experiments}

\subsection{Additional Quantitative Evaluation on Yearly Data Splits}
We provide additional quantitative evaluation for the models trained on the yearly train-validation-test data split.

\paragraph{Training variability.}
In order to quantify the variance of the training algorithm, we trained 6 instances of \modelname, each trained with a different sequence of training examples.
We show the performance for CSI and pooled CRPS (with one standard deviation error bars) in Supplementary figures, \ref{fig:uk_yearly_test_results_rand_seed} and \ref{fig:uk_yearly_test_results2_rand_seed}, respectively.

\begin{figure}[htp]
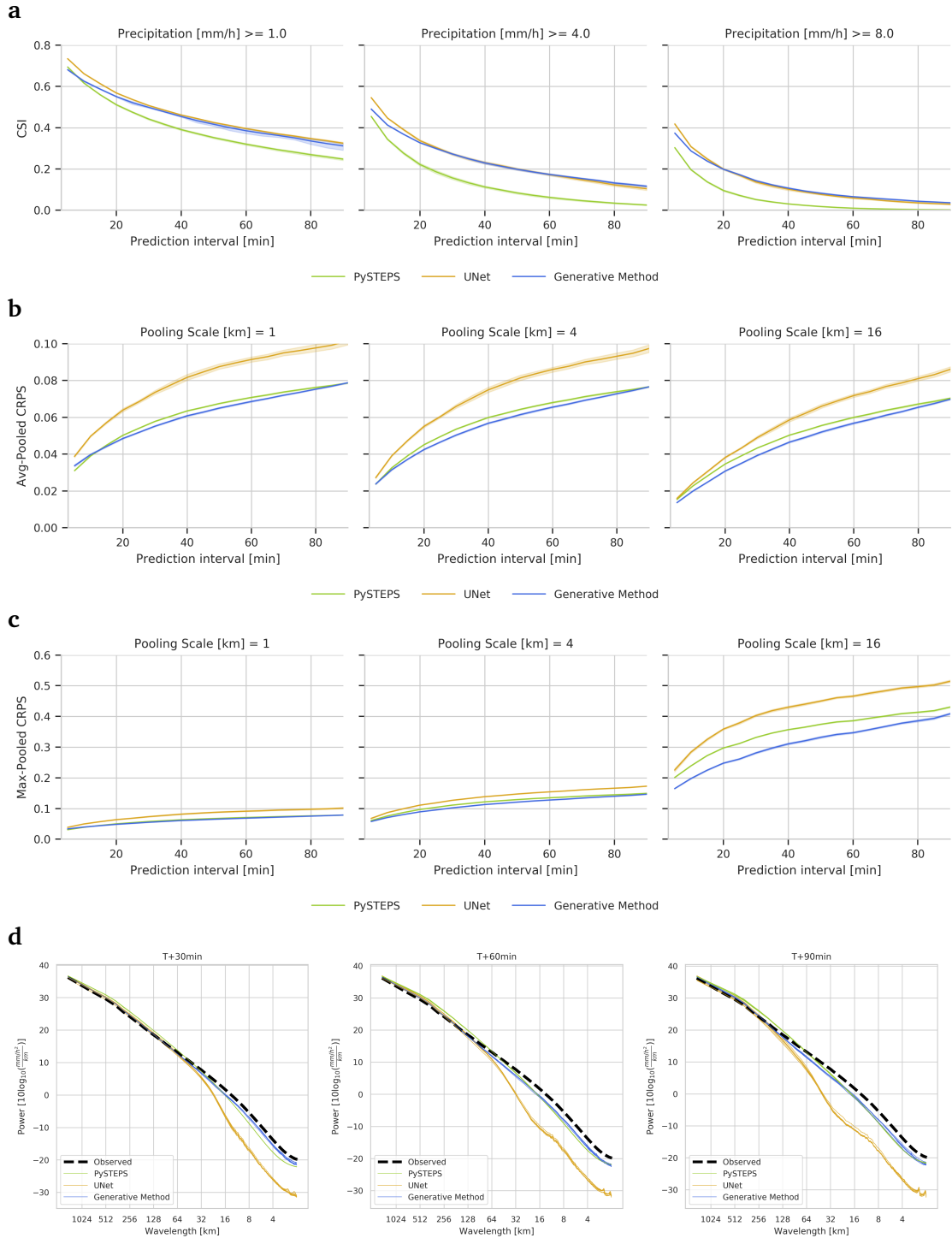

    \centering
    \newcommand{\insertfigure}[3]{\includegraphics[width=#1\textwidth]{figures/figures_UK/yearly/random_seed/#2/#3}}
    \begin{minipage}{0.9\textwidth}
        \begin{minipage}{0.99\textwidth}
        \textbf{a}\\
        \insertfigure{0.99}{20210210}{probseed-uk512-yearly-test-seedclassifier-sample_mean-matching_thresholds-csi.png}
        \end{minipage} \\
        \begin{minipage}{0.99\textwidth}
        \textbf{b}\\
        \insertfigure{0.99}{20210210}{probseed-uk512-yearly-test-seedcrps-crps.png}
        \end{minipage} \\
        \begin{minipage}{0.99\textwidth}
        \textbf{c}\\
        \insertfigure{0.99}{20210210}{probseed-uk512-yearly-test-seedcrps_max-crps.png}
        \end{minipage} \\
        \begin{minipage}{0.99\textwidth}
        \textbf{d} \\
        \insertfigure{0.32}{20210210}{psd_nimrod_yearly_splits_seed_30min_16kmObs.png}
        \insertfigure{0.32}{20210210}{psd_nimrod_yearly_splits_seed_60min_16kmObs.png}
        \insertfigure{0.32}{20210210}{psd_nimrod_yearly_splits_seed_90min_16kmObs.png}
        \end{minipage} \\
    \end{minipage}
    
    \caption{ \textbf{Verification scores for the United Kingdom in 2019 for six Generative Method and UNet initializations and six runs of PySTEPS.}
    \textbf{a:} Critical Success Index across 20 samples of different models across precipitation thresholds \unit[1]{mm/hr} (left), \unit[4]{mm/hr}, \unit[8]{mm/hr} (right) with 95\% confidence interval.
    \textbf{b:} Average-pooled CRPS of various models for original predictions (left), average rain rate over a \unit[4]{km} $\times$ \unit[4]{km} catchment area (middle), and average rain rate over a \unit[16]{km} $\times$ \unit[16]{km} catchment area (right) with 95\% confidence interval. 
     \textbf{c:} Max-pooled CRPS of various models for original predictions (left), maximum rain rate over a \unit[4]{km} $\times$ \unit[4]{km} catchment area (middle), and maximum rain rate over a \unit[16]{km} $\times$ \unit[16]{km} catchment area (right) with 95\% confidence interval. 
     \textbf{d:} Radially-averaged power spectral density for full-frame 2019 predictions for different models across the 6 initializations.  }
    \label{fig:uk_yearly_test_results_rand_seed}
\end{figure}

\begin{figure}[htp]
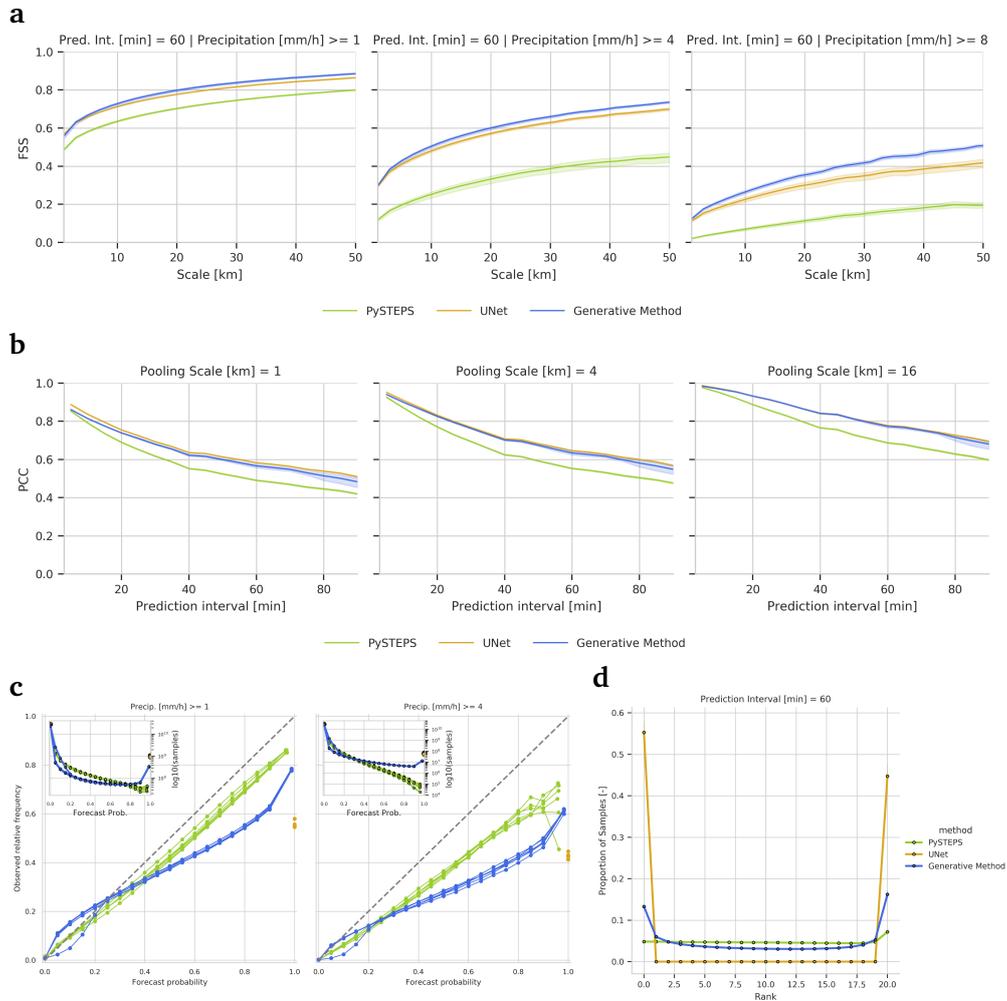

    \centering
    \newcommand{\insertfigure}[3]{\includegraphics[width=#1\textwidth]{figures/figures_UK/yearly/random_seed/#2/#3}}
    \begin{minipage}{0.8\textwidth}
        \begin{minipage}{0.99\textwidth}
        \textbf{a}\\
        \insertfigure{0.99}{20210210}{probseed-uk512-yearly-test-seed-1-4-8fss.png}
        \end{minipage} \\
        \begin{minipage}{0.99\textwidth}
        \textbf{b}\\
        \insertfigure{0.99}{20210210}{probseed-uk512-yearly-test-seedcross_correlation.png}
        \end{minipage} \\
        \begin{minipage}{0.57\textwidth}
        \textbf{c} \\
        \insertfigure{0.99}{20210210}{probseed-uk512-yearly-test-seedreliability.png}
        \end{minipage}
        \begin{minipage}{0.42\textwidth}
        \textbf{d} \\
        \insertfigure{0.99}{20210210}{probseed-uk512-yearly-test-seedrank-histogram.png}
        \end{minipage} 
    \end{minipage}
    
    \caption{ \textbf{Further verification scores for the United Kingdom in 2019 for six initializations.}
    \textbf{a:} Fractions Skill Score across different spatial scales at thresholds \unit[1]{mm/hr} (left), \unit[4]{mm/hr}, \unit[8]{mm/hr} (right) with 95\% confidence interval.
    \textbf{b:} Pearson Correlation of various models for original predictions (left), average rain rate over a \unit[4]{km} $\times$ \unit[4]{km} catchment area (middle), and average rain rate over a \unit[16]{km} $\times$ \unit[16]{km} catchment area (right) with 95\% confidence interval.
     \textbf{c:} Reliability Plot across individual initializations.
     \textbf{d:} Rank Histogram with 95\% confidence interval.}
    \label{fig:uk_yearly_test_results2_rand_seed}
\end{figure}

\paragraph{Justifying the choice of loss.}
In Supplementary figure \ref{fig:uk_yrly_ablation_results}, we show the influence of the different combinations of losses on the test set performance.
It can be seen that the proposed combination of losses is important for obtaining favourable results across a combination of metrics. For example, using only the per-grid-cell regularization loss leads to best results with regards to per-grid-cell CSI (a), however it leads to a significant increase of the CRPS metrics (c) and unfavorable spatial frequency (PSD) characteristics (d). 
\begin{figure}[tp]
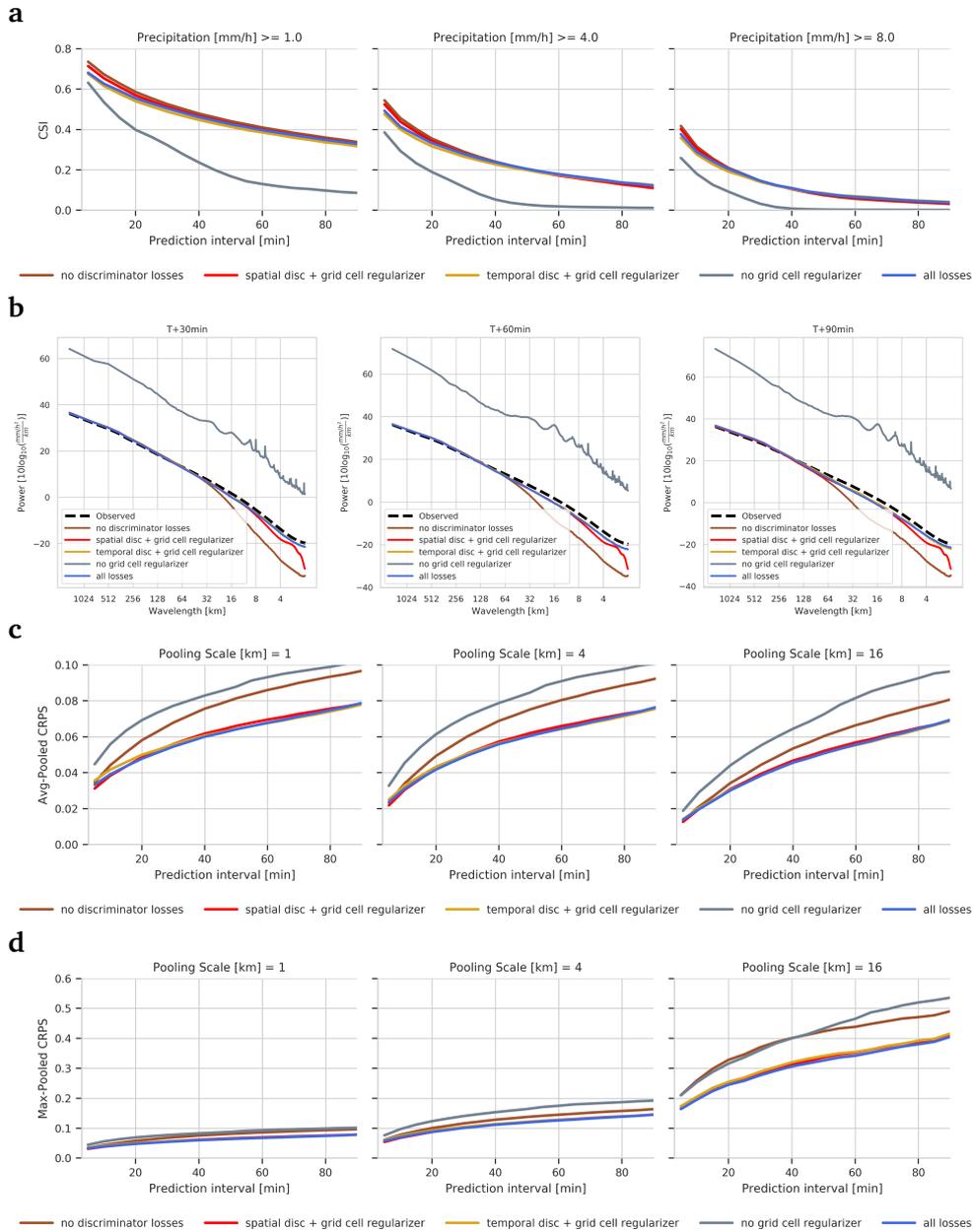

    \centering
    \newcommand{\insertfigure}[3]{\includegraphics[width=#1\textwidth]{figures/uk_yearly_single_run_ablation_expt/#2/#3}}
    \begin{minipage}{0.8\textwidth}
        \begin{minipage}{0.99\textwidth}
        \textbf{a}\\
        \insertfigure{0.99}{20210210}{prob-discabl-uk512-yearly-test-1-4-8classifier-sample_mean-matching_thresholds-csi.png}
        \end{minipage} \\
                \begin{minipage}{0.99\textwidth}
        \textbf{b} \\
        \insertfigure{0.32}{20210210}{psd_ablations_nimrod_yearly_splits_30min.png}
        \insertfigure{0.32}{20210210}{psd_ablations_nimrod_yearly_splits_60min.png}
        \insertfigure{0.32}{20210210}{psd_ablations_nimrod_yearly_splits_90min.png}
                \end{minipage} \\
        \begin{minipage}{0.99\textwidth}
        \textbf{c}\\
        \insertfigure{0.99}{20210210}{prob-discabl-uk512-yearly-test-1-4-8crps-crps.png}
        \end{minipage} \\
        \begin{minipage}{0.99\textwidth}
        \textbf{d}\\
        \insertfigure{0.99}{20210210}{prob-discabl-uk512-yearly-test-1-4-8crps_max-crps.png}
        \end{minipage} \\
    \end{minipage}
    
    \caption{ \textbf{Verification scores for the United Kingdom in 2019 with ablations of the Generative Method.} Compared losses are grid cell regularization with no discriminator losses, spatial discriminator with grid cell regularization, temporal discriminator with grid cell regularization, discriminator losses without grid cell regularization, and all three losses (the two discriminator losses and the regularization).
    \textbf{a:} Critical Success Index across 20 samples of different models across precipitation thresholds \unit[1]{mm/hr} (left), \unit[4]{mm/hr}, \unit[8]{mm/hr} (right).
    \textbf{b:} Radially-averaged power spectral density for full-frame 2019 predictions for different models. 
    \textbf{c:} CRPS of various models for original predictions (left), average rain rate over a \unit[4]{km} $\times$ \unit[4]{km} catchment area (middle), and average rain rate over a \unit[16]{km} $\times$ \unit[16]{km} catchment area (right).
     \textbf{c:} CRPS of various models for original predictions (left), maximum rain rate over a \unit[4]{km} $\times$ \unit[4]{km} catchment area (middle), and maximum rain rate over a \unit[16]{km} $\times$ \unit[16]{km} catchment area (right).
     }
    \label{fig:uk_yrly_ablation_results}
\end{figure}

\paragraph{Evaluation by Precipitation Type}
\label{sub:qualitative_event_selection}
The Methods and Extended Data \ref{fig:uk_results_rain_type} showed performance on different rain types. To assign labels for rain types, each example is labeled with the following properties. An example is labeled as frontal if directed movement of air masses with a visible front line exists, non-frontal if no precipitation exists, or mixed if part of the example contains a front. An example is labeled scattered if scattered rain exists, unscattered if none exists, and mixed if this changes during the example. Finally, the motion of precipitation is labeled coherent if the precipitation field was moving coherently, in the same direction, non-coherent otherwise, or mixed if both types exist in an example. This last property is used to classify whether or not precipitation is advective. We  then compared the performance of our model and baselines using the quantitative metrics across the different subtypes.

\subsection{NWP Results}
\label{ssec:nwp_results}

The Methods and Extended Data \ref{fig:uk_yearly_test_results_nwp_preliminary} reference comparison to how the NWP would perform within the nowcasting timescales. We perform a basic comparison against the NWP performance. In our case, we use the rainflux variable from the Met Office Deterministic UK model (UKV) \cite{bush2020first}.
The rainflux variable in this model has 5 minutes temporal and \unit[1.5]{km} spatial resolution, which we upsample to the OSGB36 \unit[1]{km} scale reference grid.
In practice, data assimilation is performed multiple times a day to initialize a new NWP with the most recent observations.
Therefore, we conduct the evaluation of the NWP baseline by restricting the test set to examples that align the first prediction target with the initialization of a NWP.
This gives two advantages to the NWP.
First, it evaluates its performance in the ideal case where it has just been updated with observations.
Second, the NWP is given instantaneous access to its prediction after initialization.
In an operational setting, the time taken by data assimilation would make the first few predictions of the model unavailable in real time.
We use data generated by NWPs initialized four times a day to construct this evaluation over the UK.
For the CSI and CRPS, we use the $512\times512$ test set containing 3704 examples.
For PSD, we use the $1536\times1280$ test set containing 1409 examples.
Note that because of their reduced size, those datasets contain limited number of high-precipitation events.
Supplementary figure \ref{fig:uk_yearly_test_results_nwp_preliminary} shows a comparison of PySTEPS, UNet and \modelname~to NWP on CSI, CRPS and PSD.
Overall, the NWP performs poorly compared to other baselines and \modelname~at nowcasting timescales on CSI and CRPS, and since it preserves physical properties, makes predictions with good spectral characteristics. 

\subsection{Empirical Comparison of the Sub-sampled and Full-frame Dataset}
\label{ssec:ssot_vs_ff}
We provide additional details of the empirical Comparison of the $512\times512$ sub-sampled and $1536\times1280$ full-frame Dataset here. 
In Supplementary figure \ref{fig:uk_yearly_ssot_vs_ff}, we show a quantitative comparison of CSI scores obtained on the full-frame UK dataset (yearly splits, test set) versus the results obtained on the proposed sub-sampled dataset. Because computing the predictions for the full-frame dataset is computationally prohibitive, we evaluate the STEPS and \modelname~only on an ensemble of 5 samples (instead of 20, which is used in other experiments).
We observe no quantitative difference, further motivating the use of the $512\times512$ sub-sampled dataset for all metrics but PSD.

\begin{figure}[tp]
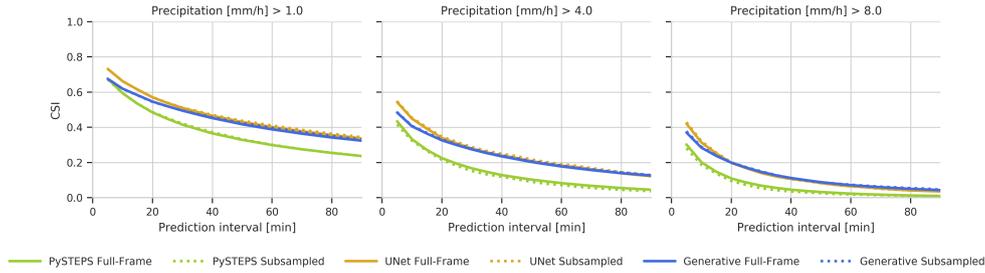

    \centering
    \newcommand{\insertfigure}[3]{\includegraphics[width=#1\textwidth]{figures/figures_UK/ssot_vs_ff/#2/#3}}
    \begin{minipage}{0.8\textwidth}
     	\insertfigure{0.99}{20210210}{deterministic-uk-yearly-test-dsetcompclassifier-matching_thresholds-csi.png}
    \end{minipage}
    
    \caption{ \textbf{Verification scores for the United Kingdom on yearly splits show no significant difference between the sub-sampled and full-frame datasets.} Results computed over an ensemble of 5 samples for PySTEPS and the Generative Method. Datasets are specified in \cref{tab:dataset-sampling-param}.}
    \label{fig:uk_yearly_ssot_vs_ff}
\end{figure}

\subsection{Computational Speed}
In \cref{tab:execution-speed} we show execution speed of some of the selected models.
We evaluate the speed of sampling by comparing speed on both CPU (10 cores of AMD EPYC processor) and GPU (NVIDIA V100) hardware for the deep learning models. We generate 10 samples and report the median time. 
Because the Axial attention model requires tiling at evaluation, it is not directly comparable and is not included here.

\begin{table}[tb]
	\caption{\textbf{Execution speed of selected models.} Computed for a single samples of the full-resolution UK data sample (1536 $\times$ 1280 locations), and reported as median time across 10 samples.
	}
	\label{tab:execution-speed}
	\begin{center}
		\begin{small}
			\begin{sc}
				\begin{tabular}{c | cccc}
					\toprule
					Model & PySTEPS & UNet & \modelname \\
					\midrule
					CPU Speed [s] & 69.54 & 2.78 & 25.66 \\
					GPU Speed [s] & - & 0.1 & 1.27 \\
					\bottomrule
				\end{tabular}
			\end{sc}
		\end{small}
	\end{center}
\end{table}

\section{Axial Attention Model and MetNet Adaptations}
\label{sec:suppl_baselines}
We focused our analysis on two strong baselines, PySTEPS \citep{pulkkinen2019pysteps} and MetNet \citep{sonderby2020metnet}, in addition to an NWP reference baseline, and this section provides additional details of our implementation of these baseline methods. We focus here on the adaptations of axial attention-based methods for radar data. 

MetNet~\cite{sonderby2020metnet} is a deep learning method for precipitation nowcasting that was demonstrated to outperform optical flow-based methods and numerical weather prediction models on the MRMS US dataset, and that was evaluated on prediction horizons up to 8 hours. MetNet's computation follows three steps:
\begin{itemize}
    \item Spatial downsampling. Input frames are transformed using a convolutional network to 256-dimensional feature maps (spatial shape $64\times64$).
    \item Temporal encoding. Feature maps from consecutive input time steps are integrated using the Convolutional LSTM architecture~\cite{xingjian2015convolutional} into a single 384-layer map (spatial shape $64\times64$).
    \item An axial attention-based aggregation~\cite{ho2019axial} is applied to that map to produce a single prediction frame of spatial shape $64\times64$.
\end{itemize}

We made several adaptations to the MetNet algorithm\footnote{The axial attention implementation is available at \url{https://github.com/lucidrains/axial-attention}} for use with radar data only. For disambiguation, we refer to the original implementation from~\cite{sonderby2020metnet} as MetNet and to our implementation as the Axial Attention model.

Each input and output pixel in MetNet corresponds to 0.01 deg of latitude and 0.01 deg of longitude, i.e. about \unit[1]{km} (not accounting for the approximations due to the Mercator projection).
The spatial extent of MetNet inputs was $1024\times1024$, and the method was trained to predict the central $64\times64$ crop. \cite{sonderby2020metnet}  assumes that precipitation moves at \unit[60]{km/hr}, leaving the 480 pixel margins on the input to account for rain cloud advection over eight hours and to ensure that the model makes predictions using only data within the input frames. 
In our study, we operate at a scale of \unit[1]{km} per pixel (UK data) and 0.01 deg per pixel (US data), making predictions at horizons up to 90 minutes. Assuming similar rain cloud maximum speed, we set margins at 96 pixels, with $256\times256$ inputs and $64\times64$ outputs.

To make the axial attention model comparable to other methods in this paper, we reduce the temporal extent of the input context from seven frames (covering 90 minutes with 15 minute intervals, in the MetNet implementation) down to four frames covering 20 minutes with five minute intervals. In the ablation studies in the original paper, two input frames sufficed for equivalent CSI performance at rain rate \unit[1]{mm/hr}.

MetNet relies on several layers of input data: precipitation measured by radar, Geostationary Operational Environmental Satellite
16 (GOES-16), per-pixel elevation embedding, per-pixel latitude and longitude position embeddings, per-frame time embeddings. As we did not have access to geostationary data for the UK, we conducted our study using only elevation, position and time embeddings.  We extracted SRTM elevation data from CGIAR\footnote{SRTM elevation data \url{http://srtm.csi.cgiar.org/srtmdata/}}, made available via the Google Earth Engine\footnote{Google Earth Engine \url{https://earthengine.google.com/}} and pre-processed it to be a single layer aligned with the OSGB coordinates of UK data or with the WGS-84 coordinates of US data. We capped elevation at \unit[4000]{m} (US) and \unit[1000]{m} (UK) and normalized values to be between 0 and 10 to match the scale of precipitation data. We computed positional embeddings by calculating the cosine and sine values of $x$ and $y$ coordinates at 4 different scales, following~\cite{vaswani2017attention}, resulting in 16 additional input layers. The paper did not specify how latitude and longitude were embedded. Similar to the original paper, we added three layers for temporal embeddings, corresponding to values of month/12, day/31 and hour/24 to the axial attention model.

In \cite{sonderby2020metnet}, this additional context was added to account for orographic rain, differing climate and seasonality in precipitation. We evaluated the contribution of these additional 20 layers of elevation, positional and temporal embeddings to the performance of the axial attention model, as measured by CSI. Counter-intuitively, we observed that this additional data did not improve performance, as changes in CSI were not statistically significant. We hypothesize that, at the time scale at which nowcasting operates, the phenomena represented by these additional embeddings are modeled in the dynamics of precipitation over four input frames.
MetNet applied a transformation $\tanh(\log(x + 0.01) / 4)$ to input precipitation data. In our analysis, we found that transforming the input data had no effect, so we included no transformation.

MetNet is trained with lead times of 15 to 480 minutes, with a temporal resolution of 15 minutes. For each choice of prediction target, the target time is specified as an input to the model, using one-hot embeddings concatenated with input frames; models with different prediction lead times thus share the same parameters. In our case, we train MetNet to predict at lead times 5 to 90 minutes with a temporal resolution of 5 minutes (UK), and 6 to 90 minutes with a temporal resolution of 6 minutes (US).

While the axial attention model is trained similarly to other deterministic methods such as UNet (i.e., by using a separate loss term for each grid cell), it outputs a distribution over precipitation levels for each pixel, corresponding to the logits of a multinomial. MetNet predicted precipitation over 512 bins of width \unit[0.2]{mm/hr}, from 0 to \unit[102.4]{mm/hr}. Such a binning scheme does not conform to the empirical distribution of precipitation, with higher amounts of rain becoming increasingly rare, and requires a large number of network parameters to represent the output distribution. For this reason, we reduce the number of bins $N$ and rescale target data $x$, starting with normalizing it to $x_n = (x - m) / (M - m)$, where the maximum of precipitation is set at $M=60$ and the minimum at $m=0$, followed by $\mu$-law re-scaling to $x_{\mu} = \frac{\log(1 + \mu*x_n)}{\log(1 + \mu)}$ and finally binning $N \times x_{\mu}$ on integer values between 0 and $N-1$. The number of bins $N=32$, and the $\mu$-law coefficient $\mu=256$ were chosen by cross-validation. Note that for $N=32$ and $\mu=256$, all precipitation values above \unit[50]{mm/hr} are clustered in the last bin.

Maximum likelihood training of the axial attention model corresponds to minimizing the cross-entropy between the predicted distribution and the ground truth categorical label. The per-pixel loss function was weighted by the square root of precipitation level, as we found this weighting helpful for predicting heavier rainfall. During evaluation, the mode of the output distribution is selected as the deterministic prediction. Note that the samples generated by MetNet have each pixel sampled independently of its neighbors. While these samples are useful for assessing the uncertainty of the per-grid-cell precipitation values, they result in noisy realizations that display grainy image features. For the purpose of human evaluation, we only presented the maximum likelihood estimate predictions (by taking the mode of the predicted distribution for each pixel) to the human judges.

Since MetNet produces the logits $y = \log p(x)$ of the probability distribution over precipitation values, we can use it to sample different realizations and estimate the model uncertainty for CRPS metrics. To do so, we compute the probability $p(x) = \frac{ \exp (y_i/T) }{ \sum _j \exp (y_j/T) }$ using softmax with different scaling coefficients (temperatures) $T$ and sample from that estimated distribution. $T=1$ corresponds to the scaling used during training, the limit case $T \rightarrow 0$ corresponds to taking the mode. Using CRPS scores, we chose temperature $T=0.5$ on the validation set.

\section{Related Work}
\label{sup:related_work}

Nowcasting is a long-standing problem in weather prediction, and our work is informed by a broad range of existing approaches and considerations. We defer to \citep{prudden2020review} for a general overview of nowcasting. In this section, we elaborate on the context of existing work and the facets of the nowcasting problem they address and that were developed in the main paper. 

\paragraph{Deep learning architectures on lower-resolution radar.}
A majority of papers required the dataset to be resized to overcome computer memory limitations, and thus operated at much coarser resolutions than \unit[1]{km} $\times$ \unit[1]{km}. One of the earliest models was a Convolutional Long Short-Term Memory (ConvLSTM) recurrent neural network for deterministic 90-minute rain/no-rain prediction on a $480\times480$ HKO-7 dataset resized to $100\times100$ \citep{shi2017deep}. On that same resized dataset, an alternative spatio-temporal architecture with stacked RNN layers was proposed \citep{wang2017predrnn}. Similar PredRNN++ and PredNet architectures were used for precipitation nowcasting over Sao Paulo, Brazil and Kyoto, Japan, respectively \cite{bonnet2020precipitation, marrocu2020performance}, although their work did not include standard metrics of performance. A ``star-bridge'' architecture, which uses precipitation-specific ConvLSTM layers, connections among ConvLSTM layers, and a two-threshold loss (at zero and three \unit{mm/hr}), was used for deterministic prediction on a Shanghai dataset, resized from $500\times500$ to $100\times100$ \cite{cao2019precipitation}. An architecture combining 3DCNNs and bidirectional ConvLSTMs was evaluated on a radar dataset over Guangdong, Hong Kong, and Macao, resized from $500\times500$ to $100\times100$ for deterministic prediction with horizons of 48 minutes. A hierarchical RNN architecture was used for deterministic predictions up to 3 hours ahead on the $480\times480$ HKO-7 dataset, resized to $128\times128$ \cite{jing2020hprnn}.

\paragraph{High-resolution nowcasting.}
Fewer works propose deterministic models that can operate at the same resolution as the underlying data and do high-resolution forecasting. These include the Convolutional Gated Recurrent Units (ConvGRU) and Trajectory GRU (TrajGRU) \cite{shi2017deep} (the latter model incorporates an optical flow module in a ConvGRU-like layer to model convective dynamics), as well as UNets with classification and regression outputs, respectively \cite{agrawal2019machine} and \cite{ayzel2020rainnet}. The star-bridge ConvLSTM architecture was made to work on a full-resolution $500\times500$ dataset over China \cite{chen2020strong}. Finally, \cite{sonderby2020metnet} use a ConvLSTM encoder and axial attention decoder to perform eight hour pointwise probabilistic prediction  by sliding a $64\times64$ window over a much larger area covering the United States.

\paragraph{Loss functions to address blurry predictions.}
As noted by some authors, predictions from these deterministic methods tend to be blurry.
As a result, there have been attempts to introduce other alternative losses to increase prediction realism. One method is to ``adversarially regularize'' neural networks, which instead of creating a probabilistic model, adds discriminator losses to deterministic predictions. Most of the proposed methods generate predictions on limited spatial or temporal resolution, for example operating on $64\times64$ data \cite{singh2017deep}, or requiring to resize a $480\times480$ dataset to $256\times256$ and to make predictions at three time steps (30, 60, and 90 minutes) only \cite{Jing2019AENNAG}. One model used multi-elevation inputs, resized from $480\times480$ to $128\times128$, for 60-minute prediction, and found promising performance on low-threshold rainfall (below or above \unit[0.5]{mm/hr}), but noted temporally inconsistent predictions \cite{jing2019mlc}. One notable exception to the limited resolution of inputs used a UNet model with a deterministic \emph{pix2pix} discriminator \cite{isola2017image} on full resolution data, but its CSI performance was similar to optical flow for prediction horizons up to 60 minutes \cite{veillette2020sevir}. Thus far, ensemble methods have received limited attention, with the exception of recent work on training four separate TrajGRUs with the proposed loss modified with different thresholds \cite{Franch_2020}.
Computer vision-inspired losses have been used to improve sharpness in deterministic predictions \cite{tran2019computer, tran2019multi}.
\paragraph{Preliminary GAN approaches.}
A probabilistic Conditional GAN model was recently proposed, which uses an autoregressive generator, conditioned on the regression output of a ConvLSTM model, and trained with adversarial losses to decrease blurriness in predictions up to 60 minutes \cite{liu2020mpl}; the authors did not report accuracy results though and noted the poor performance of the model; moreover, and perhaps owing to its autoregressive generator, predictions became blurrier over time.

\paragraph{Related problems and further data sources.}
Broadening the literature survey, we notice that deep learning models have been used for super-resolution (also called \emph{downscaling}) instead of prediction \cite{leinonen2020,adewoyin2020tru}, including GAN-based downscaling of radar images of precipitation \cite{chen2019generative, leinonen2020} and snowwater equivalent prediction from topographical data and meteorological forcings \cite{manepalli2019emulating}. There has been some recent work on nowcasting using satellite data, which tends to have much lower spatial (\unit[10]{km}) and temporal (\unit[30]{min}) resolution, but can cover the entire globe \cite{kumar2020convcast, zantedeschitowards,dewitt2020rainbench}. Finally, machine learning techniques other than deep learning models, such as a Koopman operator \cite{zheng2020hybrid} or Support Vector Machines (SVM) \cite{han2017machine} have been considered for nowcasting.

\section{Verification Metrics}
\label{sec:suppl_detailed_metrics}

For completeness, we provide further details about the per-grid-cell metrics for point predictions  (\ref{sub:per-grid-cell-metrics-for-point-pred}), ensemble predictions (\ref{sub:per-grid-cell-metrics-for-ensemble-pred}), pooled neighborhood metrics (\ref{sub:pooled-neigh-metrics}) and the whole-frame metric (\ref{sub:whole-frame-metrics}) that we relied on in the paper.

With the exception of full-frame metrics, we evaluate using the subsampled datasets described in \ref{sec:importance-sampling-scheme}. In these datasets each example consists of $T \times h \times w$ grid cells, where $T = M + N$. Models condition on the first $M$ context frames, and make predictions which are evaluated against the subsequent $N$ frames, using the central $M \times 64 \times 64$ grid cells only to ensure that models are not penalized by boundary effects. We refer to these as the example's \emph{target grid cells}. $M$ and $N$ are 4 and 18 for the UK data and 4 and 15 for the US data.

\subsection{Per-grid-cell metrics for point predictions}
\label{sub:per-grid-cell-metrics-for-point-pred}

Per-grid-cell metrics are computed over all target grid cells in all examples in the evaluation dataset. We index these target grid cells using a single index $i$,
and note that a single observed grid cell may occur multiple times as a target, where it is forecast at different lead times by different overlapping examples. We will write $F_i$ for the model's forecast for target grid cell $i$, and $O_i$ for the corresponding ground truth observation. Each target grid cell $i$ is associated with a weight
$
w_i = m_{radar,i} \cdot q_{n_i}^{-1},
$
which is applied whenever we sum over all grid cells. Here $m_{radar}$ is a binary mask which excludes grid cells for which no radar observation is available, and $q_{n_i}^{-1}$ is a per-example importance weight. This is the inverse of the inclusion probability $q_{n_i}$ defined in \cref{sec:importance-sampling-scheme} \cref{eqn:inclusion-probability}, for the example $n_i$ containing target grid cell $i$. It is used to implement the importance sampling scheme described in \cref{sec:importance-sampling-scheme}.
For convenience, in the following we write $\hat{w}_i$ for the normalized weight $w_i / \sum_{i'} w_{i'}$.

\subsubsection{Mean squared error (MSE) and Pearson Correlation Coefficient (PCC)}

These metrics give a continuous measure of the accuracy of real-valued point predictions:
\begin{equation}
    MSE = \sum_i \hat{w}_i (F_i - O_i)^2; \qquad
    PCC = \sum_i \hat{w}_i \frac{(F_i - \mu_F)}{\sigma_F} \frac{(O_i - \mu_O)}{\sigma_O},
\end{equation}
where $\mu_F, \mu_O, \sigma_F, \sigma_O$ are $w$-weighted means and standard deviations over all $F_i$ and $O_i$ respectively. Lower is better for MSE, and higher is better for PC.

\subsubsection{Critical Success Index (CSI)}

CSI \cite{schaefer1990critical} evaluates binary forecasts of whether or not rainfall exceeds a threshold $t$, for example low rain ($t = \unit[2]{mm/hr}$), medium rain ($t = \unit[5]{mm/hr}$) or heavy rain ($t = \unit[10]{mm/hr}$). It aims to give a single summary of binary classification performance that rewards both precision and recall, and is popular in the forecasting community. It is defined as
$$CSI = \frac{TP}{TP + FP + FN}.$$
We compute TP, FP and FN as sums of weights $w_i$ over grid cells for which the forecast is respectively a true positive ($F_i \geq t, O_i \geq t$), false positive ($F_i \geq t, O_i < t$) and false negative ($F_i < t, O_i \geq t$).

CSI is a monotonic transformation of the more-widely-known $f_1$ classifier score (\textrm{CSI} $= f_1 / (2 - f_1)$), 
and can also be viewed as a Jaccard similarity or Intersection over Union (IoU) metric computed over all target grid cells. Higher is better for CSI.

\subsection{Per-grid-cell Metrics for Ensemble Predictions}
\label{sub:per-grid-cell-metrics-for-ensemble-pred}

Ensemble models give an ensemble of $N > 1$ forecasts for each example, which we will think of as i.i.d. samples from the predictive distribution of a probabilistic model. In our evaluation we use $N=20$.

\subsubsection{Continuous Ranked Probability Score (CRPS)}

CRPS \cite{matheson1976scoring,gneiting2007strictly} is a proper scoring rule \cite{gneiting2007strictly} for univariate distributions, which we use to score the per-grid-cell marginals of a model's predictive distribution against observations. It is defined per grid cell as:
$$
\mathbb{E}|F - O| - \frac12 \mathbb{E}|F - F'|,
$$
where $F$ and $F'$ are drawn independently from the predictive distribution and $O$ is the observation. Lower is better for CRPS.
We compute unbiased estimates of CRPS using $\widehat{CRPS}_{PWM}$ from \cite{zamo2018estimation}, with the $N$ ensemble members as samples. These are then averaged over all grid cells as $\sum_i \hat{w}_i \widehat{CRPS}_{PWM,i}$.

\subsubsection{Reliability and Sharpness Diagrams}

We use reliability diagrams \cite{murphy1977reliability} to measure calibration for  ensemble forecasts of whether rain exceeds a threshold $t$. The diagram plots the forecast probability against corresponding observed frequencies. For a perfectly-calibrated model the resulting curve should be aligned with the diagonal, subject to some error due to the finite-sample estimates used \cite{brocker2007increasing}. It is accompanied by a sharpness diagram showing the frequency with which each probability is forecast.

We estimate per-grid-cell predictive probabilities as the proportion $\hat{p}_{i}$ of the $N$ ensemble forecasts which exceed the threshold. For each of the $N+1$ possible values $p$ for $\hat{p}_{i}$, we compute the frequency with which the forecast is made (for the sharpness plot), and the observed frequency conditional on this forecast (for the reliability plot):
$$
f_{pred}(p) = \sum_i \hat{w}_i \mathbb{1}[\hat{p}_{i} = p]; \qquad
f_{obs|pred}(p) = \frac{\sum_i w_i \mathbb{1}[\hat{p}_{i} = p \wedge O_i \geq t]}{\sum_i w_i \mathbb{1}[\hat{p}_{i} = p]}.
$$

\subsubsection{Rank Histograms}

We use rank histograms \cite{hamill1997verification, hamill2001interpretation} to measure calibration of ensemble forecasts of continuous rainfall values. For each grid cell, the $N$ ensemble forecasts are ranked in increasing order, and the position of the observation in this ranking (from $0$ to $N$ inclusive) is computed, with tie-breaking as in \cite{hamill1997verification}. We then display the frequency of these ranks in a histogram pooled over all grid cells. For perfectly-calibrated forecasts this histogram is expected to be uniform.

We compute the histogram using the implementation in PySTEPS \cite{pulkkinen2019pysteps}. While we exclude masked grid cells, this approach does not allow us to incorporate the importance weights as we do for other metrics.
This does not change the property that a uniform histogram should be expected in the ideal case, however it will be biased to focus more on deviations from uniformity due to higher-rainfall examples.

\subsection{Pooled Neighborhood Metrics}
\label{sub:pooled-neigh-metrics}

These metrics are computed for a range of spatial scales $K$, and evaluate forecasts of observations that are pooled over local $K \times K$-grid-cell neighborhoods. 
This pooling allows some credit for a forecast which gets the ``big picture'' correct, even if smaller-scale weather patterns are not predicted correctly.
We average over all $K \times K$ neighborhoods of target grid cells in all examples in the evaluation dataset, subject to a horizontal and vertical stride of $\lceil K / 4 \rceil$. We weight each neighborhood $j$ using a weight
\begin{equation*}
w_j = \Bigl(\prod_{i \in j} m_{radar,i}\Bigr) \cdot q_{n_j}^{-1},
\end{equation*}
which is zero if any grid cell within it is masked by $m_{radar}$. These partially-masked neighborhoods are excluded as they do not in effect have the advertised scale of $K \times K$. Here $q_{n_j}^{-1}$ is the importance weight for the containing example $n_j$, see \cref{eqn:inclusion-probability}. We write $\hat{w}_j = w_j/\sum_{j'} w_{j'}$.

\subsubsection{Fractions Skills Score (FSS).}

FSS is defined by \cite{roberts2008scale}. In each neighborhood it evaluates the forecast
proportion of grid cells where rainfall exceeds a threshold $t$, against corresponding observed proportions, denoted respectively:
$$
P_F = \frac1{K^2} \sum_i \mathbb{1}[F_i \geq t]; \qquad
P_O = \frac1{K^2} \sum_i \mathbb{1}[O_i \geq t]
$$
where $i$ sums over the grid cells in the $K\times K$ neighborhood. For probabilistic forecasts, we generalize FSS to evaluate the expected forecast proportions $\mathbb{E}[P_F]$, similarly to the Neighborhood Ensemble Probability approach of \cite{schwartz2010toward}.
For each neighborhood we define the Fractions Brier Score (FBS), together with an upper bound $FBS_{worst}$, as:
$$
FBS = \bigl( \mathbb{E}[P_F] - P_O \bigr)^2; \qquad
FBS_{worst} = \mathbb{E}[P_F]^2 + P_O^2.
$$

Given an ensemble of $N$ forecasts $\{P_F^{(n)}\}$, we use biased estimators %
$$
\widehat{FBS} = \Bigl( \frac1N \sum_n P_F^{(n)} - P_O \Bigr)^2; \qquad
\widehat{FBS}_{worst} = \Bigl(\frac1N \sum_n P_F^{(n)}\Bigr)^2 + P_O^2
$$
Finally these are averaged over all neighborhoods $j$ and we compute the FSS, which lies in $[0, 1]$ with 1 the best score:
$$
\widehat{FSS} = 1 - \frac{\sum_j w_j \widehat{FBS}_j}{\sum_j w_j \widehat{FBS}_{worst,j}}
$$

\subsubsection{Pooled CRPS.}
We also compute CRPS using forecasts and observations which are pooled over local neighborhoods, using both average and max pooling. These are weighted using the neighborhood weights $w_j$.
Pooled CRPS evaluates more than just the per-grid-cell marginals of the predictive distribution, requiring some modelling of the dependence between nearby grid cells to accurately forecast the marginals for pooled values.
It can be motivated as a crude proxy for performance on tasks such as flood prediction, which require probabilistic forecasts of aggregate rainfall or maximum rainfall over broader catchment areas.

\subsection{Radially-averaged power spectral density (PSD)}
\label{sub:whole-frame-metrics}
We report radially-averaged power spectral density \cite{harris2001multiscale, sinclair2005empirical}, using the implementation from PySTEPS \cite{pulkkinen2019pysteps}. This measures how power is distributed across a range of spatial frequencies in each model's forecasts, compared with observations.

\putbib[refs]
\end{bibunit}

\end{document}